%% file: Colt2022_Final.tex
\documentclass[final,12pt]{colt2022}
 \pdfoutput=1
\input{defs}

\title[Rate-Distortion Theoretic Generalization Bounds]{Rate-Distortion Theoretic Generalization Bounds\\ for Stochastic Learning Algorithms}

\coltauthor{\Name{Milad Sefidgaran} \Email{milad.sefidgaran@telecom-paris.fr} \\ \addr LTCI, Télécom Paris, Institut Polyetechnique de Paris
\AND
\Name{Amin Gohari} \Email{amin.aminzadeh@gmail.com} \\ \addr Tehran Institute for Advanced Studies, Khatam University
\AND
\Name{Gaël Richard} \Email{gael.richard@telecom-paris.fr} \\ \addr  LTCI, Télécom Paris, Institut Polyetechnique de Paris
\AND
\Name{Umut \c{S}im\c{s}ekli} \Email{umut.simsekli@inria.fr} \\ \addr  INRIA \& ENS -- PSL Research University}

\begin{document}

\maketitle

\begin{abstract}%

Understanding generalization in modern machine learning settings has been one of the major challenges in statistical learning theory. In this context, recent years have witnessed the development of various generalization bounds suggesting different complexity notions such as the mutual information between the data sample and the algorithm output, compressibility of the hypothesis space, and the fractal dimension of the hypothesis space.  
While these bounds have illuminated the problem at hand from different angles, their suggested complexity notions might appear seemingly unrelated, thereby restricting their high-level impact.
In this study, we prove novel generalization bounds through the lens of rate-distortion theory, and explicitly relate the concepts of mutual information, compressibility, and fractal dimensions in a single mathematical framework. 
Our approach consists of (i) defining a generalized notion of compressibility by using \emph{source coding} concepts, and (ii) showing that the `compression error rate' can be linked to the generalization error both in expectation and with high probability. 
We show that in the `lossless compression' setting, we recover and improve existing mutual information-based bounds, whereas a `lossy compression' scheme allows us to link generalization to the \emph{rate-distortion dimension} -- a particular notion of fractal dimension. 
Our results bring a more unified perspective on generalization and open up several future research directions.

\end{abstract}


\tableofcontents


\section{Introduction}

Many important problems in statistical learning can be cast as the \emph{population risk minimization} problem, which is defined as follows \citep{shalev2014understanding}:
\begin{align}
    \min\limits_{w \in \mathcal{W}} \Bigl\{ \mathcal{L}(w)\coloneq \mathbb{E}_{Z \sim \mu}\left[\ell(Z,w)\right] \Bigr\},
\end{align}
where $\mathcal{W} \subset \mathbb{R}^d$ denotes a parametric \emph{hypothesis class}, $Z \in \mathcal{Z}$ denotes the \emph{input data} with $\mathcal{Z}$ being the \emph{data space}, $\mu$ denotes an unknown \emph{data distribution} over $\mathcal{Z}$, and $\ell \colon \mathcal{Z} \times \mathcal{W} \to \mathbb{R}^+$ is a \emph{loss function} that measures the quality of a hypothesis $w \in \mathcal{W}$.
As the data distribution $\mu$ is unknown in practice, we instead consider the \emph{empirical risk minimization} problem, given as follows:
\begin{align}
\label{eqn:erm}
    \min\limits_{w \in \mathcal{W}} \Bigl\{ \hat{\mathcal{L}}(S,w)\coloneq \frac{1}{n}\sum\limits_{i=1}^n\ell(Z_i,w) \Bigr\},
\end{align}
where $S:=\{Z_1,\ldots,Z_n\}$ denotes a \emph{training dataset} with independent and identically distributed (i.i.d.) elements, i.e., each $Z_i\sim_{\text{i.i.d.}} \mu$. 

To attack the optimization problem \eqref{eqn:erm}, arguably, the most common approach is to utilize a stochastic optimization algorithm $\mathcal{A} : \mathcal{Z}^n \to \mathcal{W}$ (e.g., stochastic gradient descent),
such that the algorithm outputs a \emph{random} hypothesis, i.e., $\mathcal{A}(S) = W \in \mathcal{W}$. 
One of the main challenges in statistical learning theory has been then to understand the behavior of the so-called \emph{generalization error} associated with the algorithm output, that is the difference between the population and empirical risks induced by the algorithm output: $\gen(S,\mathcal{A}(S))\coloneq \mathcal{L}(\mathcal{A}(S))-\hat{\mathcal{L}}(S,\mathcal{A}(S))$.
It has been illustrated that classical algorithm-independent generalization bounds fall short at explaining the (perhaps unexpected) success of modern machine learning systems \citep{Zhang16}. This has motivated the development of algorithm-dependent generalization bounds, a field that has been evolving in different directions.

An important direction in this context, and the one that is closest to our study, is based on analyzing the generalization error by using information-theoretic tools. Initiated by  \citet{russozhou16} and  \citet{xu2017information}, these approaches link the generalization error to the \emph{mutual information} between the data sample $S$ and the algorithm output $W$; suggesting that a lower statistical dependence between $S$ and $W$ implies better generalization. 
Their initial results were later improved by using different conditional versions of the mutual information \citep{harutyunyan2021,haghifam2021towards,negrea2020it,steinke2020reasoning,Bu2020,haghifam2020sharpened}, and were further generalized to more general notions of the mutual information that are defined through f-divergences (rather than the Kullback-Leibler divergence)  \citep{esposito2020,Hellstrom2020,masiha2021}.

 A second approach has been based on the observation that the algorithm output $W$ can be `compressible' in different senses. \citet{littlestone1986relating} in a pioneer work, considered a compressibility framework for the binary classification problem, in which \emph{compressed hypothesis} are chosen based on a subset of length $k$ of $S$ such that the picked hypothesis predicts correctly the label for all $Z_i \in S$. They showed that whenever such a compressing strategy exists,  the algorithm generalizes well. 
 The compressibility approach is later applied in different ways especially to  overparametrized neural networks 
 \citep{arora2018stronger,suzuki2018spectral, suzuki2020compression,negrea2020defense,hsu2021generalization,barsbey2021heavy,baykal2018datadependent,kuhn2021robustness}.
Loosely speaking, under different compressibility assumptions for $W$, these studies showed that a higher level of compressibility indicates a lower generalization error since the hypothesis class $\mathcal{W}$ can be approximated by a smaller, `compressed' space, which intuitively induces a lower worst-case error.

Finally, a recently initiated line of research has illustrated that when $\mathcal{A}$ is chosen as an iterative optimization algorithm, due to its recursive nature, $\mathcal{A}$ might generate a `fractal structure', either in its optimization trajectories \citep{simsekli2020hausdorff,birdal2021intrinsic,hodgkinson2021generalization}, or in the support of its stationary distribution  \citep{camuto2021fractal}. These studies showed that the generalization error can be linked to the `intrinsic dimension' of the fractal structure that is generated by the algorithm; suggesting that a smaller intrinsic dimension implies improved generalization.

Even though these three research directions have shed light on different fa\c{c}ades of the problem of understanding the generalization error, the mathematical frameworks that underlie their theoretical results and their implied take-home messages might be seemingly unrelated, thereby restricting their high-level impact.
In this paper, we prove novel generalization bounds through the lens of rate-distortion theory \citep{Berger1975}, and explicitly relate the concepts of mutual information, compressibility, and fractal dimensions in a single mathematical framework. 
 
To achieve this goal, we first define a generalized notion of compressibility by using \emph{source coding} concepts from information theory, which then allows us to use `information-theoretic coverings' for $\mathcal{W}$ that we will detail in Section~\ref{sec:compressibility}. Within this context, we show that the `compression error rate' of an algorithm $\mathcal{A}$ can be linked to its generalization error both in expectation and with high probability. 
Next, we show that the aforementioned
information-theoretical frameworks can be obtained as a special case of our setup, which is referred to as `lossless compression'. Thanks to this connection, the
results of \cite{xu2017information} can be re-derived. The bound in \cite[Theorem~1]{xu2017information} is in terms of the \emph{mutual information} between $S$ and $W$, denoted as $I(S;W)$, which was previously viewed as the dataset dependency of the algorithm. However, our framework reveals that it is an upper-bound on the compression rate in terms of \emph{lossless algorithm compressibility}. This new perspective allows us to introduce the notion of 
\emph{lossy algorithm compressibility} to handle continuous or large alphabets where \cite[Theorem~1]{xu2017information} can be vacuous as $I(S;W)$ can be very large; implying that a large $I(S;W)$ does not necessarily indicate the algorithm will not generalize as long as the algorithm is `lossily' compressible. We further established novel \emph{tail bounds} suggesting that $I(S;W)$ (or its lossy version) needs to be small not only for the underlying distribution of $(S,W)$, but also for any distribution in its vicinity. This is in the spirit of \emph{stability}: the algorithm should be compressible under any small perturbation of the dataset and hypothesis. The new tail bounds are established using a new `information-theoretic covering' technique, highlighted in Section~\ref{sec:proofOutline}.

Similarly, we derive and improve the results based on `conditional mutual information' \citep{steinke2020reasoning} in Appendix~\ref{sec:condComp}. Thanks to this approach, we established tail and in expectation bounds that recover the VC-dimension bounds (Corollaries~\ref{cor:VCexp} and \ref{cor:VCprob}); the recovery in terms of the tail bound is novel.
By exploiting the flexibility of our lossy compression framework, we further extend our results and  obtain bounds in terms of the intrinsic dimension of the marginal distribution of $W$, namely the \emph{rate-distortion dimension} \citep{Kawabata1994}. 
Our results bring a unified perspective on mutual information, compressibility, and fractal dimensions, and open up several future research directions as we will point out in Section~\ref{sec:conc}.
%


\section{Preliminaries}


\subsection{Notation and problem setup}


Random variables, their realizations, and their domains are denoted by upper-case letters, lower-case letters, and calligraphy fonts, \emph{e.g.} $X$, $x$, and $\mathcal{X}$. We assume that all the domains are endowed with their Borel sigma fields. By $P_X$, we denote the distribution of $X$, defined on some measurable space $(\mathcal{X},\mathcal{F})$, and by $\Supp(P_X)$ we denote its support. The expected value of $X$ is denoted by $\mathbb{E}[X]$. We call a random variable $X$ (absolutely) continuous if it admits a density with respect to the Lebesgue measure. 
The random variable $X$ is called $\sigma$-subgaussian, if $\mathbb{E}[\exp(t(X-\mathbb{E}[X]))
]\leq \exp(\sigma^2t^2/2)$, $\forall t \in \mathbb{R}$. A collection of $m \in \mathbb{N}$ random variables is denoted by $X^m=(X_1,\ldots,X_m)$, or simply by bold letters $\vc{X}$, when $m$ is known by the context. A sequence of $m$ real numbers  $x_1,\ldots,x_m$ is denoted by $\{x_i\}_{i=1}^m$. Similar conventions are used for sequences of sets or functions. The set of integers $\{1,\ldots,m\}$ is denoted by $[m]$. We use $\mathbb{R}^+$ to denote nonnegative real numbers.

As mentioned in the introduction, we consider a generic \emph{randomized algorithm}  $\mathcal{A} : \mathcal{Z}^n \to \mathcal{W}$, that has access to dataset $S=\{Z_1,\ldots,Z_n\}$. This randomized algorithm induces a conditional distribution $P_{W|S}$. 
We denote the joint distribution of the dataset $S$ and the hypothesis $W$ by $P_{S,W}=\mu^{\otimes n}P_{W|S}$ and the marginal distribution of $W$ by $P_W$. 

Most of our results are expressed in terms of information-theoretic constructs, which we define as follows. For discrete random variables, the Shannon entropy function is defined as $H(X)\coloneq \mathbb{E}[\log(1/P_X(X))]$. Similarly, conditional entropy is defined as $H(X|Y)\coloneq \mathbb{E}[\log(1/P_{X|Y}(X|Y))]$.  The mutual information between $X$ and $Y$ is defined as $I(X;Y)\coloneq H(X)-H(X|Y)$, and intuitively measures the amount of \emph{information} these random variables contain about each other. For continuous random variables, the differential entropy $h(X)$ is defined as $-D_{KL}(P_{X}\|\leb)$, where $\leb$ is the Lebesgue measure on Euclidean spaces. In particular, if $X$ has pdf $\mathrm{p}$, then $h(X)=\mathbb{E}[\log(1/\mathrm{p}(X))]$, and $-\infty$ otherwise. Similarly, $h(X|Y)$ and $I(X;Y)\coloneq h(X)-h(X|Y)$ are defined. The Kullback–Leibler (KL) divergence between two distributions $Q$ and $P$ defined on the same measurable space is defined as $D_{KL}(Q\|P) \coloneq \mathbb{E}_{Q}\left[\log \frac{\mathrm{d}Q}{\mathrm{d}P}\right]$, when $Q\ll P$, and equals $\infty$, otherwise. Here, $\frac{\mathrm{d}Q}{\mathrm{d}P}$ is the Radon-Nikodym derivative of $Q$ with respect to $P$. 


\subsection{Technical background on source coding}
In this section, we will briefly review some results from the literature on source coding that will ease the introduction of our theoretical framework.\footnote{For a more detailed introduction, we refer the reader to  \citep{Berger1975,CoverTho06,CsisKor82,elgamal_kim_2011,polyanskiy2014lecture}.}
Consider a random variable $W$ taking values in a finite set $\mathcal{W}$. It is well-known that one can represent $W$ using $\lceil\log_2(|\mathcal{W}|)\rceil$\footnote{For $a \in \mathbb{R}$, $\lceil a \rceil$ denotes the ceiling of $a$,\emph{i.e.} $\min_{n \in \mathbb{N}}$, such that $n \geq a$. }  bits,\footnote{Depending on the base of logarithm in the Shannon entropy function, the unit of information is either \emph{bit} (base 2: $\log_2$) or \emph{nat} (base $e$: $\log_e$). We state all results with base $e$ for simplicity and compatibility with previous results. However, for the unit of information we use bit, as it is more common in the (digital) source coding context.} from which $W$ can be recovered with no error. For instance, if 
$W$ is a Bernoulli random variable, i.e., $\mathbb{P}(W = 1) = 1-\mathbb{P}(W = 0) = \theta$,
one bit suffices to represent $W$. However, intuitively speaking if $\theta$ is very close to zero or very close to one, using one full bit to represent $W$ is wasteful because in such cases $W$ is almost deterministic. 

In his seminal paper, \citet{Shan49} formalized this intuition by introducing the concept of \emph{`block-coding'}, where he showed that the \emph{`source'} $W$ can be represented in a \emph{compressed} way by using a significantly smaller number of bits, provided we can allow for a negligible probability of recovery error. The main idea behind block coding can be summarized as follows. As opposed to considering a single realization of the source $W$, we instead assume that we have access to a vector of $m$ \emph{independent} realizations of the source, denoted by $W^m$, and we are allowed to compress these $m$ instances \emph{simultaneously}. Moreover, the zero-error constraint in recovering $m$-instances is replaced by the `asymptotically negligible error' criterion (i.e., the reconstruction error vanishes as $m \ra \infty$). It turns out that joint description of such independent sources is more efficient than their individual description \citep{CoverTho06}. For instance, in our running example of  $W$ being a Bernoulli variable with parameter $\theta$, $W^m$ is a binary string of length $m$. By the law of large numbers, we expect $W^m$ to have around $m\theta$ ones in it. So, even though there are $2^m$ binary strings of length $m$, roughly speaking $W^m$ has about
\begin{align}\binom{m}{m\theta}\approx e^{mH(W)}\label{SF}
\end{align}
`effective' possibilities where in \eqref{SF} we use Stirling's approximation of factorial to express the number of possibilities in terms of the $H(W)$, the Shannon entropy of $W$. 
Intuitively, for a discrete random variable $W$, `the effective size' of $m$ independent realizations is asymptotically about $e^{mH(W)}$, rather than $|\mathcal{W}|^m$. Concretely, there exist sets $\{\mathcal{C}_m\}_{m \in \mathbb{N}}$, $\mathcal{C}_m \subseteq \mathcal{W}^m$ with $|\mathcal{C}_m|=e^{m(H(W)+\varepsilon_m)}$ such that $\lim_{m \ra \infty} \mathbb{P}(W^m \in \mathcal{C}_m)=1$ and $\lim_{m \ra \infty} \varepsilon_m =0$. Thus, Shannon showed that the fundamental limit for the compression of information is determined by the Shannon entropy function $H(W)$, which can be much smaller than $\log(|\mathcal{W}|)$. 

Unfortunately, the number of bits required to represent even a single realization of a continuous random variable $W$ is infinity. For instance, if $W$ is a uniform random variable on $[0,1]$, we need infinitely many bits to convey it. However, one bit is enough to represent a single realization of $W$ within distance $0.5$ by mapping $W$ to either $\hat W=0$ if $W<0.5$, and to $\hat W=1$ if  $W\geq 0.5$, and then conveying $\hat W$ instead of $W$. In this example, the reconstruction space (or the set of quantization points) is $\hat{\mathcal{W}}=\{0,1\}$, the reconstruction is \emph{lossy} (almost surely we never recover the original $W$) and we measure the distance (or the distortion) between $W$ and its reconstruction $\hat W$ by $|W-\hat W|$. 

Let us begin by describing the lossy compression of a \emph{single instance} of an arbitrary random variable $W$. In many information-theoretic and signal processing applications, it suffices to recover a \emph{distorted} version $\hat{W} \in \hat{\mathcal{W}}$ of $W$, as long as the incurred distortion is within an `acceptable' range, \emph{i.e.} for $\epsilon\geq 0$ and a chosen \emph{distortion function} $\dw \colon \mathcal{W}\times\hat{\mathcal{W}}\to \mathbb{R}^+$, we have $\dw(W;\hat{W})\leq \epsilon$. While the quantized (distorted) space $\hat{\mathcal{W}}$ is equal to $\mathcal{W}$ in many cases, it can be different in general.\footnote{For example, to convey the sign of $W \in \mathcal{W}=\mathbb{R}$, it is natural to consider $\hat{\mathcal{W}}=\{-1,+1\}$ and $\dw(w,\hat{w})=\mathbbm{1}_{\{w \hat{w}<0\}}$.} To facilitate the explanation, assume $\hat{\mathcal{W}}=\mathcal{W}$ for the rest of this section, \emph{i.e.,} we are required to produce  $\hat W\in\mathcal{W}$.
Next, let us consider the case of block coding where instead of a single realization of the source $W$, we have access to $W^m$ which is a vector of $m$ \emph{independent} realizations of the source. This problem is known as the  vector-quantization problem. Intuitively speaking, to compress $W^m$ we can take a collection of $k$ quantization points $\hat{\vc{w}}_1, \hat{\vc{w}}_2, \ldots, \hat{\vc{w}}_k$ in $\mathcal{W}^m$, where $\hat{\vc{w}}_j=(\hat{w}_{j,1},\ldots,\hat{w}_{j,m})$ for $j \in [k]$, and map $W^m=(W_1,\ldots,W_m)$ to its closest quantization point. Here, the distortion function between $W^m$ and a quantization point $\hat{\vc{w}}_j$ should be defined; it is often chosen to be the average of coordinate-wise distortions:
\begin{align*}
\frac{1}{m}\sum \nolimits_{i=1}^m \dw(W_i,\hat{w}_{j,i}).   
\end{align*}
Because the number of quantization points is $k$, we require $\log_2(k)$ bits to convey the index of the quantization point. The ratio $\log_2(k)/m$ is called the compression \emph{rate}, because it represents the number of compression bits per source realization.
For the selection of quantization points $\hat{\vc{w}}_1, \hat{\vc{w}}_2, \ldots, \hat{\vc{w}}_k$ to succeed, we can consider balls of radius $\epsilon$ around these quantization points and require that with high probability $W^m$ falls into the union of these balls. This can be seen as a ``block covering'' of 
$\mathcal{W}^m$ with average distortion $\epsilon$. 
Note that block covering may need a smaller number of quantization points than the case where the complete covering of the space $\mathcal{W}^m$ with the worst-case distortion $\epsilon$ is required, as in $\epsilon$-net coverings \citep{anthonyBartlett99}. 

In this context, the goal becomes finding the minimum number of bits that is required to compress i.i.d.\ repetitions of the source $W$ so that it can be recovered within a given distortion margin. 
\cite{Shan49} showed that the minimum compression rate needed for recovering a source with distortion $\epsilon$ is determined by the \emph{rate-distortion function} 
\begin{align*}
\mathfrak{RD}(\epsilon;P_W,\dw)\coloneq \inf I(W;\hat{W}), \quad \text{such that} \quad
    \mathbb{E}_{W,\hat{W}}[\dw(W,\hat{W})]\leq \epsilon,
\end{align*}
where the infimum is over all conditional probability distributions (Markov kernels) $P_{\hat{W}|W}$. Specifically, Shannon showed that for any \emph{rate} $R>\mathfrak{RD}(\epsilon;P_W,\dw)$,  a sequence of \emph{quantization codebooks}  $\{\mathcal{C}_m\}_{m \in \mathbb{N}}$,  $\mathcal{C}_m{=}\{\hat{\vc{w}}_j{=}(\hat{w}_{j,1},\ldots,\hat{w}_{j,m}),j\in[l_m]\} \subseteq \mathcal{W}^m$ exists such that $l_m \leq e^{mR}$ and 
\begin{align}
    \lim_{m \ra \infty}\mathbb{P}_{W^{\otimes m}}\left(\forall j\colon \frac{1}{m}\sum \nolimits_{i=1}^m \dw(W_i,\hat{w}_{j,i}) > \epsilon \right)=0.\label{error-prob-eq}
\end{align}
Intuitively, for discrete variables, the \emph{effective size} of $m$ independent realizations of $W$ is about $e^{mH(W)}$, and each codeword $\hat{\vc{w}}_j$ covers about $e^{mH(W|\hat{W})}$ of them, and thus, the total needed codewords to cover $\mathcal{W}^m$ with high probability is about $e^{mI(W;\hat{W})}$. Similar intuition holds for the continuous $W$, by considering $h(W)$ and  $h(W|\hat{W})$, and by considering the \emph{effective volume} of $W^m$.

Finally, a series of works, \emph{e.g.} \citep{marton1974,Han2000,Iriyama2005,bakshi2005error}, studied the rate of convergence of the probability in \eqref{error-prob-eq} to zero for a fixed rate $R$. Equivalently, one can formulate this problem as the minimum needed rate $R$ to have the above error probability decaying at least as fast as $\delta^m$. For sources with finite alphabets, this quantity is equal to \cite[Theorem~1]{marton1974} $ \sup \limits_{\substack{Q \colon D_{KL}(Q\|P_{W}) \leq \log\left(1/\delta\right)}} \mathfrak{RD}(\epsilon;Q,\dw)$, where the supremum is over all distributions $Q$ defined over $\mathcal{W}$ such that $Q \ll P_W$. Intuitively, the empirical distribution $\hat{P}_{W^m}$ of a vector of realizations $W^m$ satisfies $D_{KL}(\hat{P}_{W^m}\|P_{W}) \leq \log\left(1/\delta\right)$, with probability at least $1-\delta^m$. The idea is to ``cover'' all such high probable realizations in the balls with radius $\epsilon$. It turns out the needed rate is the supremum of the needed rate for each empirical distribution $Q$.  Similar error exponent term for continuous sources can be found in \cite[Theorem~1]{Iriyama2005}.

In this work, we apply source coding concepts and techniques to establish bounds on the generalization error. To this end, we attempt to `reliably compress' the hypothesis space with respect to a distortion that depends on the excess generalization error induced by compression. We allow the compression to be lossy within a distortion level. Then, we establish bounds on the generalization error in terms of the compression rate, amount of distortion, and  reliability level.\footnote{The rate-distortion theory was previously used in \citep{Bu2021ModelCompression,masiha2021}. For instance, in \citep{Bu2021ModelCompression}, it is used to compare the expectation of the generalization error of a \emph{compressed} learning model with respect to the \emph{original} model. Herein, we use it to analyze the generalization performance of the \emph{original} learning model. In \cite{masiha2021}, generalization error is related to the rate-distortion theory by noting the similarity of the related formulas. The connection provided in this work is operational and thus much deeper. }


\section{Generalization Bounds via Rate Distortion Theory}
\label{sec:compressibility}
We start by explaining our notion of compressibility adapted to algorithms.


\subsection{Compressibility of an algorithm}
The compression, in its classical source coding sense, aims to save a compressed version of a source that is \emph{close enough} to the source and requires a smaller storage capacity. Similarly, for a learning algorithm $\mathcal{A}\colon \mathcal{S}\to \mathcal{W}$, where $\mathcal{S}=\mathcal{Z}^n$, by having a dataset $S$ and a picked hypothesis choice $\mathcal{A}(S)=W \in \mathcal{W}$, we are interested in finding another algorithm $\hat{\mathcal{A}}(S,W)=\hat{W} \in \hat{\mathcal{W}} \subseteq \mathcal{W}$ that has fewer number of probable output hypotheses and \emph{performs closely} to the original algorithm. In this work, we consider the generalization error as the compression performance. Consider a training dataset $s$ and two hypotheses $w$ and $\hat{w}$. We define the distortion function $ \dd \colon  \mathcal{W}  \times \hat{\mathcal{W}} \times \mathcal{S} \to \mathbb{R}$ between these two pairs of realizations as the difference of their generalization performances:\footnote{While $d$ is clearly not a metric, it also depends on the underlying distribution $\mu$; we drop this dependence for ease of notations.}
\begin{align}
  \dd(w,\hat{w};s) \coloneq & \gen(s,w)-\gen(s,\hat{w}). \label{def:distanceAbs}
\end{align}
Note that here, unlike the source-coding literature, we allow the distortion function to take negative values.

To guarantee that this distortion (between single outputs of the original and compressed algorithms) does not exceed a threshold, we need to control the \emph{worst-case} distortion caused by compression, among all probable $w$ and $\hat{w}$, which might end up with overly pessimistic results. To avoid this, we utilize the \emph{block coding} technique as follows. For a block of $m \in \mathbb{N}$ independent datasets $s^m=(s_1,\ldots,s_m)$ and a block of picked hypotheses $W^m=(W_1,\ldots,W_m)$, where $W_i$ is a hypothesis choice based on dataset $s_i$, {i.e.} $\mathcal{A}(s_i)=W_i$, $i \in [m]$, with a slight abuse of notations, denote $\mathcal{A}(s^m)=W^m$. We then consider a compression algorithm  $\hat{\mathcal{A}}_m \colon \mathcal{S}^m \times \mathcal{W}^m \to \hat{\mathcal{W}}^m$ that takes as input particular realizations $(s^m,w^m)$, where $\mathcal{A}(s^m)=w^m$, and outputs a block of hypotheses $\hat{W}^m=(\hat{W}_1,\ldots,\hat{W}_m)$. We also need to extend our definition in \eqref{def:distanceAbs} to measure the distortion between two blocks of algorithm realizations $\mathcal{A}(s^m)=w^m$ and $\hat{\mathcal{A}}_m(s^m,w^m)=\hat{w}^m$. 
For now, let us use $\ddm \colon  \mathcal{S}^m \times \mathcal{W}^m \times \hat{\mathcal{W}}^m \to \mathbb{R}$ for $m\in\mathbb{N}$ to denote this extended distortion function, whose details will be provided in the next section. 
In particular, we will use the extended distortion function defined in \eqref{def:vectorDistanceE} to obtain in expectation bounds and an alternative definition given in \eqref{def:vectorDistanceP} to obtain tail bounds on the generalization gap.

Next, we define our compression algorithm. Fix a set $\mathcal{H}_m \subseteq \hat{\mathcal{W}}^{m}$, that we coin a \emph{hypothesis book} (as an analogy to code book), and denote its cardinality by $l_m$. Denote the elements of $\mathcal{H}_m$ by $\hat{\vc{w}}_j=(\hat{w}_{j,1},\ldots,\hat{w}_{j,m}) \in \hat{\mathcal{W}}^{m}$, where $j \in [l_m]$, i.e., $\mathcal{H}_m = \{\hat{\vc{w}}_j\}_{j=1}^{l_m}$. Having defined a distortion function $d_m$ and fixed a set $\mathcal{H}_m$, among all compression algorithms  $\hat{\mathcal{A}}_m$ such that $\hat{\mathcal{A}}_m(s^m,w^m) \in \mathcal{H}_m$, we consider the \emph{optimal} compression algorithm, denoted by $\hat{\mathcal{A}}^*_m(s^m,w^m;\mathcal{H}_m)=\vc{\hat{w}}_j$, where $j= \argmin \nolimits_{j \in [l_m]} d_m(w^m,\hat{\vc{w}}_j;s^m)$. With this choice and for a fixed distortion level $\epsilon$, we define the error event that happens when the average distortion between the original and the optimal compressed algorithm exceeds $\epsilon$: 
\begin{align}
    \mathcal{E}_m(\mathcal{H}_m,\epsilon;d_m) \coloneq \left\{\min \nolimits_{j \in [l_m]} d_m\left(\mathcal{A}(S^m),\hat{\vc{w}}_j;S^m\right) > \epsilon\right\}. \label{def:error}
\end{align}

Now, we are ready to define our compressibility notion, which will lay the basis of our generalization bounds.
\begin{definition} \label{def:compressibility}
The learning algorithm $\mathcal{A}$ 
is $(R,\epsilon;\left\{d_m\right\}_{m})$-compressible\footnote{While many terms in this work, including $R$ and the rate-distortion terms in the rest of the text, depend
on $\mu$, $P_{W|S}$, $n$, and the loss function, we drop these dependencies for ease of exposition.} for some $R \in \mathbb{R}^+$ and $\epsilon \in \mathbb{R}$, if there exists a sequence of hypothesis books $\left\{\mathcal{H}_m\right\}_{m\in \mathbb{N}}$, $\mathcal{H}_m=\{\vc{\hat{w}}_j, j \in [l_m]\} \subseteq \mathcal{\hat{W}}^m$ such that  $l_m \leq  e^{mR}$ and 
\begin{align}
   \lim_{m \ra \infty}  \mathbb{P}_{(S,W)^{\otimes m}}\left(\mathcal{E}_m(\mathcal{H}_m,\epsilon;d_m)\right)=0, \label{eq:errorZero}
\end{align} 
where $\mathcal{E}_m(\mathcal{H}_m,\epsilon;d_m)$ is defined in \eqref{def:error} and $\mathbb{P}_{(S,W)^{\otimes m}}$ denotes the $m$-times product measure of the joint distribution of $W$ and $S$.
\end{definition}


\subsection{Bounds on the expected value of the generalization gap}
In this section, we prove bounds on the expected generalization error, provided $\mathcal{A}$ is compressible. Intuitively, we first find compression schemes that \emph{cover} $(S^m,W^m)$ with high probability, in a sense that is defined in \eqref{eq:errorZero}, such that on average the difference of generalization errors of the original and compressed algorithms does not exceed a threshold. Then, we show that the expected generalization error can be bounded in terms of the parameters of this compressed algorithm. To do so, by borrowing from the source coding literature, we define a distortion function between $m$ realizations of the two algorithms as:
\begin{align}
    \dem(w^m,\hat{w}^m;s^m)\coloneq \frac{1}{m} \sum  \nolimits_{i=1}^m \dd(w_i,\hat{w}_i;s_i), \label{def:vectorDistanceE}
\end{align}
where $\dd(w,\hat{w};s)$ was defined in \eqref{def:distanceAbs}.

Having condition \eqref{eq:errorZero} for this distortion function guarantees that the expectation of the difference of the generalization errors  of the original and compressed algorithms does not exceed $\epsilon$. This is stated in Lemma~\ref{lem:probToExpec}, which is used in the proof of the following result, 
proved in Appendix~\ref{pr:TheoremExpComp}.

\begin{theorem} \label{th:expectationCompressible}   If a learning algorithm $\mathcal{A}(S)$ 
is $(R,\epsilon;\left\{|\dem|\right\}_{m})$-compressible,\footnote{By $|\dem|$ we mean simply the distortion function which is  equal to $|\dem(w^m,\hat{w}^m;s^m)|$ for any $w^m,\hat{w}^m,s^m$.}, if $\mathbb{E}_{S,W}[|\gen(S,W)|]<\infty$, and if for all $w \in \mathcal{W}$, $\ell(Z,w)$ is $\sigma$-subgaussian, then $\left|\mathbb{E}\left[\gen(S,W)\right]\right| \le\sqrt{2\sigma^2 R/n}+\epsilon$.
\end{theorem}
This result shows that the compressibility of an algorithm directly translates into having a good generalization performance, which can be seen as an information theoretic counterpart of the existing compression bounds, e.g., \citep{arora2018stronger,suzuki2020compression}. To make the above bound more explicit, we establish the following bound on the compressibility of any arbitrary algorithm, whose proof is given in Appendix~\ref{pr:compressibilityBound}. Let\footnote{Intuitively, $\mathbb{E} \left[\gen(S,W)-\gen(S,\hat{W})\right]$ can be seen as the limit of the distortion function $\dem$ when $m \ra \infty$.}
\begin{align}
    R_E(\epsilon) =& \inf \limits_{P_{\hat{W}|S}}I(S;\hat{W}),\quad \text{such that} \quad \left|\mathbb{E} \left[\gen(S,W)-\gen(S,\hat{W})\right]\right|\leq \epsilon, \label{eq:inExpTerm}
\end{align} 
where the expectation is with respect to $P_{S,W}$ and $P_{S} \times P_{\hat{W}|S}$.
\begin{theorem}\label{th:compressibilityBound} Assume that the algorithm $\mathcal{A}(S)=W$ 
induces $P_{S,W}$, where $\mathcal{S}$ and $\mathcal{W}$ are finite sets. Then, for every $\epsilon \in \mathbb{R}$ and any $\nu_1,\nu_2>0$, the algorithm $\mathcal{A}(s)$ is $\left(R_E(\epsilon)+\nu_1,\epsilon+\nu_2;\{|\dem|\}_m\right)$-compressible.
\end{theorem}
This theorem can be extended to infinite sets, with some further assumptions on separability of $\gen(S,W)$ with respect to $(S,W)$ and using the quantization technique used in the proof of \cite[Theorem~3.6]{elgamal_kim_2011}. Now, combining Theorems~\ref{th:expectationCompressible} and \ref{th:compressibilityBound} yields:
\begin{theorem} \label{th:expecRdDis} Assume that the algorithm $\mathcal{A}(S)=W$ 
induces $P_{S,W}$ and for all $w \in \mathcal{W}$, $\ell(Z,w)$ is $\sigma$-subgaussian. Then, for any $\epsilon \in \mathbb{R}$, $\left|\mathbb{E}\left[\gen(S,W)\right]\right| \leq \sqrt{2\sigma^2 R_E(\epsilon)/n}+\epsilon$.
\end{theorem}
The extended versions of Theorems~\ref{th:expectationCompressible} and \ref{th:expecRdDis} that include bounds on $\mathbb{E}\left[\gen(S,W)\right]$ and $\mathbb{E}\left[|\gen(S,W)|\right]$ as well, can be found in Appendix~\ref{sec:expecRdDis2}.\footnote{The mild sufficient condition $\mathbb{E}[|\gen(S,W)|]{<}\infty$ is used to bound  $\mathbb{E}[|\gen(S,W)|]$, $\mathbb{E}[\gen(S,W)]$, and $|\mathbb{E}[\gen(S,W)]|$  in the extended version of Theorem~{\ref{th:expectationCompressible}}. The sufficiency of the condition $|\mathbb{E}[\gen(S,W)]|{<}\infty$ for bounding $|\mathbb{E}[\gen(S,W)]|$, although seemingly true, is not shown in this work.}

In addition to finite sets, Theorem~\ref{th:expecRdDis}  can be derived using Theorems~\ref{th:expectationCompressible} and \ref{th:compressibilityBound} also for the infinite sets that satisfy some further separability assumptions. However, for the infinite set, without any further assumptions, we show this alternatively and trivially in Appendix~\ref{pr:expecRdDis} by using and extending the existing results of \citet[Theorems 1, 4]{xu2017information}, corresponding to Theorem~\ref{th:expecRdDis} with $\epsilon=0$ (see Corollary~\ref{cor:losslessExp} in below). Theorem~\ref{th:expecRdDis} is extended similarly to \citep{Bu2020} in Theorem~\ref{th:expecRdDis2} (Appendix~\ref{sec:expecRdDis2}) that recovers and potentially improves over \cite[Proposition~1]{Bu2020}.

\begin{corollary} \label{cor:losslessExp}
Suppose the algorithm $\mathcal{A}(S)=W$ 
induces $P_{S,W}$ and the loss function $\ell(Z,w)$ is $\sigma$-subgaussian for any $w \in \mathcal{W}$. Then, $ \left|\mathbb{E}\left[\gen(S,W)\right]\right| \le\sqrt{2\sigma^2I(S;W)/n}$.
\end{corollary}
In this corollary, by applying a compressibility approach we could recover the results obtained using the Donsker–Varadhan's identity. Indeed, in Appendix~\ref{sec:donskerCompr} we showed that this identity can be interpreted and derived via a compressibility approach. 

The case of $\epsilon=0$, considered in \citep{xu2017information,Bu2020}, corresponds to the lossless compression in source coding. While for countable sets of $S$ or $W$, we can reliably cover $(S^m,W^m)$ with $\epsilon=0$ (in the sense of \eqref{eq:errorZero}) and bounded $R$, for continuous sources and hypotheses, this term could be infinite. In contrast, considering $\epsilon \neq 0$, corresponds to the lossy compression in source coding, which allows to reliably cover $(S^m,W^m)$ with bounded $R$ within distortion $\epsilon$. Note that even for countable sets, $\epsilon \neq 0$ can give better bounds.\footnote{The approach applied in in \citep{negrea2020defense} for studying $\mathbb{E}[\gen(S,W)]$ also can be seen as lossy compression. They considered choosing a randomized `surrogate hypothesis' $\hat{W}$ for each $W$, and argue that to establish a good bound on $\mathbb{E}[\gen(S,W)]$, one could benefit from the trade-off between $\mathbb{E}[\hat{\mathcal{L}}(S,\hat{W})-\hat{\mathcal{L}}(S,W)]$ and $\mathbb{E}[\mathcal{L}(\hat{W})-\hat{\mathcal{L}}(S,\hat{W})]$. However, they have not proposed general explicit bounds on these terms, and rather considered  ad-hoc strategies for overparameterized linear regression and hypercube classification problems, when $\hat{\mathcal{L}}(S,W)=0$.}

The benefit of $\epsilon \neq 0$ becomes more clear by having a Lipschitz loss assumption. Combining this assumption with the above theorem directly yields an upper bound on the expected  generalization error in terms of the rate-distortion function of the hypothesis.

\begin{corollary}[Lipschitz loss] \label{cor:lipschitzExp} Suppose that for a distortion function $\dw\colon \mathcal{W} \times \hat{\mathcal{W}} \to \mathbb{R}^+$ and every $z,w,\hat{w}$, $|\ell(z,w)-\ell(z,\hat{w})| \leq \mathfrak{L} \dw(w,\hat{w})$\footnote{Note that this condition and imposing the Markov chain $\hat{W}-W-S$ yield $\left|\mathbb{E} \left[\gen(S,W)-\gen(S,\hat{W})\right]\right|\leq 2 \mathfrak{L} \mathbb{E}_{W,\hat{W}}\left[\dw(W,\hat{W})\right]$ and $I(S;\hat{W})\leq I(W;\hat{W})$ by data processing inequality.} and $\ell(Z,w)$ is $\sigma$-subgaussian. Then, for any $\epsilon \in \mathbb{R}^+$, we have $ \left|\mathbb{E}\left[\gen(S,W)\right]\right| \leq  \sqrt{2\sigma^2\mathfrak{RD} (\epsilon/(2\mathfrak{L});P_W,\dw)/n}+\epsilon$, where $\mathfrak{RD}(\epsilon;P_W,\dw)$ is the rate-distortion function with respect to the distortion function $\dw$: 
\begin{align}
    \mathfrak{RD} (\epsilon;P_W,\dw)\coloneq \inf \limits_{\substack{P_{\hat{W}|W}}} I(W;\hat{W}), \quad \text{such that} \quad \mathbb{E}_{W,\hat{W}}\left[\dw(W,\hat{W})\right]\leq \epsilon.
\end{align}
\end{corollary}

While the term $R_E(\epsilon)$ in \eqref{eq:inExpTerm} is in general intractable, the above bound is amenable to computation once the marginal distribution $P_{W}$ is known; a more relaxed constraint than knowing $P_{S,W}$ which is needed in many of the information-theoretic bounds on generalization error. The rate-distortion computation is a convex minimization problem over $P_{\hat W|W}$ and $\epsilon$, that can be effectively computed for finite alphabets using Blahut-Arimoto algorithm \citep{Blahut72,Arimoto72}. Note that using Carathéodory's theorem \cite[Appendix~C]{elgamal_kim_2011}, it can be shown that it is sufficient to consider $\hat{\mathcal{W}}$ such that $|\hat{\mathcal{W}}|\leq |\mathcal{W}|+1$. For the continuous alphabets, this terms can be efficiently estimated using the fine quantization technique (\emph{e.g.} \cite[Proof of Theorem~3.6]{elgamal_kim_2011} or by using the existing lower bounds, \emph{e.g.} \citep{Riegler2018}, that are almost tight in the small $\epsilon$ regime.

As an analytical example, suppose that the data $Z \in \mathbb{R}^d$ is composed of $d$ i.i.d. elements, each one distributed according to the normal distribution $\mathcal{N}(0,\sigma_0^2)$ and suppose that we choose $W$ as $W=\frac{1}{n}\sum_{i=1}^n Z_i\sim \mathcal{N}(0,(\sigma_0^2/n)\mathrm{I}_d)$, where $\mathrm{I}_d$ is the $d \times d$ identity matrix. Further, suppose that $\ell(Z,w)$ is $\sigma$-subgaussian for any $w \in \mathbb{R}^d$ and $|\ell(z,w)-\ell(z,\hat{w})| \leq \mathfrak{L} \|w-\hat{w}\|^2$.  Then, while  $I(S;W)=\infty$, Corollary~\ref{cor:lipschitzExp} together with \cite[Theorem~10.3.2]{CoverTho06} yield $ \left|\mathbb{E}\left[\gen(S,W)\right]\right|$ is bounded by $\min_{0 < \epsilon \leq  d \sigma_0^2/n}\sqrt{\sigma^2 d \log\left(2\mathfrak{L}d\sigma_0^2/(n\epsilon)\right)/n}+ \epsilon$, which equals  $2\mathfrak{L}d\sigma_0^2/n$ for $\epsilon = 2\mathfrak{L}d\sigma_0^2/n$. 

The optimal order of $\epsilon$ (and the corresponding rate-distortion terms) with respect to $n$ depends on $P_{S,W}$, as well as the loss function. In the above example, $\epsilon$ is chosen as $\mathcal{O}(1/n)$, and in the following corollary as $\mathcal{O}(1/\sqrt{n})$, resulting in the rate-distortion terms of order $\mathcal{O}(1)$ and $\mathcal{O}(\log(n))$, respectively.

In our next result, we show that our bound in terms of the rate-distortion function yields a fractal dimension-based bound as well. Let us define the rate-distortion dimension \citep{Kawabata1994} for a distribution $Q$ as $\dim_{\mathrm{R}}(Q)\coloneq \limsup \nolimits_{\epsilon \ra 0} \mathfrak{RD} (\epsilon;Q,\dw)/\log(1/\epsilon)$. 
\begin{corollary}[Rate-distortion dimension] \label{cor:dimExpec} Suppose that for a distortion function $\dw \colon \mathcal{W} \times \hat{\mathcal{W}} \to \mathbb{R}^+$ and every $z,w,\hat{w}$, $|\ell(z,w)-\ell(z,\hat{w})| \leq \mathfrak{L} \dw(w,\hat{w})$ and $\ell(Z,w)$ is $\sigma$-subgaussian. Moreover, assume that $\sup \nolimits_{\epsilon \leq \epsilon_0}  \mathfrak{RD} (\epsilon;P_W,\dw)/\log(1/\epsilon)$ converges uniformly over $n$ as $\epsilon_0 \to 0$.  Then, there exists a  $n_0$ such that for every $n \geq n_0$, $ \left|\mathbb{E}\left[\gen(S,W)\right]\right| \leq \sqrt{4\sigma^2 \dim_{\mathrm{R}}(P_W) \log(n\mathfrak{L}^2)/n}$.
\end{corollary}
This corollary, proved in Appendix~\ref{pr:dimExpec}, shows the relation of our approach with dimension-based bounds. The rate-distortion dimension is a lower bound to the Minkowski (box-counting) dimension of the set $\mathcal{W}$ (\cite{Kawabata1994}), which was considered in \citep{simsekli2020hausdorff,birdal2021intrinsic}.
%
Moreover $\dim_{\mathrm{R}}(P_W)$ is equal to the R\'enyi information dimension under certain conditions \cite[Proposition~3.3]{Kawabata1994}. The latter dimension is shown to be related to the fundamental limits of the almost lossless compression \citep{Wu2010Renyi}. 


\subsection{Tail bounds on the generalization gap} \label{sec:tailMain}
To establish a tail bound on the generalization performance, we need to find a compression scheme that not only \emph{covers} $(S^m,W^m)$ with high probability (in a sense of \eqref{eq:errorZero}), but also its probability of covering failure is exponentially decreasing with $m$, which leads us to the following notion. 
\begin{definition} \label{def:expCompressibility}
The learning algorithm $\mathcal{A}$ 
is $(R,\epsilon,\delta;\left\{d_m\right\}_{m})$- exponentially compressible for some $\delta>0$, if conditions of Definition~\ref{def:compressibility} hold and 
\begin{align}
 \lim_{m \ra \infty}  \left[-\frac{1}{m} \log\left( \mathbb{P}_{(S,W)^{\otimes m}}\left(\mathcal{E}_m(\mathcal{H}_m,\epsilon;d_m)\right)\right)   \right]  \geq \log(1/\delta). \label{eq:errorExponet}
\end{align}
In other words, the error probability is asymptotically bounded by $\delta^m$.
\end{definition}
On the other hand, instead of considering \eqref{def:vectorDistanceE}, it turns out that it is sufficient to keep the difference between the \emph{average} generalization error of the compressed algorithm and the \emph{lowest} error of the original algorithm within a threshold. More precisely, we define the new distortion function:\footnote{Note that $\dpm(w^m,\hat{w}^m;s^m)\leq \dem(w^m,\hat{w}^m;s^m)$. For further discussion on this distortion function, refer to Section~\ref{sec:proofOutline}.}
\begin{align}
    \dpm(w^m,\hat{w}^m;s^m)\coloneq \min_{j \in [m]} \gen(s_j,w_j)-\frac{1}{m} \sum \nolimits_{i=1}^m \gen(s_i,\hat{w}_i). \label{def:vectorDistanceP}
\end{align}
By using this notion, our first tail bound on the generalization performance of an algorithm is stated in the following theorem, which is proved in Appendix~\ref{sec:prCompressibility}.
\begin{theorem} \label{th:compressibility}  If a learning algorithm $\mathcal{A}$ 
is $(R,\epsilon,\delta;\left\{\dpm\right\}_{m})$-exponentially compressible and if for all $w \in \mathcal{W}$, the loss $\ell(Z,w)$ is $\sigma$-subgaussian, then with probability at least $1-\delta$, we have that 
$\gen(S,W) \leq \sqrt{2\sigma^2(R+\log\left(1/\delta\right))/n}+\epsilon$.
\end{theorem}
This result shows that exponentially compressible algorithms, with small $(R,\epsilon)$, generalize well with probability $1-\delta$. Next, in our main tail bound, we will show that any arbitrary algorithm is exponentially compressible, and we will
establish a bound on its compressibility triplet $(R,\epsilon,\delta)$. 
To state this result, we need some definitions. For a given distribution $Q$ over $\mathcal{S} \times \mathcal{W}$, let \begin{align}
    d_{Q}(\hat{w};s)\coloneq \inf_{\substack{(s',w') \in \Supp(Q)}}  
      \left[\gen(s',w')\right]-\gen(s,\hat{w}).
\end{align}  
Intuitively, $\mathbb{E}_{S,\hat{W}}[d_{Q}(\hat{W};S)]$ can be seen as the limit of the distortion function $\dpm$ when $m \ra \infty$. Moreover, for a distribution $Q$ defined over $\mathcal{S} \times \mathcal{W}$, let
\begin{align}
 \mathfrak{RD}^{*} (\epsilon;Q) \coloneq & \inf \limits_{\substack{P_{\hat{W}|S}:\\\mathbb{E}\left[d_{Q}(\hat{W};S)\right]\leq \epsilon}} I(S;\hat{W})\leq \inf \limits_{\substack{P_{\hat{W}|S}:\\\mathbb{E}\left[  \gen(S,W)-\gen(S,\hat{W})\right]\leq \epsilon}} I(S;\hat{W}), \label{eq:rdStar}
\end{align}
where the infimum is over all Markov kernels $P_{\hat{W}|S}\colon \hat{\mathcal{W}}\times \mathcal{S} \to \mathbb{R}^+$ and the expectations and the mutual information are with respect to joint distributions $Q$ and $Q_{S} \times P_{\hat{W}|S}$, where $Q_{S}$ is the marginal distribution of $S$. Now, we state our main tail bound result, proved in Appendix~\ref{pr:rdCompDist}.

\begin{theorem} \label{th:rdCompDist}
 Suppose that
the algorithm $\mathcal{A}(S)=W$ 
 induces $P_{S,W}$ and 
 for all $w \in \mathcal{W}$, $\ell(Z,w)$ is $\sigma$-subgaussian. Then, for every $\epsilon \in \mathbb{R}$ and $\delta \geq 0$, with probability at least $1-\delta$,
\begin{align*}
    \gen(S,W) \leq \sqrt{\frac{2\sigma^2(R_p(\delta,\epsilon)+\log\left(1/\delta\right))}{n}}+\epsilon,~
R_p(\delta,\epsilon) \coloneq \sup \limits_{\substack{Q \colon D_{KL}(Q\|P_{S,W}) \leq \log\left(1/\delta\right)}} \mathfrak{RD}^{*} (\epsilon;Q),
\end{align*}
where the supremum is over all probability distributions $Q$ over $\mathcal{S}\times \mathcal{W}$. 
\end{theorem}
To the best of our knowledge, this is the first information-theoretic tail bound on the generalization error with the logarithmic dependence on $1/\delta$. The bound does not reduce to previous results even for $\epsilon=0$. In this case, $\mathfrak{RD}^{*} (0;Q)\leq I_Q(S;W)$ as $\hat{\mathcal{W}}=\mathcal{W}$ and $P_{\hat{W}|S}=Q_{W|S}$ are valid choices, where $I_Q(S;W)$ implies the mutual information under the distribution $Q$. Hence, as a corollary of the above theorem, with probability at least $1-\delta$, we have
\begin{align}
    \gen(S,W) \leq \sqrt{2\sigma^2(\sup \limits_{\substack{Q \colon D_{KL}(Q\|P_{S,W}) \leq \log\left(1/\delta\right)}} I_Q(S;W)+\log\left(1/\delta\right))/n}.
\end{align}

The bound in Theorem~\ref{th:rdCompDist} does not only depend on $P_{S,W}$, but on all $Q$ close to $P_{S,W}$. This is similar to the error exponent result of \cite[Theorem~1]{marton1974}. Intuitively, by considering all $Q$ satisfying $D_{KL}(Q\|P_{S,W}) \leq \log\left(1/\delta\right)$, we cover realizations of $(S^m,W^m)$ with probability at least $1-\delta^m$. In other words, to have a good generalization bound with high probability, the algorithm should be compressible under all such $Q$ that are close enough to $P_{S,W}$.

Theorem~\ref{th:rdCompDist} does not take into account any additional stochasticity of the algorithm, as considered in \citep{harutyunyan2021}. Considering such a scenario yields stronger results, presented in Appendix~\ref{sec:tailRdDis2}.

Similar to the in expectation part, by having a Lipschitzness property and using \eqref{eq:rdStar},  Theorem~\ref{th:rdCompDist} can be upper-bounded in terms of the rate-distortion functions of the hypothesis set.
\begin{corollary}[Lipschitz loss] \label{cor:lipschitz} Suppose that for a distortion function $\dw\colon \mathcal{W} \times \hat{\mathcal{W}} \to \mathbb{R}^+$ and every $z,w,\hat{w}$, $|\ell(z,w)-\ell(z,\hat{w})| \leq \mathfrak{L} \dw(w,\hat{w})$ and $\ell(Z,w)$ is $\sigma$-subgaussian. 
Then, for any $\epsilon \in \mathbb{R}^+$, the term $R_p(\delta,\epsilon)$ in Theorem~\ref{th:rdCompDist} can be upper bounded by
\begin{align}
R_p(\delta,\epsilon) \leq  \sup \limits_{\substack{Q_W \colon D_{KL}(Q_W\|P_{W}) \leq \log\left(1/\delta\right)}} \mathfrak{RD} (\epsilon/(2\mathfrak{L});Q_W,\dw),
\end{align}
where the supremum is over all possible distributions $Q_W$ over $\mathcal{W}$.
\end{corollary}

The above corollary recovers some classical results, \emph{e.g.} the bound obtained by using $\epsilon$-net coverings. This result, together with some other concrete examples are presented in Appendix~\ref{sec:LipsCor}. Furthermore, similar to Corollary~\ref{cor:dimExpec}, one can derive a dimension-based bound by using Corollary~\ref{cor:LipschExamples}.

Note that Theorem~\ref{th:rdCompDist} can be made data-dependent using ideas of \cite{negrea2020it}. 
Finally, 
%
we further extend our results in Appendix~\ref{sec:condComp}, that recovers (and improves in specific cases) the conditional mutual information based results of \citet{steinke2020reasoning,harutyunyan2021}. 

\section{Proof Outline} \label{sec:proofOutline}
Our main results are new bounds on the generalization gap. However, we also develop new techniques which are rather general and applicable to arbitrary random variables. In particular, we derive a variational representation of the tail probability in Lemma~\ref{lem:variationalTail} (Appendix~\ref{sec:tailX}) that results the following tail bound for any arbitrary random variable $X$:

\begin{theorem}\label{th:tailX} For arbitrary random variables $X\in\mathbb{R}$,  $Y\in\mathcal{Y}$ distributed according to $(X,Y)\sim\mu_{X,Y}$, with marginals $X\sim \mu_X$ and $Y \sim \mu_Y$ and for any $\delta\geq 0$ and $\epsilon,\Delta \in \mathbb{R}$, we have
\begin{align}& 
\log\mu_X\left([\Delta,\infty)\right)
\leq\label{eq:tailX}
\\& \max\left[\log(\delta),  \sup_{\nu_{X,Y}\in\mathcal{G}}\,\inf_{p_{\hat X|Y}\in\mathcal{Q}(\nu)}\,\inf_{q_{\hat X|Y},\,\lambda\geq 0}\left\{D_{KL}(p_{\hat X|Y}\nu_Y\|q_{\hat X|Y}\nu_Y)-\lambda(\Delta-\epsilon)+\log\mathbb{E}_{\mu_Y q_{\hat X|Y}}[e^{\lambda \hat X}]\right\}\right]\nonumber
\end{align}
where
$\mathcal{G}\coloneq \{\nu_{X,Y}: D_{KL}(\nu_{X,Y}\|\mu_{X,Y})\leq \log(1/\delta) \}$, $\hat X$ is a real valued random variable and  $\mathcal{Q}(\nu)$ is the set of all conditional distributions
$p_{\hat X|Y}$ such that under the joint distribution $p_{\hat X|Y}\nu_{X,Y}$ we have: $\left[\inf_{x\in \Supp(\nu_X)}x\right]-\mathbb{E}[\hat X] \leq \epsilon$, $\nu_X$ and $\nu_Y$ are marginals of $\nu_{X,Y}$ with respect to $X$ and $Y$, respectively, and the inner infimum is over all conditional distributions $q_{\hat X|Y}$.
\end{theorem}
This theorem is proved in Appendix~\ref{pr:tailX} and implies Theorem \ref{th:rdCompDist} by considering $X$ as $\gen(S,W)$. A key idea used in this paper is leveraging the block covering technique to establish tail bounds. In the following  subsection, we explain our general approach for this.

\subsection{Tail bound via information-theoretic covering}\label{secti}
Covering is a technique that allows to provide upper bounds on the tail probability or the expectation of an arbitrary random variable. The standard covering technique works as follows (see \cite[Chapter 7]{vershynin2018high}): consider a random process $(Y_t)_{t\in T}$, and the random variable
$X=\sup_{t\in T}Y_t$. An $\epsilon$-covering (or $\epsilon$-net) of the set $T$ is a finite number of points $\mathcal{N}=\{t_1, t_2, \ldots, t_k\}\subset T$ such that every point $t\in T$ is
within distortion $\epsilon$ of some point of $\mathcal{N}$. In the information theory literature, the points $t_i\in\mathcal{N}$ are called ``quantization points", and the process of mapping an arbitrary point $t\in T$ to its closest point in set $\mathcal{N}$ is called \emph{compression} because it allows one to describe each point in set $t$ by just a number from the set $\{1,2,\ldots, k\}$, \emph{i.e.,} the index of its closest point in $\mathcal{N}$. Let $\hat X_i=Y_{t_i}$ for $1\leq i\leq k$ be the value of the random process at points in the $\epsilon$-net.
The idea of covering is to approximate $X=\sup_{t\in T}Y_t$ by $\max(\hat X_1, \hat X_2, \ldots, \hat X_k)$. Since any arbitrary $t\in T$ is close to some point $t_i\in\mathcal{N}$, random variable $X$ too should be close to $\hat X_i$ for some $i$. Once we relate $\sup_{t\in T}Y_t$ to the maximum of finitely many terms $\max(\hat X_1, \hat X_2, \ldots, \hat X_k)$, one can use tools such as the union bound to study the latter maximum. 

The underlying idea of covering is fairly general. Suppose we have an arbitrary random variable $X$ that is not necessarily arising as the supremum of an underlying random process. We can still apply similar ideas if we  ``cover" $X$ by a finite collection of random variables $\{\hat X_{1}, \hat X_{2}, \ldots, \hat X_{k}\}$. In this paper, we take a similar approach but with two crucial differences: (i) instead of covering a random variable $X$, we start off by taking a vector $X^m$ of $m$ i.i.d.\ repetitions of $X$, and cover the vector $X^m$. This technique is known in the information literature as ``block-coding" and allows for a certain concentration of measure phenomenon to occur when we let $m$, the number of repetitions of  $X$ to go to infinity. Moreover, it allows to utilize classical results on compression from information theory
(ii) instead of  covering the entire space as in an $\epsilon$-net, we allow for a vanishing fraction of the space to remain uncovered. This is in line with the information-theoretic notion of covering. More precisely, we cover the subspace in which $X^m$ concentrates on. 

To see the idea of block covering in action, let $X_1, \ldots, X_m$ be $m$~i.i.d.\ repetitions of $X$. Then, 
$
    \mathbb{P}\left(X\geq \Delta\right)^m \nonumber
    = \mathbb{P}\left(\min_i X_i \geq \Delta\right)
$.
In order to relate $\min_i X_i$ to an average, we introduce the following distortion function between two sequences.\footnote{This distortion function is new and not previously used in the information theory literature to the best of our knowledge.}
\begin{definition}\label{rho-def}
Given two sequences $\vec{a}=(a_1,  \ldots, a_m)$ and $\vec{b}=(b_1,  \ldots, b_m)$, define $\rho(\vec a,\vec b)\coloneq \min_{i}a_i-\frac1m\sum_{i}b_i$. 
\end{definition}

Then, we have the following result (see Appendix \ref{pr:tailTwoTerms} for a proof):
\begin{theorem}\label{th:tailTwoTerms}
Let $X$ be an arbitrary random variable. Take  $m\in\mathbb{N}$ and let $X^m$ be its $m$ i.i.d. repetitions. Let
$\hat{X}^m(1), \ldots, \hat{X}^m(k)$ be
an arbitrary set of $k\in\mathbb{N}$ random variables produced from some arbitrary conditional distribution $Q_{\hat{X}^m(1), \cdots, \hat{X}^m(k)|X^m}$. 
Then, for any $\epsilon \in \mathbb{R}$,

\begin{align}
\mathbb{P}\left(X\geq \Delta\right)^m
&\leq \mathbb{P}\left(\exists j:\frac1m \sum_{i=1}^m\hat X_i(j)\geq \Delta- \epsilon\right)
+\mathbb{P}\left(\forall j\in[k]: \rho(X^m,{\hat X}^m(j))>  \epsilon \right)\label{eqnFirstl1}
\\&\leq\sum_{j=1}^k\mathbb{P}\left(\frac1m \sum_{i=1}^m\hat X_i(j)\geq \Delta- \epsilon\right)+\mathbb{P}\left(\forall j\in[k]: \rho(X^m,{\hat X}^m(j))>  \epsilon \right).\nonumber
\end{align}
\end{theorem}
The random vectors $\hat{X}^m(1), \hat{X}^m(2), \ldots, \hat{X}^m(k)$ represent ``quantizations" of the sequences $X^m$. The term $\mathbb{P}\left(\forall j\in[k]: \rho(X^m,{\hat X}^m(j))>  \epsilon \right)$ represents probability of excess distortion (with respect to $\rho$) when covering $X^m$ by $\hat{X}^m(1), \ldots, \hat{X}^m(k)$. The term $\mathbb{P}\left(\frac1m \sum_i\hat X_i(j)\geq \Delta- \epsilon\right)$ 
is a tail bound inequality on the quantizations points. We make this more clear by the following example.

\begin{example}
Let $k=1$, $X{\sim}\text{Bernoulli}(1/2)$, $\Delta{=}0.5$, and $\epsilon{=}0$. Then, $\mathbb{P}(X\geq \Delta)=0.5$. Let $\hat X_i=0$ with probability one. Then, $\mathbb{P}(\sum \hat X_i\geq \Delta-\epsilon)=0$. The term $\rho(X^m,\hat X^m)$ is zero if and only if $X_i=0$ for some $i$. Thus, $\mathbb{P}(\rho(X^m,\hat X^m)>\epsilon)=(1/2)^m$. Hence, we have equality for this example.
\end{example}

\section{Conclusion}
\label{sec:conc}
In this work, using the source coding literature, we developed a compressibility framework to study the generalization error of the stochastic learning algorithms. This framework allows establishing bounds on the generalization gap in terms of rate-distortion function. Further, our defined compressibility notion makes the connection between several different research approaches in studying the generalization gap, \emph{e.g.} information-theoretic and dimension-based approaches. This study opens up new directions, including: (i) making our bounds computational by applying the numerical methods to compute or bound the rate-distortion function and rate-distortion dimension, (ii) investigating the relation between our bounds and other dimensions-based bounds, by exploiting the relation between rate-distortion dimension and fractal dimensions, \emph{e.g.} correlation dimension, (iii) making the connection between our compressibility framework and PAC-Bayesian approaches \citep{mcallester1999some},\footnote{This relation have been previously established by \citet{blum2003pac} for the compressibility framework of \citep{littlestone1986relating}. The connection  of the PAC-Bayesian approaches, particularly when applied for neural networks \citep{mackay1995probable,langford2001not,dziugaite2017computing,neyshabur2018pacbayesian}, with our framework also seems promising. In these approaches the propagated error at the output of the network due to small perturbation of the weights are studied. Perturbing $W$ can be seen as letting $\hat{W}=W+N$, where $N$ is an independent noise, and the propagated error as the induced distortion. Then, one needs to properly bound $I(\hat{W};W)$.} (iv) to establish general bounds on the generalization error by combining rate-distortion theoretic results of this work and the approach of using surrogate hypothesis \citep{negrea2020defense}, and contrariwise, to use the ad-hoc approaches of the latter for our compressibility framework to derive alternative bounds, (v) and finally combining the information-theoretic covering approach, introduced in Section~\ref{secti}, with other related techniques such as chaining.

\section*{Acknowledgments}
This work is partly supported by the French National Research Agency grant ANR-16-CE23-0014 (FBIMATRIX). U\c{S}'s research is supported by the French government under management of Agence Nationale de la Recherche as part of the ``Investissements d’avenir'' program, reference ANR-19-P3IA-0001 (PRAIRIE 3IA Institute).

\bibliography{references}

\newpage

\appendix

\begin{center}   {\Large \textbf{Appendices}}\end{center}


The organization of the appendices is as follows. 
\begin{itemize}
    \item In Appendix~\ref{sec:condComp}, we introduce the notion of conditional compressibility. Using this concept, we derive several bounds on the generalization performance that recover (and for certain cases improve) some previous bounds by \cite{steinke2020reasoning,harutyunyan2021,Vapnik1998}. 
    \item In Appendix~\ref{sec:donsker}, we discuss the  Donsker-Varadhan's inequality. The relation with compressibility is shown and a variational representation of the expectation of a random variable is presented.
    \item In Appendix~\ref{sec:tailX}, the tail bound on the arbitrary random variable (Theorem~\ref{th:tailX}) is discussed. In particular, a variational representation of the tail probability is presented, which is a key lemma to derive this tail bound.
    \item In Appendix~\ref{sec:otherResults}, extensions of Theorems~\ref{th:expectationCompressible}, \ref{th:expecRdDis}, and \ref{th:rdCompDist} are stated, in addition to some concrete examples of  Corollary~\ref{cor:lipschitz}.
    \item Finally, in Appendix~\ref{sec:proofs}, proofs of our results are presented.
\end{itemize}

\paragraph{Type of a sequence} Through the appendices, we use the notion of the \emph{type} \citep{CoverTho06}. Here, we give its definition. We say that two sequences $x^m,x'^m \in \mathcal{X}^m$ have the same type if their empirical distributions are the same. The type of a sequence $x^m$ is its empirical distribution and is denoted by $\mathcal{T}(x^m)$. An $m$-type $Q_m$ refers to all sequences of length $m$ whose empirical distributions equal $Q_m$. Note that for any $x \in \mathcal{X}$, $Q_m(x)=k/m$ where $k \in \{0,1,\ldots,m\}$. For ease of notation, the type $Q_m(x)$ is simply denoted  by $Q(x)$ or $q(x)$, whenever $m$ is known from the context.


\section{Conditional Compressibility} \label{sec:condComp}
In this section, we introduce conditional compressibility, using concepts from \cite{steinke2020reasoning}. Building based on this notion, 
we derive both in expectation and tail bounds that recover and improve over some previous results. 
Theorem~\ref{th:expCond} recovers (and potentially improves over) \cite[Theorem~1.2.1]{steinke2020reasoning} and \cite[Corollary~2]{harutyunyan2021}. Corollaries~\ref{cor:VCexp} and \ref{cor:VCprob} recover the the in-expectation and tail bound results when a learning algorithm has a bounded  VC-dimension \citep{Vapnik1998}.

Through this section, assume $\Scap \in \mathcal{Z}^{n \times 2}$ be a super-dataset of length $2n$, distributed according to $P_{\Scap}=\mu^{\otimes 2n}$, containing the dataset $S=\Scap_{\Kbold}$ and a ghost dataset $\bar{S}=\Scap_{\bar{\Kbold}}$, where
$\Scap_{\Kbold}\coloneq (\Scap_{1,\Knorm_1},\ldots,\Scap_{n,\Knorm_n})$ and $\Kbold \coloneq (\Knorm_1,\ldots,\Knorm_n)$ and $\bar{\Kbold}\coloneq (\bar{\Knorm}_1,\ldots,\bar{\Knorm}_n)$ are vectors of length $n$ such that each $\Knorm_i$ takes values uniformly over $\{1,2\}$ independent of $\left\{\Knorm_j\colon j\neq i\right\}$, and $\bar{\Knorm}_i\coloneq \{1,2\} \setminus \Knorm_i$. Denote 
\begin{align}
    f(\Smin,\kbold,w) \coloneq \hat{\mathcal{L}}(\Smin_{\bar{\kbold}},w)-\hat{\mathcal{L}}(\Smin_{\kbold},w)= \frac{1}{n}\sum\nolimits_{j=1}^n (-1)^{\knorm_j}\left(\ell(\Smin_{j,1},w)-\ell(\Smin_{j,2},w)\right).
\end{align}
Let $d_m\coloneq \mathcal{W}^m \times \hat{\mathcal{W}}^m \times \mathcal{Z}^{2nm} \times \{1,2\}^{nm} \to \mathbb{R}$ be a function, measuring a distortion between $m$ realizations of $ f(\Smin,\kbold,w)$ and $ f(\Smin,\kbold,\hat{w})$. In particular, we use the following distortion functions:
\begin{align}
    \dspm(w^m,\hat{w}^m;\Smin^m,\kbold^m) & \coloneq \min \limits_{j \in [m]}f(\Smin_j,\kbold_j,w_j)-\frac{1}{m}\sum \nolimits_{i=1}^m f(\Smin_i,\kbold_i,\hat{w}_i),\\
    \dsm(w^m,\hat{w}^m;\Smin^m,\kbold^m) & \coloneq \frac{1}{m}\sum \nolimits_{i=1}^m \left(f(\Smin_i,\kbold_i,w_i)-f(\Smin_i,\kbold_i,\hat{w}_i)\right),\\
    \dsAm(w^m,\hat{w}^m;\Smin^m,\kbold^m) & \coloneq \frac{1}{m}\sum \nolimits_{i=1}^m \left(|f(\Smin_i,\kbold_i,w_i)|-|f(\Smin_i,\kbold_i,\hat{w}_i)|\right),
\end{align}
where $\kbold_i=(\knorm_{i,1},\ldots,\knorm_{i,n})$. In general when $\Smin^m=(\Smin,\ldots,\Smin)$, the distortion functions are denoted by $d_m(w^m,\hat{w}^m;\Smin,\kbold^m)$.

\begin{definition} \label{def:compressCond} The learning algorithm $\mathcal{A}(S)$ 
is $(R(\Smin),\epsilon(\Smin);\left\{d_m\right\}_{m})$-conditionally compressible for  some $R(\Smin) \in \mathbb{R}^+$ and $\epsilon(\Smin) \in \mathbb{R}$, if for any $\Smin \in \mathcal{Z}^{2 \times n}$, there exists a sequence of hypothesis books $\left\{\mathcal{H}_{m}(\Smin)\right\}_{m\in \mathbb{N}}$, $\mathcal{H}_{m}(\Smin)=\{\vc{\hat{w}}_{j}(\Smin), j \in [l_{m}(\Smin)]\} \subseteq \mathcal{\hat{W}}^m$ such that $l_{m}(\Smin) \leq  e^{mR(\Smin)}$ and 
\begin{align}
   \lim_{m \ra \infty}  \mathbb{P}_{(\Kbold,W|\Smin)^{\otimes m}}\left(\min \limits_{j \in [l_m(\Smin)]}d_m(W^m,\hat{\vc{w}}_j(\Smin);\Smin,\Kbold^m) >\epsilon(\Smin) \right)=0, \label{eq:errorZeroCond}
\end{align}
where $P_{\Kbold,W|\Smin}=\frac{1}{2^n} P_{W|\Smin_{\Kbold}}$.

The learning algorithm is $(R(\Smin),\epsilon(\Smin),\delta;\left\{d_m\right\}_{m})$- exponentially and conditionally compressible for some $\delta>0$, if in addition to above conditions, the following also  holds:
\begin{align}
 \lim_{m \ra \infty}  \left[-\frac{1}{m} \log\left(  \mathbb{P}_{(\Kbold,W|\Smin)^{\otimes m}}\left(\min \limits_{j \in [l_m(\Smin)]}d_m(W^m,\hat{\vc{w}}_j(\Smin);\Smin,\Kbold^m) >\epsilon(\Smin) \right)\right)   \right]  \geq \log(1/\delta). \label{eq:errorExponetCond}
\end{align}
In other words, asymptotically the error probability is bounded by  $\delta^m$. 

\end{definition}

Similar to the unconditional part, we state in expectation and tail bounds.
\subsection{Bounds on the expected value of the generalization gap}
The first theorem is a bound on the expectation of the generalization performance of the conditionally compressible algorithms.
\begin{theorem} \label{th:expectationCompressibleCond}  Consider  a learning algorithm $\mathcal{A}(S)$ 
and a bounded loss function $\ell(z,w) \in [0,1]$. 

\begin{itemize}
    \item[i.] If $\mathcal{A}(S)$ is $(R(\Smin),\epsilon(\Smin);\left\{\dsm\right\}_{m})$-conditionally compressible, then 
    \begin{align*}
    \mathbb{E}\left[\gen(S,W)\right] \leq \mathbb{E}_{\Scap} \left[\sqrt{\frac{2 R(\Scap)}{n}}+\epsilon(\Scap)\right].
\end{align*}
\item[ii.] If $\mathcal{A}(S)$ is $(R(\Smin),\epsilon(\Smin);\left\{|\dsm|\right\}_{m})$-conditionally compressible, then 
    \begin{align*}
    \left|\mathbb{E}\left[\gen(S,W)\right]\right| \leq \mathbb{E}_{\Scap} \left[\sqrt{\frac{2 R(\Scap)}{n}}+\epsilon(\Scap)\right].
\end{align*}
\item[ii.] If $\mathcal{A}(S)$ is $(R(\Smin),\epsilon(\Smin);\left\{\dsAm\right\}_{m})$-conditionally compressible, then 
    \begin{align*}
    \mathbb{E}\left[\left|\gen(S,W)\right|\right] \leq \mathbb{E}_{\Scap} \left[\sqrt{\frac{2 (R(\Scap)+\log(2))}{n}}+\epsilon(\Scap)\right].
\end{align*}
\end{itemize}
\end{theorem}
This theorem is proved in Appendix~\ref{pr:expectationCompressibleCond}. We use this result to derive a bound on the generalization gap of an arbitrary learning algorithm, in the next theorem. This theorem can be derived from \cite[Theorem~1.2]{steinke2020reasoning} in the same manner as we have proved Theorem~\ref{th:expecRdDis} using \cite[Theorems~1,4]{xu2017information}. It can be alternatively derived using Theorem~\ref{th:expectationCompressibleCond} and by bounding the conditional compressibility parameters of an arbitrary learning algorithm, similar to the proof of  Theorem~\ref{th:compressibilityBound}. We omit the proof, as it is similar to the proofs of Theorems~\ref{th:compressibilityBound} and \ref{th:expecRdDis}.
\begin{theorem} \label{th:expCond}  Suppose the algorithm $\mathcal{A}(S)=W$ induces $P_{S,W}$ and the loss function $\ell(z,w)$ is bounded in the range $[0,1]$. Consider any auxiliary random variable $U$\footnote{Here, $U$ represents the stochasticity of the algorithm. Note that $U$ being a constant is always a valid choice. For further discussions, refer to Appendix~\ref{sec:tailRdDis2}.} defined by the conditional distribution $P_{U|S,W}$ and satisfying $P_{U,S,W}=P_U P_S P_{W|U,S}$\footnote{Note that $P_{U,\Scap,\Kbold,W}= P_{\Scap} P_U P_{W|U,\Scap_{\Kbold}}$.}.  Then, for any $\epsilon \in \mathbb{R}$
\begin{itemize}[leftmargin=*]
    \item[i.]\begin{align}
    \left|\mathbb{E}_{S,W}\left[\gen(S,W)\right]\right| \leq & \mathbb{E}_{\Scap,U} \left[\sqrt{\frac{2 R_{E,\Scap,U}(\epsilon)}{n}}+\epsilon \right],\nonumber\\
    \mathbb{E}_{S,W}\left[\gen(S,W)\right] \leq & \mathbb{E}_{\Scap,U} \left[\sqrt{\frac{2 R'_{E,\Scap,U}(\epsilon)}{n}}+\epsilon \right],\nonumber\\
    \mathbb{E}_{S,W}\left[\left|\gen(S,W)\right|\right] \leq &\mathbb{E}_{\Scap,U} \left[ \sqrt{\frac{2(R''_{E,\Scap,U}(\epsilon)+\log(2))}{n}}+\epsilon\right],
\end{align}
where 
\begin{align}
    R_{E,\Smin,u}(\epsilon) = & \inf \limits_{\substack{P_{\hat{W}|\Smin_{\Kbold},u}}} I(\Kbold;\hat{W}|\Smin,u),\quad \text{such that} \quad  \left|\mathbb{E} \left[f(\Smin,\Kbold,W)-f(\Smin,\Kbold,\hat{W})\right]\right|\leq \epsilon,\nonumber\\
    R'_{E,\Smin,u}(\epsilon) =& \inf \limits_{\substack{P_{\hat{W}|\Smin_{\Kbold},u}}} I(\Kbold;\hat{W}|\Smin,u),\quad \text{such that} \quad  \mathbb{E} \left[f(\Smin,\Kbold,W)-f(\Smin,\Kbold,\hat{W})\right]\leq \epsilon, \nonumber\\
    R''_{E,\Smin,u}(\epsilon) = & \inf \limits_{\substack{P_{\hat{W}|\Smin_{\Kbold},u}}} I(\Kbold;\hat{W}|\Smin,u),\quad \text{such that} \quad \mathbb{E} \left[|f(\Smin,\Kbold,W)|-|f(\Smin,\Kbold,\hat{W})|\right]\leq \epsilon. \label{eq:expecRECond}
\end{align}
The expectations  are with respect to $P_{\Kbold}P_{W|\Smin_{\Kbold},u}$ and $P_{\Kbold} P_{\hat{W}|\Smin_{\Kbold},u}$.
\item[ii.]  \begin{align}
    \left|\mathbb{E}_{S,W}\left[\gen(S,W)\right]\right| &\leq  \frac{1}{n}\sum \limits_{i=1}^n \mathbb{E}_{\Scap,U} \left[\sqrt{2 R_{E,\Scap,U,i}(\epsilon)}+\epsilon \right],\nonumber\\
    \mathbb{E}_{S,W}\left[\gen(S,W)\right] &\leq  \frac{1}{n}\sum \limits_{i=1}^n \mathbb{E}_{\Scap,U} \left[\sqrt{2 R'_{E,\Scap,U,i}(\epsilon)}+\epsilon \right],\nonumber \\
    \mathbb{E}_{S,W}\left[\left|\gen(S,W)\right|\right] & \leq \frac{1}{n}\sum \limits_{i=1}^n \mathbb{E}_{\Scap,U} \left[ \sqrt{2(R''_{E,\Scap,U,i}(\epsilon)+\log(2))}+\epsilon\right],
\end{align}
where 
\begin{align}
    R_{E,\Smin,u,i}(\epsilon) =& \inf \limits_{\substack{P_{\hat{W}|\Smin_{\Knorm_i},u} }} I(\Knorm_i;\hat{W}|
    \Smin_i,u),\quad \text{such that} \quad \left|\mathbb{E} \left[f(\Smin_i,\Knorm_i,W)-f(\Smin_i,\Knorm_i,\hat{W})\right]\right|\leq \epsilon, \nonumber \\
    R'_{E,\Smin,u,i}(\epsilon) =& \inf \limits_{\substack{P_{\hat{W}|\Smin_{\Knorm_i},u}}} I(\Knorm_i;\hat{W}|
    \Smin_i,u),\quad \text{such that} \quad \mathbb{E} \left[f(\Smin_i,\Knorm_i,W)-f(\Smin_i,\Knorm_i,\hat{W})\right] \leq \epsilon, \nonumber\\ 
    R''_{E,\Smin,u,i}(\epsilon) = &\inf \limits_{\substack{P_{\hat{W}|\Smin_{\Knorm_i},u}}} I(\Knorm_i;\hat{W}|\Smin_i,u),\quad \text{such that} \quad \mathbb{E} \left[|f(\Smin_i,\Knorm_i,W)|-|f(\Smin_i,\Knorm_i,\hat{W})|\right]\leq \epsilon, \label{eq:expecRECondN}
\end{align}
where $  f(\Smin_i,\Knorm_i,w) \coloneq (-1)^{\knorm_i}\left(\ell(\Smin_{i,1},w)-\ell(\Smin_{i,2},w)\right)$ and the expectations  are with respect to $P_{\Knorm_i}P_{W|\Smin_{i,\knorm_i},u}$ and $P_{\Knorm_i} P_{\hat{W}|\Smin_{i,\Knorm_i},u}$.
\end{itemize}
\end{theorem}
The above bound trivially recovers \cite[Theorem~1.2.1]{steinke2020reasoning} and \cite[Corollary~2]{harutyunyan2021} by letting $U=\text{Constant}$, $\epsilon=0$, and $\hat{W}=W$.

Next, we show that we can recover the bound in terms of VC-dimension using the above result.
\begin{corollary} \label{cor:VCexp}
If a learning algorithm has the VC-dimension $d$ and the loss function $\ell(z,w) \in [0,1]$, then
\begin{align*}
    \left|\mathbb{E}\left[\gen(S,W)\right]\right| &\leq \sqrt{\frac{2 d\log(2e n/d)}{n}},~~~~ \mathbb{E}\left[\left|\gen(S,W)\right|\right] \leq \sqrt{\frac{2( d\log(2en/d)+\log(2))}{n}}.
\end{align*}
\end{corollary}
The corollary is proved in Appendix~\ref{pr:VCexp}.

\subsection{Tail bounds on the generalization gap}
In this section, we propose tail bounds on the generalization performance using exponentially and conditionally compressibility.

\begin{theorem} \label{th:compressibilityCond} If the learning algorithm $\mathcal{A}(S)$ 
is $(R(\Smin),\epsilon(\Smin),\delta/2;\left\{d_{\Smin,p}\right\}_{m})$- exponentially and conditionally compressible, then with probability at least $1-\delta$ 
for the bounded loss function $\ell(z,w) \in [0,1]$, 
\begin{align*}
    \gen(S,W) \leq \sup \limits_{\Smin \in \mathcal{Z}^{2n}}\left[\sqrt{\frac{2(R(\Smin)+\log\left(2/\delta\right))}{n}}+\epsilon(\Smin)\right]+\sqrt{\frac{\log(2/\delta)}{n}}.
\end{align*}
\end{theorem}
The theorem is proved in Appendix~\ref{pr:compressibilityCond}.

Now, we establish a tail bound on the generalization performance of the arbitrary learning algorithm. For a given distribution $Q_{\kbold,W}$ over $\{1,2\}^n \times \mathcal{W}$, let \begin{align}
    d_{Q_{\Kbold,W}}(\hat{w};\Smin,\kbold)\coloneq \inf_{\substack{(\kbold',w')\in \Supp(Q_{\Kbold,W})}}        \left[f(\Smin,\kbold',w')\right]-f(\Smin,\kbold,\hat{w}).
\end{align}  
Moreover, for a set $\mathcal{U}$\footnote{As mentioned before, $U$ represents the stochasticity of the algorithm. For further discussions, refer to Appendix~\ref{sec:tailRdDis2}.} and a distribution $Q$ defined over $\{1,2\}^{n} \times \mathcal{W} \times \mathcal{U}$,  define 
\begin{align}
 \mathfrak{RD}^{\dagger} (\epsilon;Q|
\Smin) \coloneq & \inf \limits_{\substack{P_{\hat{W}|\Smin_{\Kbold},U}:\\\mathbb{E}\left[  d_{Q_{\Kbold,W}}(\hat{W};\Smin,\Kbold)\right]\leq \epsilon}} I(\Kbold;\hat{W}|U)\nonumber\\\leq&\inf \limits_{\substack{P_{\hat{W}|\Smin_{\Kbold},U}:\\\mathbb{E}\left[  f(\Smin,\Kbold,W)-f(\Smin,\Kbold,\hat{W})\right]\leq \epsilon}} I(\Kbold;\hat{W}|U)\\=&\inf \limits_{\substack{P_{\hat{W}|\Smin_{\Kbold},U}:\\\mathbb{E}\left[\gen(\Smin_{\Kbold},W)-\gen(\Smin_{\Kbold},\hat{W}) \right]\leq \epsilon}} I(\Kbold;\hat{W}|U),
\end{align}
where $Q_{\Kbold,W}$ is the marginal distribution of $(\Kbold,W)$, the infimum is over all conditional probability distributions (Markov kernels) $P_{\hat{W}|S,U}: \hat{\mathcal{W}}\times \mathcal{S} \times \mathcal{U} \to \mathbb{R}^+$, the expectation and the mutual information are with respect to joint distributions $Q$ and $Q_{U,\Kbold} \times P_{\hat{W}|\Smin_{\Kbold},U}$, where $Q_{U,\Kbold}$ is the marginal distribution of $(U,\Kbold)$ . Then, we have the below tail bound, proved in Appendix~\ref{pr:rdCompDistCond}.

\begin{theorem} \label{th:rdCompDistCond}
 Suppose the algorithm $\mathcal{A}(S)=W$ 
 induces $P_{S,W}$ and $\ell(z,w)$ is  bounded in the range $[0,1]$. Consider any auxiliary random variable $U$ defined by the conditional distribution $P_{U|S,W}$ and satisfying $P_{U,S,W}=P_U P_S P_{W|U,S}$\footnote{Note that $P_{\Scap,\Kbold,W,U}= P_{\Scap} P_{\Kbold} P_U P_{W|U,\Scap_{\Kbold}}=\frac{1}{2^n}\mu^{\otimes 2n}P_U P_{W|U,\Scap_{\Kbold}}$.}. Then, for any values of $\{\epsilon(\Smin)\}_{\Smin}$ and $\delta \geq 0$, with probability at least $1-\delta$ 
\begin{align}
   \gen(S,W) \leq \sup \limits_{\Smin}\left[\sqrt{\frac{2(R(\Smin,\delta,\epsilon)+\log\left(2/\delta\right))}{n}}+\epsilon(\Smin)\right]+\sqrt{\frac{\log(2/\delta)}{n}}. \label{eq:rdTailCond}
\end{align}
where
\begin{align}
R(\Smin,\delta,\epsilon) \coloneq \sup \limits_{\substack{Q \colon D_{KL}(Q\|P_{\Kbold,W,U|\Smin}) \leq \log\left(2/\delta\right)}} \mathfrak{RD}^{\dagger} (\epsilon(\Smin);Q|\Smin),
\end{align}
where the supremum is over all possible distributions $Q$ over $\{1,2\}^n\times \mathcal{W} \times \mathcal{U}$. 
\end{theorem}

\begin{remark} By considering the exponentially and conditionally compressibility with respect to $\mathbb{P}_{(\Kbold,W,\Scap)^{\otimes m}}$ rather than $\mathbb{P}_{(\Kbold,W|\Smin)^{\otimes m}}$ in \eqref{eq:errorExponetCond}, the following result also can be achieved with the assumptions of Theorem~\ref{th:rdCompDistCond}. For any $\epsilon$ and $\delta \geq 0$, with probability at least $1-\delta$ 
\begin{align}
   \gen(S,W) \leq \sqrt{\frac{2(R(\delta,\epsilon)+\log\left(2/\delta\right))}{n}}+\epsilon+\sqrt{\frac{\log(2/\delta)}{n}}.
\end{align}
where
\begin{align}
R(\delta,\epsilon) \coloneq \sup \limits_{\substack{Q \colon D_{KL}(Q\|P_{K,W,\Scap,U}) \leq \log\left(2/\delta\right)}} \mathbb{E}_{\Scap \sim Q_{\Scap}}\left[\mathfrak{RD}^{\dagger} (\epsilon;Q_{\Kbold,W,U|\Scap}|\Scap)\right],
\end{align}
where the supremum is over all possible distributions $Q$ over $\{1,2\}^n \times \mathcal{W} \times \mathcal{Z}^{2n} \times \mathcal{U}$, $Q_{\Kbold,W,U|\Scap}$ is the conditional distribution of $(\Kbold,W,U)$ given $\Scap$, and $Q_{\Scap}$ is the marginal distribution of $\Scap$. \end{remark}

Finally, we use Theorem~\ref{th:rdCompDistCond} to recover the generalization bound for the algorithms having a bounded VC-dimension \cite{Vapnik1998}.

\begin{corollary} \label{cor:VCprob} If a learning algorithm has VC-dimension $d$ and the loss function $\ell(z,w) \in [0,1]$, then with probability at least $1-\delta$ 
\begin{align*}
   \gen(S,W) &\leq \sqrt{\frac{2 (d\log(2e n/d)+\log(2/\delta))}{n}}+\sqrt{\frac{\log(2/\delta)}{n}}.
\end{align*}
\end{corollary}
The corollary is proved in Appendix~\ref{pr:VCprob}.


\section{On the Donsker-Varadhan's Inequality} \label{sec:donsker}
The Donsker-Varadhan's identity implies that for arbitrary distributions $p(x)$ and $q(x)$ on a  set $\mathcal{X}$ and for any arbitrary function $\Phi:\mathcal{X}\to \mathbb{R}$ we have
\begin{align}
D_{KL}(q\|p)\geq \mathbb{E}_{X\sim q}[\Phi(X)]-\log\left(\mathbb{E}_{X\sim p}\left[e^{\Phi(X)}\right]\right). \label{eq:DonskerVardhan}    
\end{align}
In this appendix, we first  show that this inequality can be proved using a compressibility approach for a finite set $\mathcal{X}$.  Then, we also derive a lemma based on \eqref{eq:DonskerVardhan}  that is used to derive a tail bound on an arbitrary random variable.

\subsection{Donsker-Varadhan's inequality via compression} \label{sec:donskerCompr}
Take some arbitrary function $\Phi:\mathcal{X}\to \mathbb{R}$.
Generate $2^{mR}$ sequences \[X^m(1), X^m(2), \ldots, X^m(2^{mR}),\] 
where $X^m(i)=(X_1(i),\ldots,X_m(i))$, 
in an i.i.d.\ fashion from $p(x)$, \emph{i.e.} each $X_j(i)\sim p$ for $i \in [2^{mR}], j \in [m]$, independent of other instances. Consider the expression
\[\mathbb{E}\left[\max_{k}\sum_{i=1}^m\Phi(X_i(k))\right]
.\] 
On the one hand,
\[e^{\mathbb{E}\left[\max_{k}\sum_{i}\Phi(X_i(k))\right]}\leq
\mathbb{E}\left[e^{\max_{k}\sum_{i}\Phi(X_i(k))}\right]\leq
\mathbb{E}\left[\sum_ke^{\sum_{i}\Phi(X_i(k))}\right]
=2^{mR}\left(\mathbb{E}_{X\sim p}\left[e^{\Phi(X)}\right]\right)^m.
\]
Therefore,
\begin{align}
    \mathbb{E}\left[\max_{k}\sum_{i}\Phi(X_i(k))\right]\leq
mR+m\log\left(\mathbb{E}_{X\sim p}\left[e^{\Phi(X)}\right]\right).\label{eq:dons1}
\end{align}
On the other hand, let $\gamma_m$ be the probability that at least one of the sequences $X^m(k)$ for some $k$ will have type $q(x)$. It is known that (for example by using \citet[Theorem~11.1.4]{CoverTho06})  $\gamma_m\rightarrow 1$ as $m$ tends to infinity if $R>D_{KL}(q\|p)$.
Under the event that the sequence $X^m(k)$ has type $q(x)$, $\sum_{i}\Phi(X_i(k))$ equals $m\mathbb{E}_{X\sim q}[\Phi(X)]$. Thus, 
\begin{align}\mathbb{E}\left[\max_{k}\sum_{i}\Phi(X_i(k))\right]\geq 
m\gamma_m\mathbb{E}_{X\sim q}[\Phi(X)]+m(1-\gamma_m)\min_{x}\Phi(x).\label{eq:dons2}
\end{align}
From \eqref{eq:dons1} and \eqref{eq:dons2} and by letting $m$ tend to infinity, we obtain
\[
D_{KL}(q\|p)+\log\left(\mathbb{E}_{X\sim p}\left[e^{\Phi(X)}\right]\right)\geq \mathbb{E}_{X\sim q}[\Phi(X)].\]
This yields the desired inequality.
\subsection{Variational representation of \texorpdfstring{$\mathbb{E}[X]$}{E[X]}} 
\label{sec:donskerEx}
In this subsection, we state a variational lemma on $\mathbb{E}[X]$, used in proof of Theorem~\ref{th:tailX}. The lemma is proved in  Appendix~\ref{pr:donskerEx}, by using \eqref{eq:DonskerVardhan}.
\begin{lemma}
For every distribution $\nu$, we have
\begin{align} \mathbb{E}_{\nu}[ X]=\frac{1}{\lambda}\inf_{\mu}\left[D_{KL}(\nu\|\mu)+\log\mathbb{E}_{\mu}\left[e^{\lambda  X}\right]\right].
\end{align}\label{lemmaag2}
\end{lemma}


\section{Tail Bound on an Arbitrary Random Variable} \label{sec:tailX}
The key to the proof of the tail bound in Theorem~\ref{th:tailX} is a variational representation of the tail probability, stated in the next lemma.

\begin{lemma}\label{lem:variationalTail}
Let $\epsilon$ be an arbitrary real number.
For any arbitrary distribution $\nu_X$ on $\mathbb{R}$, let $\mathcal{P}(\nu_X)$ denote the set of distributions $p_{\hat X}$ on $\mathbb{R}$ for which
\[\left[\inf_{x\in \Supp(\nu_X)}x\right]-\mathbb{E}\left[\hat {X}\right]\leq \epsilon.
\]
Let $X\sim\mu_X$ where $\mu_X$ is an arbitrary distribution on the sample space $\mathcal{X}\subseteq \mathbb{R}$. Then, for any $\Delta \in \mathbb{R}$ we have\footnote{For $a\in \mathbb{R}$,  $[a]_+ \coloneq \max(0,a)$.}
\begin{align*}\log\mathbb{P}_{X\sim \mu_X}\left(X\geq \Delta\right)
&= \sup_{\nu_{X}\ll\mu_X}~~\inf_{p_{\hat X}\in\mathcal{P}(\nu_X), ~\lambda\geq 0}\left\{-D_{KL}(\nu_{X}\|\mu_{X})-\lambda\bigg[\Delta-\epsilon- \mathbb{E}_{p}[\hat X]\bigg]_+\right\}.
\end{align*}
\end{lemma}
We give two proofs for this lemma. The first proof is provided in Appendix~\ref{pr:variationalTail}. We give also an alternative proof (in the inequality form) when $\mathcal{X}$ and $\mathcal{S}$ are finite sets. This proof illustrates the connections between the tail bound and compression. To this end, consider the distortion function $\rho$ defined in Definition~\ref{rho-def}. Note that $\dpm$ can be expressed in terms of this distortion function, \emph{i.e.} $\dpm(w^m,\hat{w}^m;s^m)=\rho\left(\left\{\gen(s_i,w_i)\right\}_{i\in [m]},\left\{\gen(s_i,\hat{w}_i)\right\}_{i\in [m]}\right)$. To establish the tail bound, first we upper bound it in terms of the tail of some quantizations of $X^m$ and the probability of covering $X^m$ by this quanitzation points. This is exactly the bound established in Theorem~\ref{th:tailTwoTerms}. Indeed, Theorem~\ref{th:tailTwoTerms} shows the connection between the tail bound and compression. Note that Theorem~\ref{th:tailTwoTerms} also holds if the conditions $\geq \Delta-\epsilon$ and $>\epsilon$ are replaced by conditions $> \Delta-\epsilon$ and $\geq \epsilon$ respectively in \eqref{eqnFirstl1}. The rest of proof applies some information-theoretic techniques, as detailed in Appendix~\ref{pr:variationalTail2}.


\section{Other Results} \label{sec:otherResults}
In this section, we state some further obtained results.
\subsection{Extension of the in expectation bound} \label{sec:expecRdDis2}
In a similar manner as \eqref{def:vectorDistanceE}, let $\deAm(w^m,\hat{w}^m;s^m)\coloneq \frac{1}{m} \sum \nolimits_{i=1}^m  \left(|\gen(s_i,w_i)|-|\gen(s_i,\hat{w}_i)|\right)$. Here, we state the extended version of Theorem~\ref{th:expectationCompressible}.

\begin{reptheorem}{th:expectationCompressible}  Consider a learning algorithm $\mathcal{A}(S)$ 
and suppose that $\mathbb{E}_{S,W}[|\gen(S,W)|]<\infty$ and for all $w \in \mathcal{W}$, $\ell(Z,w)$ is $\sigma$-subgaussian.
\begin{itemize}[leftmargin=*]
    \item[i.] If $\mathcal{A}(S)$ is $(R,\epsilon;\left\{\dem\right\}_{m})$-compressible, then $\mathbb{E}\left[\gen(S,W)\right] \le\sqrt{2\sigma^2R/n}+\epsilon$.
\item[ii.] If $\mathcal{A}(S)$ is $(R,\epsilon;\left\{|\dem|\right\}_{m})$-compressible, then $\left|\mathbb{E}\left[\gen(S,W)\right]\right| \le\sqrt{2\sigma^2 R/n}+\epsilon$.
\item[iii.] If $\mathcal{A}(S)$ is $(R,\epsilon;\left\{\deAm\right\}_{m})$-compressible, then $\mathbb{E}\left[\,|\gen(S,W)|\,\right] \le\sqrt{2\sigma^2(R+\log(2))/n}+\epsilon$.
\end{itemize}
All above expectations are with respect to $P_{S,W}$.
\end{reptheorem}
Next, we state the extended version of Theorem~\ref{th:expecRdDis}.
\begin{reptheorem}{th:expecRdDis} Assume that the algorithm $\mathcal{A}(S)=W$ 
induces $P_{S,W}$ and for all $w \in \mathcal{W}$, $\ell(Z,w)$ is $\sigma$-subgaussian. Then, for any $\epsilon \in \mathbb{R}$,
\begin{align}       \left|\mathbb{E}\left[\gen(S,W)\right]\right| & \leq  \sqrt{\frac{2\sigma^2R_E(\epsilon)}{n}}+\epsilon,\nonumber \\
\mathbb{E}\left[\gen(S,W)\right] &\leq \sqrt{\frac{2\sigma^2 R'_E(\epsilon)}{n}}+\epsilon,\nonumber\\  \mathbb{E}\left[\,|\gen(S,W)|\,\right] &\leq \sqrt{\frac{2\sigma^2(R''_E(\epsilon)+\log(2))}{n}}+\epsilon, \label{eq:expecRDis} \end{align}
where 
\begin{align}
     R_E(\epsilon) &= \inf \limits_{P_{\hat{W}|S}} I(S;\hat{W}),\quad \text{such that} \quad \left|\mathbb{E} \left[\gen(S,W)-\gen(S,\hat{W})\right]\right|\leq \epsilon,\nonumber\\
     R'_E(\epsilon) &= \inf \limits_{P_{\hat{W}|S}} I(S;\hat{W}),\quad \text{such that} \quad  \mathbb{E} \left[\gen(S,W)-\gen(S,\hat{W})\right]\leq \epsilon, \nonumber\\
    R''_{E}(\epsilon) &= \inf \limits_{P_{\hat{W}|S}} I(S;\hat{W}),\quad \text{such that} \quad \mathbb{E} \left[|\gen(S,W)|-|\gen(S,\hat{W})|\right]\leq \epsilon. \label{eq:expecRE}
\end{align} 
All expectations in above are with respect to $P_{S,W}$ and $P_{S} \times P_{\hat{W}|S}$.
\end{reptheorem}
This theorem can be trivially extended to the case where we have access to an internal randomness $U$ of the algorithm, as defined in Appendix~\ref{sec:tailRdDis2}. For example, by using $\mathbb{E}\left[\gen(S,W)\right]=\mathbb{E}_U\mathbb{E}_{S,W|U}\left[\gen(S,W)\right]$, it can be  shown that $\mathbb{E}\left[\gen(S,W)\right] \leq \mathbb{E}_U [\sqrt{2\sigma^2 R_{E,U}/n}]+\epsilon$, where $R_{E,u}(\epsilon) \coloneq \inf  I(S;\hat{W}|U=u)$, in which the infimum is over all Markov kernels $P_{\hat{W}|S,u}$ such that $\mathbb{E} \left[\gen(S,W)-\gen(S,\hat{W})\right]\leq \epsilon$.

In the following, Theorem~\ref{th:expecRdDis} is extended similarly to \cite{Bu2020}. The proof trivially follows from the relation $\mathbb{E}\left[\gen(S,W)\right]=\frac{1}{n}\sum\limits_{i=1}^n\mathbb{E}\left[\gen(\{Z_i\},W)\right]$ and Theorem~\ref{th:expecRdDis}, where $\mathbb{E}\left[\gen(\{z_i\},w)\right]=\mathcal{L}(w)-\ell(z_i,w)$.
\begin{theorem} \label{th:expecRdDis2} Suppose the algorithm $\mathcal{A}(S)=W$ induces $P_{S,W}$ 
and for all $w \in \mathcal{W}$, $\ell(Z,w)$ is $\sigma$-subgaussian. Then, for any $\epsilon \in \mathbb{R}$,
\begin{align}
    \left|\mathbb{E}\left[\gen(S,W)\right]\right|  &\leq \frac{1}{n}\sum\limits_{i=1}^n \left[\sqrt{2\sigma^2 R_{E,i}(\epsilon)}\right]+\epsilon,\\
     \mathbb{E}\left[\gen(S,W)\right] &\leq \frac{1}{n}\sum\limits_{i=1}^n\left[\sqrt{2\sigma^2 R'_{E,i}(\epsilon)}\right]+\epsilon,\\
    \mathbb{E}\left[\,|\gen(S,W)|\,\right] &\leq \frac{1}{n}\sum\limits_{i=1}^n\left[\sqrt{2\sigma^2(R''_{E,i}(\epsilon)+\log(2))}\right]+\epsilon,
\end{align}
where the expectation is with respect to $P_{S,W}$ and
\begin{align}
     R_{E,i}(\epsilon) =& \inf \limits_{P_{\hat{W}|Z_i}} I(Z_i;\hat{W}),\quad \text{such that} \quad \left|\mathbb{E} \left[\gen(\{Z_i\},W)-\gen(\{Z_i\},\hat{W})\right]\right|\leq \epsilon,\\
    R'_{E,i}(\epsilon) =& \inf \limits_{P_{\hat{W}|Z_i}} I(Z_i;\hat{W}),\quad \text{such that} \quad \mathbb{E} \left[\gen(\{Z_i\},W)-\gen(\{Z_i\},\hat{W})\right]\leq \epsilon,\\
    R''_{E,i}(\epsilon) =& \inf \limits_{P_{\hat{W}|Z_i}} I(Z_i;\hat{W}),\quad \text{such that} \quad \mathbb{E} \left[|\gen(\{Z_i\},W)|-|\gen(\{Z_i\},\hat{W})|\right]\leq \epsilon, \label{eq:expecREN}
\end{align}
where $\mathbb{E}\left[\gen(\{z_i\},w)\right]=\mathcal{L}(w)-\ell(z_i,w)$ and the expectations are with respect to $P_{Z_i,W}$ and $P_{Z_i} \times P_{\hat{W}|Z_i}$. 
\end{theorem}

Letting $\epsilon=0$ and $\hat{W}=W$, this theorem recovers (and potentially improves over) \cite[Proposition~1]{Bu2020}.

\subsection{Extension of the tail bound}
\label{sec:tailRdDis2}
It has been already shown by \cite{harutyunyan2021} that taking into account the stochasticity of the algorithm could yield tighter bounds on the expectation of the generalization gap. Here, we apply a similar idea for the tail bound. To this end, we represent partial or full stochasticity of the algorithm which is independent of the dataset by $U \in \mathcal{U}$. This means that  the hypothesis is chosen according to $P_{W|S,U}$ (deterministically or randomly). Having this stochasticity available, we can make our compression more efficient, by letting the hypothesis books in Definition~\ref{def:compressibility} depend on $U$ as well, \emph{i.e.} for each arbitrary distribution $Q$ defined over $\mathcal{U}$, we choose a sequence of hypothesis books  $\mathcal{H}_{m}(Q)=\{\vc{\hat{w}}_j(Q), j \in [l_m(Q)]\} \subseteq \mathcal{\hat{W}}^m$, such that $l_m(Q) \leq  e^{mR(Q)}$ and
\begin{align}
 \lim_{m \ra \infty}  -\frac{1}{m} \log\left( \mathbb{P}_{(U,S,W)^{\otimes m}}\left(\min \limits_{j \in [l_m(\hat{P}_{U^m})]} \dpm\left(W^m,\hat{\vc{w}}_j(\hat{P}_{U^m});S^m\right)>\epsilon\right)\right)    \geq \log(1/\delta),
\end{align}
where $\hat{P}_{U^m}$ is the empirical distribution of $U^m$. Then, it can be shown that $R$ in Theorem~\ref{th:compressibility} can be replaced by $\sup_Q R(Q)$, where the supremum is over all $Q$ such that $D_{KL}(Q\|P_{U})\leq \log(1/\delta)$.

In order to define the extended version of Theorem~\ref{th:rdCompDist}, we need to define an extended definition of $ \mathfrak{RD}^{*} (\epsilon;Q)$, that takes $U$ also into account. For a distribution $Q$ defined over $\mathcal{S} \times \mathcal{W} \times \mathcal{U}$, let
\begin{align}
 \mathfrak{RD}^{*} (\epsilon;Q) \coloneq & \inf \limits_{\substack{P_{\hat{W}|S,U}:\\\mathbb{E}\left[d_{Q_{S,W}}(\hat{W};S)\right]\leq \epsilon}} I(S;\hat{W}|U)\leq \inf \limits_{\substack{P_{\hat{W}|S,U}:\\\mathbb{E}\left[  \gen(S,W)-\gen(S,\hat{W})\right]\leq \epsilon}} I(S;\hat{W}|U), \label{eq:rdStar2}
\end{align}
where $Q_{S,W}$ is the marginal distribution of $(S,W)$, the infimum is over all Markov kernels  $P_{\hat{W}|S,U}: \hat{\mathcal{W}}\times \mathcal{S} \times \mathcal{U} \to \mathbb{R}^+$, the expectation and the mutual information are with respect to joint distributions $Q$ and $Q_{S,U} \times P_{\hat{W}|S,U}$, where $Q_{S,U}$ is the marginal distribution of $(S,U)$. Note that letting $U$ being a constant, \eqref{eq:rdStar2} will be reduced to \eqref{eq:rdStar}. Now, we state an extended version of Theorem~\ref{th:rdCompDist}, proved in Appendix~\ref{pr:rdCompDist}. 

\begin{reptheorem}{th:rdCompDist}
 Suppose that the algorithm $\mathcal{A}(S)=W$ 
 induces $P_{S,W}$ and for all $w \in \mathcal{W}$, $\ell(Z,w)$ is $\sigma$-subgaussian. Consider any auxiliary random variable $U$ defined by  $P_{U|S,W}$ and satisfying $P_{U,S,W}=P_U P_S P_{W|U,S}$. Then, for every $\epsilon \in \mathbb{R}$ and $\delta \geq 0$, with probability at least $1-\delta$,
\begin{align*}
    \gen(S,W) \leq \sqrt{\frac{2\sigma^2(R_p(\delta,\epsilon)+\log\left(1/\delta\right))}{n}}+\epsilon,~
R_p(\delta,\epsilon) \coloneq \sup \limits_{\substack{Q \colon D_{KL}(Q\|P_{S,W,U}) \leq \log\left(1/\delta\right)}} \mathfrak{RD}^{*} (\epsilon;Q),
\end{align*}
where the supremum is over all possible distributions $Q$ over $\mathcal{S}\times \mathcal{W}\times \mathcal{U}$. 
\end{reptheorem}
Note that the above bound holds for any $U$ that satisfies the assumptions of the theorem and $U$ being a constant is always valid choice. By the choice of $U=\text{Constant}$, this extended version becomes the same as the original one, stated in Section~\ref{sec:tailMain}.

As a special case, when $S$ and $W$ are independent, the above theorem, by choosing $U=W$, results that with probability $1-\delta$, $\gen(S,W) \leq \sqrt{2\sigma^2 \log(1/\delta)/n}$. This bound is  equal to the one obtainable by direct application of  Hoeffding's inequality. However, we may not be able to achieve this bound using Theorem~\ref{th:rdCompDist} with constant $U$. Since, while for example $I(S;W)=0$ under $P_{S,W}$, it may not be equal to zero under distribution $Q$, where $D_{KL}(Q\|P_{S,W})\leq \log(1/\delta)$; as under distribution $Q$, random variables $S$ and $W$ might be (weakly) dependent.

\subsection{Examples for Lipschitz loss}  \label{sec:LipsCor}
In the following, we show some consequences of Corollary~\ref{cor:lipschitz}.

\begin{corollary} \label{cor:LipschExamples} Suppose that the loss function $\ell(Z,w)$ is $\sigma$-subgaussian for any $w \in \mathcal{W}$.
\begin{itemize}
    \item[i.] [$\varepsilon$-net covering] Let $\mathcal{W}=\mathbb{R}^d$ and let $W$ with probability one take value in the $d$-dimensional ball $\mathcal{V}_d=\{w\in \mathbb{R}^d\colon \|w\| \leq r_0\}$ and suppose that for every $z,w,\hat{w}$, $|\ell(z,w)-\ell(z,\hat{w})| \leq \mathfrak{L} \|w-\hat{w}\|$. Then, for every $\delta>0$, with probability at least $1-\delta$,
    \begin{align*}
        \gen(S,W) \leq \min \limits_{\epsilon \geq 0} \left[\sqrt{\frac{2\sigma^2(d\log(2r_0/\epsilon)+\log(1/\delta))}{n}}+2\mathfrak{L} \epsilon\right].
    \end{align*}
    In particular, for $n \geq 16$, with probability at least $1-e^{-d/2}$, we have $ \gen(S,W) \leq  (4r_0\mathfrak{L}+\sigma\sqrt{d})\sqrt{\log(n)/n}$.
    \item[ii.] Suppose that $W \in  \{0,1\}^d$ is composed of $d$ i.i.d. elements distributed according to  Bernoulli distribution with an unknown parameter  $\mathbb{P}(W_i=1)=p$  and $|\ell(z,w)-\ell(z,\hat{w})| \leq \mathfrak{L} d_H(w,\hat{w})$, where $d_H$ is the Hamming distance.\footnote{For binary vectors $x=(x_1,\ldots,x_d)$ and $y=(y_1,\ldots,y_d)$, $d_H(x,y)\coloneq \sum_{i=1}^d \mathbbm{1}_{\{x_i \neq y_i\}}$.} Then, for every $\delta>0$, with probability at least $1-\delta$,
    \begin{align*}
        \gen(S,W) \leq \min \limits_{0 \leq \epsilon \leq d}\left[ \sqrt{\frac{2\sigma^2[d\log(2)-d h_b(\epsilon/d)+\log(1/\delta)]}{n}}+ 2\mathfrak{L}\epsilon\right],
    \end{align*}
    where $h_b(\cdot)$ is the binary entropy function, \emph{i.e.} $h_b(p)=-p\log(p)-(1-p)\log(1-p)$, for $p \in [0,1]$ and $0\log(0)=0$ by convention.
    \item[iii.]  Suppose that $W \in  \mathbb{R}^d$ is composed of $d$ i.i.d. elements distributed according to the two-sided exponential distribution $p(w_i)=\frac{\lambda}{2} e^{-\lambda |w_i|},~i \in[d]$  and $|\ell(z,w)-\ell(z,\hat{w})| \leq \mathfrak{L} \|w-\hat{w}\|_1$. Then, for every $\delta>0$, with probability at least $1-\delta$,
    \begin{align*}
        \gen(S,W) \leq \min \limits_{\epsilon\geq 0}\left[\sqrt{\frac{2\sigma^2\left(d R'(\epsilon,\delta)+\log(1/\delta)\right)}{n}}+2\mathfrak{L}\epsilon\right],
    \end{align*}
where $R'(\epsilon,\delta)$ is determined by
\begin{align*}
   \log(1/\delta)=\alpha \lambda -1  -\log(\alpha \lambda),
\end{align*}
in which $\alpha \coloneq \epsilon\exp(R'(\epsilon,\delta))/d$.
    \item[iv.] Suppose that $W \in  \mathbb{R}^d$ is composed of $d$ i.i.d. elements distributed according to the normal distribution $\mathcal{N}(0,\sigma_N^2)$  and $|\ell(z,w)-\ell(z,\hat{w})| \leq \mathfrak{L} \|w-\hat{w}\|_2^2$. Then, for every $\delta>0$, with probability at least $1-\delta$,
    \begin{align*}
        \gen(S,W) \leq \min \limits_{\epsilon\geq 0}\left[\sqrt{\frac{\sigma^2\left(d\log\left(\max\left(d\alpha^2/
    \epsilon,1\right)\right)+2\log(1/\delta)\right)}{n}}+2\mathfrak{L}\epsilon\right],
    \end{align*}
where $\alpha \geq \sigma$ is determined by
\begin{align*}
    \log(1/\delta)=\frac{1}{2}\left(\frac{\alpha^2}{\sigma^2_N}-1-\log\left(\frac{\alpha^2}{\sigma^2_N}\right)\right).
\end{align*}
\end{itemize}
\end{corollary}
The corollary is proved in Appendix~\ref{sec:LipschExamples}.


\section{Proofs} \label{sec:proofs}
In this section, we present the proofs of all our results, in the order of their appearances in the paper.


\subsection{Proof of Theorem~\ref{th:expectationCompressible}}\label{pr:TheoremExpComp}
Here we state the proof for the long version of the theorem, stated in Appendix~\ref{sec:expecRdDis2}. Note that as defined in that appendix,  $\deAm(w^m,\hat{w}^m;s^m)\coloneq \frac{1}{m} \sum \nolimits_{i=1}^m  \left(|\gen(s_i,w_i)|-|\gen(s_i,\hat{w}_i)|\right)$.

Before stating the proof, we show that having condition \eqref{def:vectorDistanceE} for the distortion functions $\left\{ \dem,|\dem|,\deAm\right\}$ guarantees that the expectation of the difference of the original and compressed algorithms does not exceed $\epsilon$. 
\begin{lemma} \label{lem:probToExpec}
If $\mathbb{E}[|\gen(S,W)|]<\infty$, then for $d_m \in \left\{ \dem,|\dem|,\deAm\right\}$ condition~\eqref{eq:errorZero} yields
\begin{align}
   \lim_{m \ra \infty}  \mathbb{E}_{(S,W)^{\otimes m}}\left[\min_{j \in [l_m]} d_{m}(W^m,\hat{\vc{w}}_j;S^m)\right]\leq \epsilon. \label{eq:errorZeroExp}
\end{align}
\end{lemma}
The above lemma is proved in Appendix~\ref{pr:probToExpec}. Now, we proceed with the proof of the extended version of Theorem~\ref{th:expectationCompressible}, appeared in Appendix~\ref{sec:expecRdDis2}.

\begin{proof}
\paragraph{Part i.} Suppose that for each $(s^m,w^m)$, $\hat{\vc{w}}(s^m,w^m)\coloneq \hat{\vc{w}}_j$ where $j =\argmin \limits_{j \in [l_m]} \dem(w^m,\hat{\vc{w}}_j;s^m)$, which will be denoted by $\hat{\vc{w}}=(\hat{w}_1,\ldots,\hat{w}_m)$ for simplicity. Then,
\begin{align*}
      \mathbb{E}_{(S,W)}&\left[\gen(S,W)\right] \\
      =& \frac{1}{m}\mathbb{E}_{(S,W)^{\otimes m}}\left[\sum \limits_{i=1}^m \gen(S_i,W_i)\right]\\
      =& \frac{1}{m}\mathbb{E}_{(S,W)^{\otimes m}}\left[\sum \limits_{i=1}^m \gen(S_i,W_i)-\gen(S_i,\hat{W}_i)\right]+\frac{1}{m}\mathbb{E}_{(S,W)^{\otimes m}}\left[\sum \limits_{i=1}^m \gen(S_i,\hat{W}_i)\right]\\
      \stackrel{(a)}{\leq} & \frac{1}{m}\mathbb{E}_{(S,W)^{\otimes m}}\left[\sum \limits_{i=1}^m \gen(S_i,\hat{W}_i)\right]+\epsilon+\varepsilon_m\\
      \leq & \frac{1}{m}\mathbb{E}_{S^{\otimes m}}\left[\max_{j \in [l_m]}\sum \limits_{i=1}^m \gen(S_i,\hat{w}_{j,i})\right]+\epsilon+\varepsilon_m\\
      \stackrel{(b)}{\leq} & \frac{1}{m}\sqrt{\frac{2 \sigma^2m \log(l_m)}{n}}+\epsilon+\varepsilon_m\\
      \stackrel{(c)}{\leq} & \sqrt{\frac{2 \sigma^2 R}{n}}+\epsilon+\varepsilon_m,
\end{align*}
where $(a)$ is by Lemma~\ref{lem:probToExpec}, $(b)$ is derived since $\sum \limits_{i=1}^m \gen(S_i,\hat{w}_{j,i})$ is $\sigma\sqrt{m/n}$-subgaussian, and $(c)$ is obtained by bounding $l_m \leq e^{mR}$. Taking the limit for $m \ra \infty$ completes the proof.

\paragraph{Part ii.} Similarly, let  $\hat{\vc{w}}(s^m,w^m)\coloneq \hat{\vc{w}}_j$ where $j = \argmin \limits_{j \in [l_m]} |\dem(w^m,\hat{\vc{w}}_j;s^m)|$. Then, we have
\begin{align*}
      \big|\mathbb{E}_{(S,W)}&\left[\gen(S,W)\right]\big| \\
      \leq & \frac{1}{m}\mathbb{E}_{(S,W)^{\otimes m}}\left[\big|\sum \limits_{i=1}^m \gen(S_i,W_i)\big|\right]\\
      \leq & \frac{1}{m}\mathbb{E}_{(S,W)^{\otimes m}}\left[\big|\sum \limits_{i=1}^m \gen(S_i,W_i)-\gen(S_i,\hat{W}_i)\big|\right]+\frac{1}{m}\mathbb{E}_{(S,W)^{\otimes m}}\left[\big|\sum \limits_{i=1}^m \gen(S_i,\hat{W}_i)\big|\right]\\
      \leq & \frac{1}{m}\mathbb{E}_{S^{\otimes m}}\left[\max_{j \in [l_m]}\big|\sum \limits_{i=1}^m \gen(S_i,\hat{w}_{j,i})\big|\right]+\epsilon+\varepsilon_m\\
      \leq & \frac{1}{m}\sqrt{\frac{2 \sigma^2m \log(2l_m)}{n}}+\epsilon+\varepsilon_m\\
      \leq & \sqrt{\frac{2 \sigma^2 (R+\log(2)/m)}{n}}+\epsilon+\varepsilon_m.
\end{align*}
Taking the limit for $m \ra \infty$ completes the proof.
\paragraph{Part iii.} Similarly, let  $\hat{\vc{w}}(s^m,w^m)\coloneq \hat{\vc{w}}_j$ where $j = \argmin \limits_{j \in [l_m]} \deAm(w^m,\hat{\vc{w}}_j;s^m)$. Then, we have

\begin{align*}
      \mathbb{E}_{(S,W)}&\left[\big|\gen(S,W)\big|\right] \\
      =& \frac{1}{m}\mathbb{E}_{(S,W)^{\otimes m}}\left[\sum \limits_{i=1}^m \big|\gen(S_i,W_i)\big|-\big|\gen(S_i,\hat{W}_i)\big|\right]+\frac{1}{m}\mathbb{E}_{(S,W)^{\otimes m}}\left[\sum \limits_{i=1}^m \big|\gen(S_i,\hat{W}_i)\big|\right]\\
      \leq & \frac{1}{m}\mathbb{E}_{S^{\otimes m}}\left[\max_{j \in [l_m]}\sum \limits_{i=1}^m \big|\gen(S_i,\hat{w}_{j,i})\big|\right]+\epsilon+\varepsilon_m\\
      \leq & \frac{1}{m}\sqrt{\frac{2 \sigma^2m \log(2^m l_m)}{n}}+\epsilon+\varepsilon_m\\
      \leq & \sqrt{\frac{2 \sigma^2 (R+\log(2))}{n}}+\epsilon+\varepsilon_m.
\end{align*}
Taking the limit for $m \ra \infty$ completes the proof.
\end{proof}

\subsection{Proof of Theorem~\ref{th:compressibilityBound}}\label{pr:compressibilityBound}
\begin{proof}
Fix $\epsilon$ and $\nu_1,\nu_2>0$. Assume that there exists a $\hat{W}\in \hat{\mathcal{W}}$ defined by the conditional distribution $P_{\hat{W}|S}$, such that  $\left|\mathbb{E}\left[\gen(S,W)-\gen(S,\hat{W})\right]\right|\leq \epsilon$. It is sufficient to show $R_E(\epsilon) \leq I(S;\hat{W})+\nu_1$. 

Denote the empirical distribution of a sequence $x^m$ by $\hat{P}_{x^m}$, \emph{i.e.} 
\begin{align}
\hat{P}_{x^m}(x) \coloneq \frac{\# i \in[m] \colon x_i=x}{m}. \label{eq:swTV}
\end{align}
Then, using the proof of \cite[Theorem~3]{Cuff09}, there exists a required sequence of $\left\{\mathcal{H}_m\right\}_{m \in \mathbb{N}}$ such that  $|\mathcal{H}_m|\leq e^{m(I(S;\hat{W})+\nu_1)}$  and such that for each $S^m$, a vector $\hat{W}^m(S^m) \in \mathcal{H}_m$, that we denote for ease of notations as $\hat{W}^m$, can be chosen such that
\begin{align}
    \|\hat{P}_{(S^m,\hat{W}^m)}(s,\hat{w})-P_{(S,\hat{W})}(s,\hat{w})\|_{TV} \stackrel{p}{\longrightarrow} 0. \label{eq:tvSWhat}
\end{align}
where $\stackrel{p}{\longrightarrow}$ means convergence in probability and \emph{TV} denotes the total variation distance between two distributions \cite[Definition~4]{Cuff09}. Moreover, by strong law of large numbers, for $m$ independent instances $(S_i,W_i)$ chosen according to $P_{S,W}$, we have
\begin{align}
    \|\hat{P}_{(S^m,W^m)}(s,w)-P_{S,W}(s,w)\|_{TV} \stackrel{p}{\longrightarrow} 0. \label{eq:tvSW}
\end{align}
This yields
\begin{align}
   \left| \frac{1}{m}\sum\limits_{i=1}^m \left[\gen(S_i,W_i)-\gen(S_i,\hat{W}_i)\right]\right| \leq& \left|\mathbb{E}\left[\gen(S,W)-\gen(S,\hat{W})\right]\right|+\epsilon_m \\\leq& \epsilon+\varepsilon_m,
\end{align}
where $\varepsilon_m$ vanishes as $m \ra \infty$. Now, 
\begin{align*}
   \mathbb{P}_{(S,W)^{\otimes m}} \left(\dem(W^m,\hat{W}^m;S^m) \geq  \epsilon +\nu_2 \right) \leq & \mathbb{P}_{(S,W)^{\otimes m}} \left(\epsilon+\varepsilon_m \geq \epsilon +\nu_2\right) \\
  \rightarrow& 0,
\end{align*}
where the last line is when $m \ra \infty$. This completes the proof.
\end{proof}

\subsection{Proof of Theorem~\ref{th:expecRdDis}}\label{pr:expecRdDis}
\begin{proof} We show
\begin{align}
     \left|\mathbb{E}\left[\gen(S,W)\right]\right| \leq \sqrt{\frac{2\sigma^2 R_E(\epsilon)}{n}}+\epsilon, \label{eq:firstExpec}
\end{align}
and the proof for the rest of bounds in \eqref{eq:expecRDis} is similar. Consider any Markov kernel $P_{\hat{W}|S}$ that satisfies $\left|\mathbb{E}\left[\gen(S,W)-\gen(S,\hat{W})\right]\right|\leq \epsilon$. Then,
\begin{align*}
    \left|\mathbb{E}\left[\gen(S,W)\right]\right|\leq&\left|\mathbb{E}\left[\gen(S,\hat{W})\right]\right|+ \epsilon\\
    \leq& \sqrt{\frac{2\sigma^2 I(S;\hat{W})}{n}}+\epsilon,
\end{align*}
where the last step is deduced from \cite[Theorem~1]{xu2017information}. This completes the proof.
\end{proof}


\subsection{Proof of Corollary~\ref{cor:dimExpec}} \label{pr:dimExpec}
\begin{proof} Let $n'_0$ be large enough such that for $n \geq n'_0$ and $\epsilon\coloneq 2\mathfrak{L}/\sqrt{n\mathfrak{L}^2}=2/\sqrt{n}$,
\begin{align*}
       \mathfrak{RD} (\epsilon/(2\mathfrak{L});P_W,\dw)/\log(2\mathfrak{L}/\epsilon) \leq 2 \dim_{\mathrm{R}}(P_W).
\end{align*}
Note that this holds due to the uniform convergence assumption of the corollary. Then, using  Corollary~\ref{cor:lipschitzExp}, we have
\begin{align*}
    \left|\mathbb{E}\left[\gen(S,W)\right]\right| &\leq \sqrt{\frac{2\sigma^2 \dim_{\mathrm{R}}(P_W,\delta) \log(n\mathfrak{L}^2)}{n}}+\sqrt{\frac{4}{n}}\\
    &\leq \sqrt{\frac{4\sigma^2 \dim_{\mathrm{R}}(P_W,\delta) \log(n\mathfrak{L}^2)}{n}},
\end{align*}
where the last inequality holds for $n \geq n_0$, where $n_0\geq n'_0$ is a sufficiently large integer.
\end{proof}



\subsection{Proof of Theorem~\ref{th:compressibility}} \label{sec:prCompressibility}

\begin{proof}  Some of the steps in this proof are identical to the proof of Theorem~\ref{th:tailTwoTerms}, by considering $X^m$ as $(\gen(S_1,W_1),\ldots,\gen(S_m,W_m))$ and $\hat{X}^m(j)$ as $(\gen(S_1,\hat{w}_{j,1}),\ldots,\gen(S_m,\hat{w}_{j,m}))$. Here, for the sake of completeness, we re-state all steps for the particular setup and notations used for the generalization error problem.

For any $\nu \in (0,\log(1/\delta))$ sufficiently small, choose $m_0$ such that for $m \geq m_0$,
\begin{align}
-\frac{1}{m} \log\left( \mathbb{P}_{(S,W)^{\otimes m}}\left(\mathcal{E}_m(\mathcal{H}_m,\epsilon;\dpm)\right) \right) \geq \log(1/\delta)-\nu.
\end{align}

For ease of notations, let $\mathcal{E}_m \coloneq \mathcal{E}_m(\mathcal{H}_m,\epsilon;\dpm)$. 
Then,
\begin{align*}
    \mathbb{P}_{S,W}&\left(\gen(S,W) \geq \Delta\right)^m \\
    =& \mathbb{P}_{(S,W)^{\otimes m}}\left(\forall i, \gen(S_i,W_i) \geq \Delta\right)\\
    \leq & \mathbb{P}_{(S,W)^{\otimes m}}\left(\forall i, \gen(S_i,W_i) \geq \Delta, \mathcal{E}_m^c \right)+\mathbb{P}_{(S,W)^{\otimes m}}\left(\mathcal{E}_m \right)\\
    \leq&  \mathbb{P}_{(S,W)^{\otimes m}}\left(\forall i, \gen(S_i,W_i) \geq \Delta, \mathcal{E}_m^c \right)+ e^{-m(\log(1/\delta)-\nu)}\\
    \leq& \mathbb{P}_{S^{\otimes m}}\left(\exists w^m \colon \forall i, \gen(S_i,w_i) \geq \Delta, \mathcal{E}_m^c \right)+ e^{-m(\log(1/\delta)-\nu)}\\
    \leq&  \mathbb{P}_{S^{\otimes m}}\left(\exists j \in [l_m], \{\Delta_i\}_{i=1}^m\in \mathbb{R} \colon \forall i,  \gen(S_i,\hat{w}_{j,i}) \geq \Delta-\Delta_i ,\sum_{i=1}^m \Delta_i \leq m\epsilon \right)+ e^{-m(\log(1/\delta)-\nu)}\\
    \leq&  \mathbb{P}_{S^{\otimes m}}\left(\exists j \in [l_m], \{\Delta_i\}_{i=1}^m\in \mathbb{R} \colon   \sum_{i=1}^m \gen(S_i,\hat{w}_{j,i}) \geq \sum_{i=1}^m \left(\Delta-\Delta_i\right) ,\sum_{i=1}^m \Delta_i \leq m\epsilon \right)+ e^{-m(\log(1/\delta)-\nu)}\\
    \leq &  \mathbb{P}_{S^{\otimes m}}\left(\exists j \in [l_m] \colon   \sum_{i=1}^m \gen(S_i,\hat{w}_{j,i}) \geq m\left(\Delta-\epsilon\right)  \right)+ e^{-m(\log(1/\delta)-\nu)}\\
    \leq & \sum \limits_{j \in [l_m]} \mathbb{P}_{S^{\otimes m}}\left(\sum_{i=1}^m \gen(S_i,\hat{w}_{j,i}) \geq m\left(\Delta-\epsilon\right)  \right)+ e^{-m(\log(1/\delta)-\nu)}\\
    \stackrel{(a)}{\leq} & \sum \limits_{j \in [l_m]} e^{-mn(\Delta-\epsilon)^2/(2\sigma^2)}+ e^{-m(\log(1/\delta)-\nu)}\\
    \stackrel{(a)}{\leq} & e^{m(R-n(\Delta-\epsilon)^2/(2\sigma^2))}+ e^{-m(\log(1/\delta)-\nu)}\\
    \stackrel{(b)}{\leq} & 2 e^{-m(\log(1/\delta)-\nu)}
\end{align*}
where $(a)$ is derived using the Hoeffding's inequality, $(b)$ is derived since $l_m \leq e^{mR}$, and $(c)$  is derived by choosing $\Delta$  as $\Delta\coloneq \sqrt{\frac{2\sigma^2(R+\log(1/\delta))}{n}}+\epsilon$. The proof completes by taking the $m$'th root of both sides, and since $\nu$ can be chosen arbitrarily small.

\end{proof}


\subsection{Proof of Theorem~\ref{th:rdCompDist}}\label{pr:rdCompDist}
Theorem~\ref{th:rdCompDist} is stated in Section~\ref{sec:tailMain} for $U$ being a constant and in Appendix~\ref{sec:tailRdDis2} has been extended to take into account the stochasticity of the algorithm. In the following, we first state the proof for finite sets and for the $U$ being a constant using Theorem~\ref{th:compressibility}. The result can be extended to the case of arbitrary $U$ that satisfies the conditions of the theorem, and to infinite sets, with some further assumptions on $(S,W)$, using the quantization technique used in the proof of \cite[Theorem~3.6]{elgamal_kim_2011} and by applying \cite[Theorem~1]{Iriyama2005} and its adaptation for the memoryless sources in \cite[Theorem~3]{bakshi2005error}. However, for the general case, we state an alternative proof that applies the Donsker–Varadhan's variational representation of the KL divergence.

\subsubsection{First Proof}
\begin{proof} Suppose that $\mathcal{S} \times \mathcal{W}$ is a finite set and $U$ is a constant. We start by showing that for every $\epsilon \in \mathbb{R}$ and any $\nu_1,\nu_2>0$, the algorithm $\mathcal{A}(S)$ is $\left(R(\delta,\epsilon)+\nu_1,\epsilon+\nu_2,\delta;\{\dpm\}_m\right)$-exponentially compressible. Our proof is similar to \cite[Theorem~1]{marton1974}.

Let $Q_m$ be an arbitrary type of $\mathcal{S}^m \times \mathcal{W}^m$. For the definition of the type, refer to the beginning of the appendices. Define
\begin{align}
\mathcal{Q}_m(\delta) \coloneq \left\{Q_m \colon D_{KL}\left(Q_m \| P_{S,W}\right)\leq \log(1/\delta) \right\}.
\end{align}
Note that,
\begin{align}
    \mathbb{P}_{(S,W)^{\otimes m}}\left(\mathcal{T}(S^m,W^m) \notin \mathcal{Q}_m(\delta)\right)=& \sum \limits_{Q'_m \notin \mathcal{Q}(\delta)}  \mathbb{P}_{(S,W)^{\otimes m}}\left(\mathcal{T}(S^m,W^m) =Q'_m \right) \nonumber \\
    \stackrel{(a)}{\leq} &  \sum \limits_{Q'_m \notin \mathcal{Q}(\delta)}e^{-m  D_{KL}\left(Q'_m \| P_{S,W}\right)}\nonumber \\
    \stackrel{(b)}{\leq} & m^{\left|\mathcal{S}\right| \times \left|\mathcal{W}\right|} e^{m\log(\delta)}\nonumber \\
    = & e^{m\left(\log(\delta)+\left|\mathcal{S}\right| \times \left|\mathcal{W}\right| \log(m)/m\right)}\nonumber \\
    = & (\delta+\varepsilon_m)^m, \label{eq:typeProb}
\end{align}
where $\lim_{m \ra \infty } \varepsilon_m =0$, the step $(a)$ is due to \cite[Theorem~11.1.4]{CoverTho06}, and the step $(b)$ is deduced since number of types can be bounded by $m^{\left|\mathcal{S}\right| \times \left|\mathcal{W}\right|}$.

First, we state a variant of type covering lemma \cite[Section~6.1.2, Lemma~1]{Berger1975} (appeared also in  \cite[Lemma~9.1]{CsisKor82}), proved in Appendix~\ref{pr:typeCoveringExt}:

\begin{lemma} \label{lem:typeCoveringExt}
For any $\nu>0$ and any type $Q_m$, there exists  a  hypothesis book  $\mathcal{H}_{Q_m}=\{\vc{\hat{w}}_{Q_m,j}, j \in [l_{Q_m}]\} \subseteq \mathcal{\hat{W}}^m$, such that $l_{Q_m} \leq  e^{mR_{Q_m}}$, where
\begin{align}
    R_{Q_m} =\mathfrak{RD}^*(\epsilon;Q_m) +\varepsilon_{Q_m},
\end{align}
and $\lim_{m \ra \infty } \varepsilon_{Q_m} =0$, and such that for $m \geq m_{\nu,Q_m}$,
\begin{align}  \mathbb{P}_{(S,W)^{\otimes m}}\left(\mathcal{T}(S^m,W^m)=Q_m ,\min_{j \in  [l_{Q_m}]} \dpm\left(W^m,\vc{\hat{w}}_{Q_m,j};S^m\right)\geq \epsilon+\nu \right)=0.\label{eq:errorZeroType}
\end{align}
\end{lemma}

Let $m\geq m_{\nu}$, where $m_{\nu}$ is sufficiently large such that the above lemma holds for all types. Letting $\mathcal{H}_m \coloneq \bigcup \limits_{Q_m \in \mathcal{Q}_m(\delta) } \mathcal{H}_{Q_m}$, we have
\begin{align}
    l_m \leq & \sum \limits_{Q_m \in \mathcal{Q}_m(\delta)} l_{Q_m} \nonumber \\
    & \leq m^{\left|\mathcal{S}\right| \times \left|\mathcal{W}\right|} e^{m \left(\max \limits_{Q_m \in \mathcal{Q}_m(\delta)}\left[ \mathfrak{RD}^*(\epsilon;Q_m) +\varepsilon_{Q_m}\right]\right)} \nonumber\\
    = &e^{m \left(\max \limits_{Q_m \in \mathcal{Q}_m(\delta)} \mathfrak{RD}^*(\epsilon;Q_m) +\varepsilon'_{m}\right)}, \label{eq:CodeLength}
\end{align}
where $\lim_{m \ra \infty } \varepsilon'_m =0$. Moreover,
\begin{align*}
   \mathbb{P}_{(S,W)^{\otimes m}}&\left(\mathcal{E}_m(\mathcal{H}_m,\epsilon+\nu;\dpm)\right)\\
   \leq&   \mathbb{P}_{(S,W)^{\otimes m}}\left((S^m,W^m) \notin \mathcal{Q}_m\right)\\
   &+\mathbb{P}_{(S,W)^{\otimes m}}\left((S^m,W^m) \in \mathcal{Q}_m,\min_{Q_m \in \mathcal{Q}_m} \min_{j \in  [l_{Q_m}]} \dpm\left(W^m,\vc{\hat{w}}_{Q_m,j};S^m\right)\geq \epsilon+\nu\right)\\
   \stackrel{(a)}{\leq} &  (\delta+\varepsilon_m)^m.
\end{align*}
where the last steps is derived using Lemma~\ref{lem:typeCoveringExt} and relation \eqref{eq:typeProb}.

 Hence,
\begin{align}
   \lim \limits_{m \ra \infty}-\frac{1}{m}\log \left(\mathbb{P}_{(S,W)^{\otimes m}}\left(\mathcal{E}_m(\mathcal{H}_m,\epsilon+\nu;\dpm)\right)\right) \leq \log(1/\delta). \label{eq:CodePr}
\end{align}
The relations \eqref{eq:CodeLength} and \eqref{eq:CodePr} show that for every $\epsilon \in \mathbb{R}$ and any $\nu_1,\nu_2>0$, the algorithm $\mathcal{A}(S)$ is $\left(R(\delta,\epsilon)+\nu_1,\epsilon+\nu_2,\delta;\{\dpm\}_m\right)$-exponentially compressible. Using Theorem~\ref{th:compressibility} completes the proof.
\end{proof}
\subsubsection{Second Proof}
\begin{proof} 
In this proof, since we use Theorem~\ref{th:tailX}, we denote $P_{U,S,W}$ by $\mu_{U,S,W}$, to be compatible with the notations of Theorem~\ref{th:tailX}. Note that $\mu_{U,S,W}=\mu_S\mu_U\mu_{W|S,U}$ where $U$ is an independent noise in the algorithm and $W$ is the output hypothesis of the algorithm. 

Let  $X\coloneq \gen(W,S)$. This defines $\mu_{S,U,W,X}$. 
Writing Theorem \ref{th:tailX} with the choice of $Y=(S,U,W)$, we get that
\begin{align}&\log\mathbb{P}_{X\sim \mu_X}\left(X\geq \Delta\right)\label{eqn19982}
\\&\leq \max\bigg[\log(\delta),  ~~~\sup_{\nu_{S,U,W,X}\in\mathcal{G}}~~~\inf_{p_{\hat X|S,U,W}\in\mathcal{Q}(\nu),~q_{\hat X|S,U,W},~\lambda\geq 0}\bigg\{D_{KL}(p_{\hat X|S,U,W}\nu_{S,U,W}\|q_{\hat X|S,U,W}\nu_{S,U,W})\nonumber
\\&\qquad\qquad\qquad\qquad\qquad\qquad\qquad\qquad\qquad\qquad\qquad\qquad-\lambda(\Delta-\epsilon)+\log\mathbb{E}_{\mu_{S,U,W}q_{\hat X|S,U,W}}[e^{\lambda \hat X}]\bigg\}\bigg]\nonumber
\end{align}
where
$\mathcal{G}$ is the following set of distributions:
\begin{align*}
    \mathcal{G}=\{\nu_{X,S,U,W}: D_{KL}(\nu_{X,S,U,W}\|\mu_{X,S,U,W})\leq \log(1/\delta),
    \}
\end{align*}
and $\mathcal{Q}(\nu)$ is the set of conditional distributions
$p_{\hat X|S,U,W}$ such that 
under
$p_{\hat X|S,U,W}\nu_{S,U,W}$ we have: \[\left[\inf_{x\in \Supp(\nu_X)}x\right]-\mathbb{E}[\hat X] \leq \epsilon.\]

Since $X=\gen(W,S)$ under $\mu_{X,U,W,S}$, and $D_{KL}(\nu_{X,S,U,W}\|\mu_{X,S,U,W})<\infty$, we obtain that $X=\gen(W,S)$ under $\nu_{X,U,W,S}$ too. Therefore, the supremum is over distribution $\nu$ of the form $\nu_{X,U,W,S}=\nu_{U,W,S}\mu_{X|U,W,S}$. This implies that
\[
D_{KL}(\nu_{X,S,U,W}\|\mu_{X,S,U,W})=D_{KL}(\nu_{S,U,W}\|\mu_{S,U,W})
\]
and
\[\inf_{x\in \Supp(\nu_X)}x=\inf_{\substack{(s',w') \in \Supp(\nu_{S,W})}}  
      \left[\gen(s',w')\right].
\]

Now, take some arbitrary $\nu_{S,U,W}$, and also take some arbitrary
$p_{\hat W_1|U,S,W}$ satisfying 
\[p_{\hat W_1|U,S,W}=p_{\hat W_1|U,S}\] where $\hat W_1\in\mathcal{W}$ belongs to the hypothesis space and
\[\inf_{\substack{(s',w') \in \Supp(\nu_{S,W})}}  
      \left[\gen(s',w')\right]-\mathbb{E}_{\nu_{U,S}p_{\hat W_1|U,S}}\left[ \gen(\hat W_1,S) \right]\leq \epsilon.
\]
Let $\hat{X}_1=\gen(\hat W_1,S)$. Then, $p_{\hat X_1|S,U,W}\in \mathcal{Q}(\nu)$.

Next, we define $\hat W_2$ to have a joint distribution of the form
$p_{\hat W_2|U}\nu_{S,U,W,X}$ 
such that the marginal joint distribution of $(\hat W_2, U)$ is the same as $(\hat W_1, U)$ under $p_{\hat W_1|U,S}\nu_{S,U,W,X}$. Note that $p_{\hat W_2|U,S}=p_{\hat W_2|U}$ is assumed here by the fact that $\hat W_2$ has a joint distribution of the form
$p_{\hat W_2|U}\nu_{S,U,W,X}$.

Let  $\hat{X}_2=\gen(\hat W_2,S)$.  
 Take $q_{\hat X_2|S,U,W}$ to be the conditional distribution of $\hat{X}_2$ given $S,U,W$. Also, take $p_{\hat X_1|S,U,W}\in \mathcal{Q}(\nu)$ to be the conditional distribution of $\hat{X}_1$ given $S,U,W$. We evaluate the above bound with $q_{\hat X_2|S,U,W}$ and $p_{\hat X_1|S,U,W}$. Then,
\begin{align*}D_{KL}(p_{\hat X_2|S,U,W}\nu_{S,U,W}\|q_{\hat X_1|S,U,W}\nu_{S,U,W})&\leq
D_{KL}(p_{\hat W_1|S,U}\nu_{S,U,W}\|p_{\hat W_2|S,U}\nu_{S,U,W})
\\&=
D_{KL}(p_{\hat W_1|S,U}\nu_{S,U,W}\|p_{\hat W_1|U}\nu_{S,U,W})
\\&=
D_{KL}(p_{\hat W_1|S,U}\nu_{S,U}\|p_{\hat W_1|U}\nu_{S,U})
\\&=I_{p_{\hat{W}_1|S,U}\nu_{S,U}}(\hat{W}_1;S|U).
\end{align*}

Finally, under $\mu_{U,S,W}q_{\hat X_2|S,U,W}$ we have that $S=(Z_1, \ldots, Z_n)$ is an i.i.d. sequence according to $\mu_Z$. Moreover, in $\mu_{U,S}$ we have that $U$ is independent of $S$. Furthermore, in $q_{\hat X_2|S}$, we have $p_{\hat W_2|U,S}=p_{\hat W_2|U}$, which together with independence of $U$ and $S$ implies that $\hat W_2$ is independent of $S$. Therefore, $\gen(\hat W_2,S)$ is the sum of $n$ i.i.d.\ variables, and since $\ell(Z,\hat{w})$ is $\sigma$-subgaussian, hence $\gen(\hat W_2,S)$ is $\sigma/\sqrt{n}$-subgaussian. Therefore, we can compute its moment generating function. Thus, $\log\mathbb{E}\left[e^{\lambda \hat{X}}\right]\leq \lambda^2 \sigma^2/2n$ and letting $\lambda \coloneq n(\Delta-\epsilon)/\sigma^2$ yields
\begin{align*}
    D_{KL}(p_{\hat X|S,U}\nu_{S,U}\|q_{\hat X|S,U}\nu_{S,U})-\lambda(\Delta-\epsilon)&+\log\mathbb{E}_{\mu_{S,U}q_{\hat X|S,U}}[e^{\lambda \hat X}] \\
    &\leq I_{p_{\hat W_1|S,U}\nu_{S,U}}(\hat{W}_1;S|U)-\frac{n(\Delta-\epsilon)^2}{2 \sigma^2}.
\end{align*}
Hence, denoting $\mathcal{G}_{S,U,W}=\{\nu_{S,U,W}: D_{KL}(\nu_{S,U,W}\|\mu_{S,U,W})\leq \log(1/\delta)
    \}$, and by definition \eqref{eq:rdStar}, we have
\begin{align}&\log\mathbb{P}_{X\sim \mu_X}\left(X\geq \Delta\right)
\\&\leq \max\bigg[\log(\delta),  ~~~\sup_{\nu_{S,U,W}\in\mathcal{G}_{S,U,W}}~~~\bigg\{  \mathfrak{RD}^{*} (\epsilon;Q) -\frac{n(\Delta-\epsilon)^2}{2 \sigma^2}\bigg\}\bigg].\nonumber
\end{align}
Letting
\begin{align*}
    \Delta =\sup_{\nu_{S,U,W}\in\mathcal{G}_{S,U,W}} \sqrt{\frac{2\sigma^2(\mathfrak{RD}^{*} (\epsilon;Q)+\log\left(1/\delta\right))}{n}}+\epsilon,
\end{align*}
completes the proof.\end{proof}


\subsection{Proof of Theorem~\ref{th:tailX}} \label{pr:tailX}

We first claim that
\begin{align}\nonumber&\log\mathbb{P}_{X\sim \mu_X}\left(X\geq \Delta\right)
\\&\leq \sup_{\nu_{YX}\ll\mu_{YX}}~~\inf_{p_{\hat X|Y}\in\mathcal{Q}(\nu),~\lambda\geq 0}\left\{-D_{KL}(\nu_{YX}\|\mu_{YX})-\lambda\bigg[(\Delta-\epsilon)- \int\hat x d(p_{\hat X|Y}\nu_{Y,X})\bigg]_+\right\}.\label{eqn17611}
\end{align}
This follows from Lemma \ref{lem:variationalTail} because if we look at $\nu_{YX}$'s of the form $\nu_{YX}=\nu_X\mu_{Y|X}$, we have 
\[D_{KL}(\nu_{YX}\|\mu_{YX})= D_{KL}(\nu_{X}\|\mu_{X}).\]
Moreover, the term $\int\hat x d(p_{\hat X|Y}\nu_{Y,X})$ depends only on the marginal distribution on $\hat X$ under $p_{\hat X|Y}\nu_{Y,X}$. 

Next, given $p_{\hat X|Y}$, let
 $p_{\hat X,Y}=p_{\hat X|Y}\nu_{Y}$. For every $Y=y$, consider the conditional distribution $p_{\hat X|Y=y}$ induced by this joint distribution. Lemma 
\ref{lemmaag2} yields
\begin{align}\inf_{q_{\hat X}}\left[D_{KL}(p_{\hat X|Y=y}\|q_{\hat X})+\log\mathbb{E}_{q}[e^{\lambda \hat X}]\right]=\lambda \mathbb{E}_{p}[\hat X|Y=y],
\end{align}
By averaging this over $y$ using the distribution $\nu_Y$, we obtain
\begin{align}\inf_{q_{\hat X|Y}}\left[ D_{KL}(p_{\hat X|Y}\nu_Y\|q_{\hat X|Y}\nu_Y)+\mathbb{E}_{\nu_Y}\log\mathbb{E}_{q}[e^{\lambda \hat X}|Y]\right]=\lambda\int\hat x d(p_{\hat X|Y}\nu_{Y}).
\end{align}
This equality along with \eqref{eqn17611} yield
\begin{align*}\nonumber&\log\mathbb{P}_{X\sim \mu_X}\left(X\geq \Delta\right)
\\&\leq \sup_{\nu_{YX}\ll\mu_{YX}}~~\inf_{p_{\hat X|Y}\in\mathcal{Q}(\nu)}~\inf_{q_{\hat X|Y},~\lambda\geq 0}\bigg\{-D_{KL}(\nu_{YX}\|\mu_{YX})
\\&\qquad\qquad\qquad\qquad-\bigg[\lambda(\Delta-\epsilon)-D_{KL}(p_{\hat X|Y}\nu_Y\|q_{\hat X|Y}\nu_Y)-\mathbb{E}_{Y\sim\nu_Y}\log\mathbb{E}_{q}[e^{\lambda \hat X}|Y]\bigg]_+\bigg\}.
\end{align*}
From the Donsker-Varadhan's identity we obtain the inequality
\[
\mathbb{E}_{Y\sim\nu_Y}\log\mathbb{E}_{q}[e^{\lambda \hat X}|Y]\leq D_{KL}(\nu_{Y}\|\mu_{Y})+\log \mathbb{E}_{\mu_Y}\mathbb{E}_{q}[e^{\lambda \hat X}|Y]=
D_{KL}(\nu_{Y}\|\mu_{Y})+\log \mathbb{E}_{\mu_Yq_{\hat X|Y}}\mathbb{E}[e^{\lambda \hat X}].
\]
Therefore,
\begin{align*}\nonumber&\log\mathbb{P}_{X\sim \mu_X}\left(X\geq \Delta\right)
\\&\leq \sup_{\nu_{YX}\ll\mu_{YX}}~~\inf_{p_{\hat X|Y}\in\mathcal{Q}(\nu),~q_{\hat X|Y},~\lambda\geq 0}\bigg\{-D_{KL}(\nu_{YX}\|\mu_{YX})-\bigg[\lambda(\Delta-\epsilon)-D_{KL}(p_{\hat X|Y}\nu_Y\|q_{\hat X|Y}\nu_Y)
\\&\qquad\qquad\qquad\qquad\qquad\qquad\qquad\qquad\qquad\qquad\qquad-D_{KL}(\nu_{Y}\|\mu_{Y})-\log \mathbb{E}_{\mu_Yq_{\hat X|Y}}\mathbb{E}[e^{\lambda \hat X}]\bigg]_+\bigg\}.
\end{align*}

The desired inequality follows from here since
$
D_{KL}(\nu_{YX}\|\mu_{YX})\geq D_{KL}(\nu_{Y}\|\mu_{Y})
$.


\subsection{Proof of Theorem~\ref{th:tailTwoTerms}} \label{pr:tailTwoTerms}

\begin{proof} Let $X^m=(X_1,\ldots,X_m)$. We can write
\begin{align}
    \mathbb{P}&\left(X\geq \Delta\right)^m \nonumber\\
    =& \mathbb{P}\left(\min_{i\in [m]} X_i \geq \Delta\right)\nonumber\\
    \leq & \mathbb{P}\left(\min_{i \in [m]} X_i \geq \Delta, \min_{j \in [k]} \rho(X^m,{\hat X}^m(j))\leq  \epsilon \right)+\mathbb{P}\left(\min_{j \in [k]} \rho(X^m,{\hat X}^m(j))>  \epsilon \right)\nonumber\\
    =&  \mathbb{P}\left(\exists j \in [k]: \min_{i\in [m]} X_i \geq \Delta,  \rho(X^m,{\hat X}^m(j))\leq \epsilon \right)+\mathbb{P}\left(\min_{j \in [k]} \rho(X^m,{\hat X}^m(j))>  \epsilon \right)\nonumber\\
    = & \mathbb{P}\left(\exists j \in [k]:\rho(X^m,{\hat X}^m(j))+\frac1m \sum_{i=1}^m\hat X_i(j)\geq \Delta, \rho(X^m,{\hat X}^m(j))\leq \epsilon\right)\label{stnt1}
    \\&\qquad+\mathbb{P}\left(\min_{j \in [k]} \rho(X^m,{\hat X}^m(j))>  \epsilon \right)\nonumber
    \\
    \leq & \mathbb{P}\left(\exists j \in [k]:\frac1m \sum_{i=1}^m\hat X_i(j)\geq \Delta- \epsilon\right)+\mathbb{P}\left(\min_{j \in [k]} \rho(X^m,{\hat X}^m(j))>  \epsilon \right)\nonumber
    \\
    \leq & \sum_{j=1}^k\mathbb{P}\left(\frac1m \sum_{i=1}^m\hat X_i(j)\geq \Delta- \epsilon\right)+\mathbb{P}\left(\min_{j \in [k]} \rho(X^m,{\hat X}^m(j))>  \epsilon \right),\nonumber
\end{align}
where \eqref{stnt1} follows from the definition of $\rho$ (Definition~\ref{rho-def}). 
\end{proof}

\subsection{Proof of Theorem~\ref{th:expectationCompressibleCond}}\label{pr:expectationCompressibleCond}

\begin{proof}
First, note that as established in \cite[Proof of Theorem~5.1]{steinke2020reasoning},
\begin{align*}
    \mathbb{E}\left[\gen(S,W)\right]&= \mathbb{E}_{\Scap}\mathbb{E}_{\Kbold,W|\Scap}\left[f(\Scap,\Kbold,W)\right].
\end{align*}
The rest of the proof is similar to the proof of Theorem~\ref{th:expectationCompressible}, by considering the term $f(\Scap,\Kbold,W)$, instead of $\gen(S,W)$, and by noting that conditioned on $\Scap=\Smin$ and $W=w$, for every $j \in [n]$, $(-1)^{\Knorm_j}\left(\ell(\Smin_{j,1},w)-\ell(\Smin_{{j,2}},w)\right)$ is a bounded process in the range $[-1,1]$, with average zero, that takes values among $\ell(\Smin_{j,1},w)-\ell(\Smin_{{j,2}},w)$ and $-\left(\ell(\Smin_{j,1},w)-\ell(\Smin_{{j,2}},w)\right)$, uniformly. Hence, $f(\Smin,\Kbold,w)$ is $1/\sqrt{n}$-subgaussian. We show the proof for part i. The other parts follow similarly. Let  $\vc{\hat{w}}(\Smin,\kbold^m,w^m)\coloneq \hat{\vc{w}}_j(\Smin)$ where $j=\argmin \limits_{j \in [l_m(\Smin)]} \dsm(w^m,\hat{\vc{w}}_j(\Smin);\Smin,\kbold^m)$. We denote it simply by $\hat{\vc{w}}=(\hat{w}_1,\ldots,\hat{w}_m)$. Then, we have
\begin{align*}
      \mathbb{E}_{(\Kbold,W|\Smin)}&\left[f(\Smin,\Kbold,W)\right] \\
      \leq & \frac{1}{m}\mathbb{E}_{(\Kbold,W|\Smin)^{\otimes m}}\left[\sum \limits_{i=1}^m f(\Smin,\Kbold_i,W_i)\right]\\
      \leq & \frac{1}{m}\mathbb{E}_{(\Kbold,W|\Smin)^{\otimes m}}\left[\sum \limits_{i=1}^m f(\Smin,\Kbold_i,W_i)-f(\Smin,\Kbold_i,\hat{W}_i)\right]+\frac{1}{m}\mathbb{E}_{(\Kbold,W|\Smin)^{\otimes m}}\left[\sum \limits_{i=1}^m f(\Smin,\Kbold_i,\hat{W}_i)\right]\\
      \leq & \frac{1}{m}\mathbb{E}_{\Kbold^{\otimes m}}\left[\max_{j \in [l_m(\Smin)]}\sum \limits_{i=1}^m f(\Smin,\Kbold_i,\hat{w}_{j,i}) \right]+\epsilon(\Smin)+\varepsilon_m\\
      \leq & \frac{1}{m}\sqrt{\frac{2 m \log(l_m(\Smin))}{n}}+\epsilon(\Smin)+\varepsilon_m\\
      \leq & \sqrt{\frac{2  R(\Smin)}{n}}+\epsilon(\Smin)+\varepsilon_m.
\end{align*}
Taking the limit for $m \ra \infty$ completes the proof.
\end{proof}


\subsection{Proof of Corollary~\ref{cor:VCexp}}\label{pr:VCexp}
\begin{proof}
Let $U$ be the stochasticity of the algorithm (\emph{e.g.} the randomness in choosing the training data for each batch in the SGD algorithm) in a sense that for a given dataset $s=(z_1,\ldots,z_n)$ and based on the sequence of values $\{(\ell(z_1,w),\ldots,\ell(z_n,w))\}_{w\in \mathcal{W}}$, the algorithm chooses a fixed hypothesis $w$ conditioned on $U=u$. 

If an algorithm has the VC-dimension $d$ and for a fixed $\Scap=\Smin$, the set of possible pairs $\{(\ell(z_1,w),\ldots,\ell(z_n,w))\}_w$, where $z_i=\Smin_{\kbold}$ for some $\kbold \in \{1,2\}^n$, is bounded by the set of possible $\{(\ell(z_{1,1},w),\ell(z_{1,2},w),\ldots,\ell(z_{n,1},w),\ell(z_{n,2},w))\}_w$, and the latter is bounded by $(2en/d)^d$ due to Sauer-Shelah lemma \cite{SAUER1972145,Shelah72}. Hence, $I(\Kbold,W |\Smin,u) \leq d \log (2en/d)$. Using Theorem~\ref{th:expCond} completes the proof.
\end{proof}


\subsection{Proof of Theorem~\ref{th:compressibilityCond}}\label{pr:compressibilityCond}
\begin{proof}
Let $\Delta_1\coloneq \sqrt{\log(2/\delta)/n}$, $\Delta_2\coloneq \sup_{\Smin} \sqrt{2(R(\Smin)+\log\left(2/\delta\right))/n}+\epsilon(\Smin)$, and let $\bar{S}\sim \mu^{\otimes n}$ be independent of $(S,W)$. Then,
\begin{align*}
   \mathbb{P}\left(\gen(S,W) \geq \Delta_1+\Delta_2 \right) =&\mathbb{P}\left(\mathbb{E}_{\bar{S}}\left[\hat{\mathcal{L}}(\bar{S},W)\right]-\hat{\mathcal{L}}(\bar{S},W)+\hat{\mathcal{L}}(\bar{S},W)-\hat{\mathcal{L}}(S,W) \geq \Delta_1+\Delta_2 \right) \\
   \leq &\mathbb{P}\left(\mathbb{E}_{\bar{S}}\left[\hat{\mathcal{L}}(\bar{S},W)\right]-\hat{\mathcal{L}}(\bar{S},W) \geq  \Delta_1 \right)+\mathbb{P}\left(\hat{\mathcal{L}}(\bar{S},W)-\hat{\mathcal{L}}(S,W) \geq  \Delta_2 \right)  \\
   \leq &\delta/2+\mathbb{P}\left(\hat{\mathcal{L}}(\bar{S},W)-\hat{\mathcal{L}}(S,W) \geq  \Delta_2 \right).
\end{align*}
It remains to upper bound the second term by $\delta/2$. Denote $\Scap \in \mathcal{Z}^{2 \times n}$ as concatenation of $S$ and $\bar{S}$, such that for some $\Kbold \in \{1,2\}^n$, $Z_i=\Scap_{i,\Knorm_i}$ and $\bar{Z}_i=\Scap_{i,\bar{\Knorm}_i}$. As before, we denote $S=\Scap_{\Kbold}$ and $\bar{S}=\Scap_{\bar{K}}$. The joint distribution of $\Scap,\Kbold,W$ is $P_{\Scap}P_{\Kbold}P_{W|\Scap_{\Kbold}}$, where $P_{\Scap}=\mu^{\otimes 2n}$ and $P_{\Kbold}$ is uniform over $\{1,2\}^n$. Now,
\begin{align*}
   \mathbb{P}\left(\hat{\mathcal{L}}(\bar{S},W)-\hat{\mathcal{L}}(S,W) \geq  \Delta_2 \right)=&\mathbb{P}\left(\hat{\mathcal{L}}(\Scap_{\bar{\Kbold}},W)-\hat{\mathcal{L}}(\Scap_{\Kbold},W) \geq  \Delta_2 \right)\\
   =&\mathbb{P}\left(f(\Scap,\Kbold,W) \geq  \Delta_2 \right)\\
   \leq & \max_{\Smin} \mathbb{P}\left(f(\Smin,\Kbold,W) \geq  \Delta_2 \right).
\end{align*}
The rest of proof is to bound $\mathbb{P}_{\Kbold,W|\Smin}\left(f(\Smin,\Kbold,W) \geq  \Delta_2 \right)$ for a fixed $\Smin$. Similar to the proof of Theorem~\ref{th:compressibility}, for any $\nu \in (0,\log(2/\delta))$ sufficiently small, choose $m_0$ such that for $m \geq m_0$,

\begin{align}
 \lim_{m \ra \infty}  \left[-\frac{1}{m} \log\left(  \mathbb{P}_{(\Kbold,W|\Smin)^{\otimes m}}\left(\min \limits_{j \in [l_m(\Smin)]}\dspm(W^m,\hat{\vc{w}}_j(\Smin);\Smin,\Kbold^m) >\epsilon(\Smin) \right)\right)   \right]  \geq \log(2/\delta)-\nu. \nonumber
\end{align}
For ease of notations, let $\mathcal{E}_m$ be the event that $\min \limits_{j \in [l_m(\Smin)]}\dspm(W^m,\hat{\vc{w}}_j(\Smin);\Smin,\Kbold^m) >\epsilon(\Smin)$. Then,
\begin{align*}
     &\hspace{-0.2 cm}\mathbb{P}_{\Kbold,W|\Smin}\left(f(\Smin,\Kbold,W)\geq \Delta_2\right)^m \\=& \mathbb{P}_{(\Kbold,W|\Smin)^{\otimes m}}\left(\forall i, f(\Smin,\Kbold_i,W_i) \geq \Delta_2\right)\\
    \leq & \mathbb{P}_{(\Kbold,W|\Smin)^{\otimes m}}\left(\forall i, f(\Smin,\Kbold_i,W_i) \geq \Delta_2, \mathcal{E}_m^c \right)+\mathbb{P}_{(\Kbold,W|\Smin)^{\otimes m}}\left(\mathcal{E}_m \right)\\
    \leq&  \mathbb{P}_{(\Kbold,W|\Smin)^{\otimes m}}\left(\forall i, f(\Smin,\Kbold_i,W_i) \geq \Delta_2, \mathcal{E}_m^c \right)+ e^{-m(\log(2/\delta)-\nu)}\\
    \leq& \mathbb{P}_{\Kbold^{\otimes m}}\left(\exists w^m \colon \forall i, f(\Smin,\Kbold_i,w_i) \geq \Delta_2, \mathcal{E}_m^c \right)+ e^{-m(\log(2/\delta)-\nu)}\\
    \leq&  \mathbb{P}_{\Kbold^{\otimes m}}\left(\exists j \in [l_m(\Smin)], \{\Delta_i\}_{i=1}^m\in \mathbb{R} \colon \forall i,  f(\Smin,\Kbold_i,\hat{w}_{j,i}) \geq \Delta_2-\Delta_i ,\sum_{i=1}^m \Delta_i \leq m\epsilon(\Smin) \right)+ e^{-m(\log(2/\delta)-\nu)}\\
    \leq&  \mathbb{P}_{\Kbold^{\otimes m}}\left(\exists j \in [l_m(\Smin)], \{\Delta_i\}_{i=1}^m\in \mathbb{R} \colon   \sum_{i=1}^m f(\Smin,\Kbold_i,\hat{w}_{j,i}) \geq \sum_{i=1}^m \left(\Delta_2-\Delta_i\right) ,\sum_{i=1}^m \Delta_i \leq m\epsilon(\Smin) \right)+ e^{-m(\log(2/\delta)-\nu)}\\
    \leq &  \mathbb{P}_{\Kbold^{\otimes m}}\left(\exists j \in [l_m(\Smin)] \colon   \sum_{i=1}^m f(\Smin,\Kbold_i,\hat{w}_{j,i}) \geq m\left(\Delta_2-\epsilon(\Smin)\right)  \right)+ e^{-m(\log(2/\delta)-\nu)}\\
    \leq & \sum \limits_{j \in [l_m(\Smin)]} \mathbb{P}_{\Kbold^{\otimes m}}\left(\sum_{i=1}^m f(\Smin,\Kbold_i,\hat{w}_{j,i}) \geq m\left(\Delta_2-\epsilon(\Smin)\right)  \right)+ e^{-m(\log(2/\delta)-\nu)}\\
    \stackrel{(a)}{\leq} & \sum \limits_{j \in [l_m(\Smin)]} e^{-mn(\Delta_2-\epsilon(\Smin))^2/2}+ e^{-m(\log(2/\delta)-\nu)}\\
    \stackrel{(b)}{\leq} & e^{m(R(\Smin)-n(\Delta_2-\epsilon(\Smin))^2/2)}+ e^{-m(\log(2/\delta)-\nu)}\\
    \stackrel{(c)}{\leq} & 2 e^{-m(\log(2/\delta)-\nu)},
\end{align*}
where $(a)$ is derived using the 
Hoeffding's inequality, $(b)$ is derived since $l_m(\Smin) \leq e^{mR(\Smin)}$, and $(c)$  is derived since $\Delta_2\geq \sqrt{\frac{2(R(\Smin)+\log(2/\delta))}{n}}+\epsilon(\Smin)$. The proof completes by taking the $m$'th root of both sides, and since $\nu$ can be chosen arbitrarily small.
\end{proof}


\subsection{Proof of Theorem~\ref{th:rdCompDistCond}}\label{pr:rdCompDistCond}
\begin{proof}
First, similar to the proof of Theorem~\ref{th:compressibilityCond}, we have
\begin{align*}
   \mathbb{P}\left(\gen(S,W) \geq \Delta_1+\Delta_2 \right)   \leq &\delta/2+ \max_{\Smin} \mathbb{P}\left(f(\Smin,\Kbold,W) \geq  \Delta_2 \right).
\end{align*}
where $\Delta_1\coloneq \sqrt{\log(2/\delta)/n}$, $\Delta_2\coloneq \sup_{\Smin}\sqrt{2(R(\Smin,\delta,\epsilon)+\log\left(2/\delta\right))/n}+\epsilon(\Smin)$, and $R(\Smin,\delta,\epsilon)$ is defined in the theorem. The rest of the proof is to upper bound $\mathbb{P}_{\Kbold,W|\Smin}\left(f(\Smin,\Kbold,W) \geq  \Delta_2 \right)$ by $\delta/2$ for a fixed $\Smin$, which follows  similarly as the proof of Theorem~\ref{th:rdCompDist}, by considering $f(\Smin,\Kbold,W)$ instead of $\gen(S,W)$.
\end{proof}


\subsection{Proof of Corollary~\ref{cor:VCprob}}\label{pr:VCprob}
\begin{proof}
For any $\Smin$ and under any $Q$ such that $D_{KL}(Q\|P_{\Kbold,W,U|\Smin}) \leq \log\left(2/\delta\right) < \infty$, we have $I(\Kbold,W |u) \leq d \log (2en/d)$. Since, for any fixed $U$, similar to the proof of Corollary~\ref{cor:VCexp}, the set of possible $W$ under $P_{\Kbold,W,U|\Smin}$ is bounded by $(2en/d)^d$, and consequently under $Q$ as well. Using Theorem~\ref{th:rdCompDistCond} completes the proof.
\end{proof}


\subsection{Proof of Lemma~\ref{lemmaag2}}\label{pr:donskerEx}
\begin{proof}
The Donsker-Varadhan's identity states that
\[D_{KL}(\nu\|\mu)=\sup_{\Phi}\left\{ \mathbb{E}_{\nu}[\Phi( X)]-\log\mathbb{E}_{\mu}[e^{ \Phi( X)}]\right\}.
\]
The choice of $\Phi(\hat x)=\lambda\hat x$ implies the following inequality for any distributions $\nu$ and $\mu$:
\[D_{KL}(\nu\|\mu)\geq \lambda \mathbb{E}_\nu[ X]-\log\mathbb{E}_{\mu}[e^{\lambda X}].
\]
Therefore,
\[\inf_{\mu}\left[D_{KL}(\nu\|\mu)+\log\mathbb{E}_{\mu}[e^{\lambda  X}]\right]\geq \lambda \mathbb{E}_\nu[X].
\]
On the other hand, if $\frac{d\mu}{d\nu}$ is proportional to $e^{-\lambda x}$, 
one can directly verify that 
\[D_{KL}(\nu\|\mu)+\log\mathbb{E}_{\mu}[e^{\lambda  X}]=\lambda \mathbb{E}_\nu[X].
\]
Thus, the desired inequality is established.

\end{proof}


\subsection{Proof of Lemma~\ref{lem:variationalTail}}
We state two proofs for this lemma.

\subsubsection{First Proof}\label{pr:variationalTail}
\begin{proof}
We simplify the right hand side and reduce it to the left hand side.
For any $\nu_X$, if there exists a  distribution $p_{\hat X}\in\mathcal{P}(\nu_X)$ such that $(\Delta-\epsilon)> \mathbb{E}_{p}[\hat X]$, then one can set $\lambda=\infty$. Otherwise, it is optimal to set $\lambda=0$. Let $\mathcal{A}$ denote the set of distributions $\nu_X$ such that for any $p_{\hat X}\in\mathcal{P}(\nu_X)$ we have $(\Delta-\epsilon)\leq \mathbb{E}_{p}[\hat X]$. 
Then, the desired equality is equivalent with 
\begin{align*}\log\mathbb{P}_{X\sim\mu_X}\left(X\geq \Delta\right)
&= \sup_{\nu_{X} \in \mathcal{A}}-D_{KL}(\nu_{X}\|\mu_{X}).
\end{align*}
Remember that
$\mathcal{P}(\nu_X)$ denotes the set of distributions $p_{\hat X}$ on $\mathbb{R}$ for which
\[\left[\inf_{x\in \Supp(\nu_X)}x\right]-\mathbb{E}\left[\hat {X}\right]\leq \epsilon.
\]
Thus,  $\mathcal{A}$ is the set of distributions $\nu_X$ such that  $\Delta\leq \inf \limits _{x\in\Supp(\nu_X)} x$, or equivalently, $\Supp(\nu_X)\subseteq [\Delta,\infty)$. The minimum of $D_{KL}(\nu_{X}\|\mu_{X})$ is then obtained by a distribution that is proportional with $\mu_X$ on $[\Delta,\infty)$, and the minimum value of $D_{KL}(\nu_{X}\|\mu_{X})$ equals $-\log\mathbb{P}_{X\sim \mu_X}\left(X\geq \Delta\right)$. This completes the proof.
\end{proof}


\subsubsection{Second Proof}\label{pr:variationalTail2}
\begin{proof}
In this part, we give a second proof of Lemma~\ref{lem:variationalTail} from Appendix~\ref{sec:tailX}. Fix some natural number $m$. The  \emph{type} of a given sequence in $\mathcal{X}^m$ is defined as its empirical distribution.
For every type $\nu_{X}$ of the sequences in $\mathcal{X}^m$ of length $m$, pick an arbitrary \emph{type}  $p_\nu(\hat x)$ on a set $\hat{\mathcal{X}}$ satisfying
\begin{align}
\left[\min_{x\colon \nu(x)>0}x\right]-\mathbb{E}_{p_\nu}\left[\hat {X}\right]\leq \epsilon.\label{eqnEppi1}
\end{align}
Let
$q_\nu(\hat x)$ be another distribution such that 
\[q_\nu(\hat x)=\frac{p_\nu(\hat x)e^{-\lambda \hat x}}{\mathbb{E}_{p_\nu}e^{-\lambda \hat X}}, \qquad\forall \hat x.\] Equivalently,
\[p_\nu(\hat x)=\frac{q_\nu(\hat x)e^{\lambda \hat x}}{\mathbb{E}_{q_\nu} e^{\lambda \hat X}}, \qquad\forall \hat x.
\]
Then, one can directly verify that
\begin{align}D_{KL}(p_\nu(\hat x)\|q_\nu(\hat x))=\lambda \mathbb{E}_{p_\nu}[\hat X]-\log\mathbb{E}_{q_\nu}[e^{\lambda \hat X}].\label{eq:donsEq}\end{align}

Let $X^m=(X_1, X_2, \ldots, X_m)$ be $m$ i.i.d.\ repetitions from the distribution $p_X$.

Take some $\zeta>0$. Let
$k=\max_{\nu}e^{m (\zeta+D_{KL}(p_\nu(\hat x)\|q_\nu(\hat x)))}$
where the maximum is over all possible types $\nu$ (of sequences in $\mathcal{X}^m$). 
We now define $\hat{X}^m(1), \ldots, \hat{X}^m(k)$ jointly distributed with $X^m$. Given some $x^m$, we define the conditional distribution of $\hat{X}^m(1), \ldots, \hat{X}^m(k)$ given $x^m$ as follows. Let $\nu(x)$ be the empirical type of the sequence $x^m$. For $j\leq e^{m (\zeta+D_{KL}(p_\nu(\hat x)\|q_\nu(\hat x)))}$, generate the sequences $\hat X^m(j)$ independently and 
i.i.d.\ from the distribution $q_\nu(\hat x)$. For $j> e^{m (\zeta+D_{KL}(p_\nu(\hat x)\|q_\nu(\hat x)))}$, the sequences $\hat X^m(j)$ are all zero. 

From Theorem \ref{th:tailTwoTerms} we get the following tail bound:
\begin{align}
\mathbb{P}\left(X\geq \Delta\right)^m
&\leq \mathbb{P}\left(\exists j:\frac1m \sum_i\hat X_i(j)\geq \Delta- \epsilon\right)
+\mathbb{P}\left(\forall j\in[k]: \rho(X^m,{\hat X}^m(j))>  \epsilon \right).\label{eqnABC}
\end{align}
We have
\[\mathbb{P}\left(\forall j\in[k]: \rho(X^m,{\hat X}^m(j))>  \epsilon \right)=\sum_{x^m}p(x^m)\mathbb{P}\left(\forall j\in[k]: \rho(X^m,{\hat X}^m(j))>  \epsilon\bigg|X^m=x^m\right).
\]
Fix some $X^m=x^m$ with a type $\nu(x)$. 
Observe that if for some $j$, the sequence $\hat X^m(j)$ has type $p_\nu(\hat x)$ then $\rho(X^m,{\hat X}^m(j))\leq \epsilon$. This follows from the definition of $p_{\nu}$ in \eqref{eqnEppi1}. Therefore, the probability $\mathbb{P}\left(\forall j\in[k]: \rho(X^m,{\hat X}^m(j))>  \epsilon\bigg|X^m=x^m\right)$ is less than or equal to the probability that there is no $j$ such that the sequence $\hat X^m(j)$ has type  $p_{\nu}(\hat x)$.

We now compute the probability that there is some $j\leq e^{m (\zeta+D_{KL}(p_\nu(\hat x)\|q_\nu(\hat x)))}$ such that the sequence $\hat X^m(j)$ has type  $p_{\nu}(\hat x)$.
The probability that each sequence has type $p_{\nu}(\hat x)$ is greater than or equal to \cite[Lemma 2.6]{Csiszar1995}
$$\alpha=(m+1)^{-|\mathcal{\hat X}|}e^{-mD_{KL}(p_\nu(\hat x)\|q_\nu(\hat x))}.$$
The probability that there is no $j\leq e^{m (\zeta+D_{KL}(p_\nu(\hat x)\|q_\nu(\hat x)))}$
such that the sequence $\hat X^m(j)$ has type  $p(\hat x)$ equals
\[(1-\alpha)^{e^{m (\zeta+D_{KL}(p_\nu(\hat x)\|q_\nu(\hat x)))}}\leq \exp(-\alpha e^{m (\zeta+D_{KL}(p_\nu(\hat x)\|q_\nu(\hat x)))})=\exp(-(m+1)^{-|\mathcal{\hat X}|}e^{m\zeta})\]
where we used the inequality $(1-x)^m\leq \exp(-mx)$. Since this upper bound does not depend on our choice of $x^m$, we get
\[\mathbb{P}\left(\forall j\in[k]: \rho(X^m,{\hat X}^m(j))>  \epsilon \right)\leq \exp(-(m+1)^{-|\mathcal{\hat X}|}e^{m\zeta}).
\]
Next, note that
\[
\mathbb{P}\left[\exists j:\frac1m \sum_i\hat X_i(j)\geq \Delta- \epsilon\right]=
\sum_{x^m}p(x^m)
\mathbb{P}\left[\exists j:\frac1m \sum_i\hat X_i(j)\geq \Delta- \epsilon\bigg|x^m\right].
\]
For every $\epsilon<\Delta$, we can obtain an upper bound using the union bound as follows:
\begin{align*}
    \mathbb{P}\left[\exists j:\frac1m \sum_i\hat X_i(j)\geq \Delta- \epsilon\bigg|x^m\right]&
    = \mathbb{P}\left[\exists j\leq e^{m (\zeta+D_{KL}(p_\nu(\hat x)\|q_\nu(\hat x)))}:\frac1m \sum_i\hat X_i(j)\geq \Delta- \epsilon\bigg|x^m\right]
    \\&\leq \sum_{j=1}^{e^{m (\zeta+D_{KL}(p_\nu(\hat x)\|q_\nu(\hat x)))}}\mathbb{P}\left[\frac1m \sum_i\hat X_i(j)\geq \Delta- \epsilon\bigg|x^m\right].
\end{align*}

Take some $j\leq e^{m (\zeta+D_{KL}(p_\nu(\hat x)\|q_\nu(\hat x)))}$. 
Chernoff's bound implies that
\begin{align*}
\mathbb{P}\left[\frac1m \sum_i\hat X_i(j)\geq \Delta- \epsilon\bigg|x^m\right]
&\leq\exp(-m\lambda(\Delta-\epsilon))\prod_{i=1}^m\mathbb{E}_{q_\nu}[\exp(\lambda \hat X_i(j))|x^m]
\\&=\exp(-m\lambda(\Delta-\epsilon)+m\log\mathbb{E}_{q_\nu}[\exp(\lambda \hat X)]).
\end{align*}
We obtain
\begin{align*}
    \mathbb{P}\left[\exists j:\frac1m \sum_i\hat X_i(j)\geq \Delta- \epsilon\bigg|x^m\right]&
    \leq 
    \exp(m (\zeta+D_{KL}(p_\nu(\hat x)\|q_\nu(\hat x)))-m\lambda(\Delta-\epsilon)+m\log\mathbb{E}_{q_\nu}[\exp(\lambda \hat X)])\\
    &
    =
    \exp\left[m \zeta-m\lambda(\Delta-\epsilon)+m\lambda\mathbb{E}_{p_\nu}(\hat X)\right]
\end{align*}
where we used \eqref{eq:donsEq} in the last step. Another  trivial upper bound is 
\[\mathbb{P}\left[\exists j:\frac1m \sum_i\hat X_i(j)\geq \Delta- \epsilon\bigg|x^m\right]\leq 1.\]
Thus,
\begin{align*}
    \mathbb{P}\left[\exists j:\frac1m \sum_i\hat X_i(j)\geq \Delta- \epsilon\bigg|x^m\right]
    &
    \leq
    \exp\left\{-m \bigg[-\zeta+\lambda(\Delta-\epsilon)-\lambda\mathbb{E}_{p_\nu}(\hat X)\bigg]_+\right\}.
\end{align*}
The above bound depends only on the type $\nu$ and not on the exact sequence $x^m$. If we denote $\mathcal{T}_\nu$ the set of sequences $x^m$ with type $\nu$, we have \cite[Lemma 2.6]{Csiszar1995}
\[\sum_{x^m\in\mathcal{T}_\nu}p(x^m)\leq \exp\left(-mD_{KL}(\nu_X\|p_X)\right).
\]
Thus,
\begin{align*}
    \sum_{x^m\in\mathcal{T}_\nu}p(x^m)\mathbb{P}&\left[\exists j:\frac1m \sum_i\hat X_i(j)\geq \Delta- \epsilon\bigg|x^m\right]
    \\&
    \leq
    \exp\left\{-mD_{KL}(\nu_X\|p_X)-m \bigg[-\zeta+\lambda(\Delta-\epsilon)-\lambda\mathbb{E}_{p_\nu}(\hat X)\bigg]_+\right\}.
\end{align*}
Therefore,
\begin{align*}
\mathbb{P}&\left[\exists j:\frac1m \sum_i\hat X_i(j)\geq \Delta- \epsilon\right]
\\&\leq \sum_{\nu}
    \exp\left\{-mD_{KL}(\nu_X\|p_X)-m \bigg[-\zeta+\lambda(\Delta-\epsilon)-\lambda\mathbb{E}_{p_\nu}(\hat X)\bigg]_+\right\}
    \\
    &\leq (m+1)^{|\mathcal{X}|}\max_{\nu}
    \exp\left\{-mD_{KL}(\nu_X\|p_X)-m \bigg[-\zeta+\lambda(\Delta-\epsilon)-\lambda\mathbb{E}_{p_\nu}(\hat X)\bigg]_+\right\}.
\end{align*}
From \eqref{eqnABC} we get
\begin{align*}
\mathbb{P}\left(X\geq \Delta\right)^m
&\leq \mathbb{P}\left[\exists j:\frac1m \sum_i\hat X_i(j)\geq \Delta- \epsilon\right]
+\mathbb{P}\left(\forall j\in[k]: \rho(X^m,{\hat X}^m(j))>  \epsilon \right)
\\
&\leq 
(m+1)^{|\mathcal{X}|}\max_{\nu}
    \exp\left\{-mD_{KL}(\nu_X\|p_X)-m \bigg[-\zeta+\lambda(\Delta-\epsilon)-\lambda\mathbb{E}_{p_\nu}(\hat X)\bigg]_+\right\}
    \\&\quad+\exp\left(-(m+1)^{-|\mathcal{\hat X}|}e^{m\zeta}\right).
\end{align*}
Raising both sides of the inequality to the power $1/m$ and letting $m$ tend to infinity yields
\begin{align*}
\mathbb{P}\left(X\geq \Delta\right)
&\leq 
\max_{\nu}
    \exp\left\{-D_{KL}(\nu_X\|p_X)- \bigg[-\zeta+\lambda(\Delta-\epsilon)-\lambda\mathbb{E}_{p_\nu}(\hat X)\bigg]_+\right\}.
\end{align*}
Letting $m$ tend to infinity, we obtain the above inequality for any arbitrary
 $p_\nu(\hat x)$ in $\mathcal{P}(\nu_X)$.
Letting $\zeta$ tend to zero yields the desired result.
\end{proof}


\subsection{Proof of Corollary~\ref{cor:LipschExamples}}\label{sec:LipschExamples}
\begin{proof}
\begin{itemize}
    \item[i.] Let $\hat{W}$ be uniformly distributed over the $d$-dimensional ball with radius $\epsilon<R$ with center $W$. Let $V_d$ denote the volume of the unitary $d$-dimensional ball. Then,
    \begin{align*}
        I(W;\hat{W})=&H(\hat{W})-H(\hat{W}|W)\\
        =&H(\hat{W})-\log(\epsilon^d V_d)\\
        \leq&\log((r_0+\epsilon)^d V_d)-\log(\epsilon^d V_d)\\
        =&d\log((r_0+\epsilon)/\epsilon)\\
        \leq&d\log(2r_0/\epsilon).
    \end{align*}
    Now, using Corollary~\ref{cor:lipschitz}, we derive with probability at least $1-\delta$,
    \begin{align*}
        \gen(S,W) \leq \sqrt{\frac{2\sigma^2(d\log(2r_0/\epsilon)+\log(1/\delta))}{n}}+2\mathfrak{L} \epsilon.
    \end{align*}
    For $n \geq 16$, by letting $\epsilon \coloneq2r_0 \sqrt{d\log(n)/n}$ and $\delta\coloneq e^{-d/2}$, we derive that for $n \geq 16$, with probability at least $1-e^{-d/2}$,
    \begin{align*}
        \gen(S,W) \leq  (4r_0\mathfrak{L}+\sigma\sqrt{d})\sqrt{\log(n)/n}.
    \end{align*}
    \item[ii.] The result follows from \cite[Example~1]{marton1974} and Corollary~\ref{cor:lipschitz}.
    \item[iii.]The result follows from \cite[Example~1]{bakshi2005error} and Corollary~\ref{cor:lipschitz}.
    \item[iv.] The result follows from \cite[Theorem~2]{Ihara2000} and Corollary~\ref{cor:lipschitz}. 
\end{itemize}
\end{proof}


\subsection{Proof of Lemma~\ref{lem:probToExpec}} \label{pr:probToExpec}
\begin{proof}
Here we show the proof for $d_m\coloneq \dem$. The proof is similar for $|\dem|$ and $\deAm$. For simplicity, denote $\mathcal{E}_m\coloneq \mathcal{E}_m(\mathcal{H}_m,\epsilon;\dem)$.  
\begin{align*}
    \mathbb{E}_{(S,W)^{\otimes m}}&\left(\min_{j \in [l_m]} \dem(W^m,\hat{\vc{w}}_j;S^m)\right)\\
    =&  \mathbb{E}_{(S,W)^{\otimes m}}\left(\min_{j \in [l_m]} \dem(W^m,\hat{\vc{w}}_j;S^m)|\mathcal{E}_m\right)\mathbb{P}_{(S,W)^{\otimes m}}\left(\mathcal{E}_m\right)\\
    &+  \mathbb{E}_{(S,W)^{\otimes m}}\left(\min_{j \in [l_m]} \dem(W^m,\hat{\vc{w}}_j;S^m)|\mathcal{E}_m^c\right)\mathbb{P}_{(S,W)^{\otimes m}}\left(\mathcal{E}_m^c\right)\\
    \leq &  \mathbb{E}_{(S,W)^{\otimes m}}\left(\min_{j \in [l_m]} \dem(W^m,\hat{\vc{w}}_j;S^m)|\mathcal{E}_m\right)\mathbb{P}_{(S,W)^{\otimes m}}\left(\mathcal{E}_m\right)+   \epsilon\\
    \leq &  \mathbb{E}_{(S,W)^{\otimes m}}\left( \dem(W^m,\hat{\vc{w}}_1;S^m)|\mathcal{E}_m\right)\mathbb{P}_{(S,W)^{\otimes m}}\left(\mathcal{E}_m\right)+   \epsilon\\
    \leq &  \frac{1}{m} \mathbb{E}_{(S,W)^{\otimes m}}\left(| \sum \limits_{i=1}^m \gen(S_i,W_i) |+| \sum \limits_{i=1}^m \gen(S_i,\hat{w}_{1,i}) | \Big|\mathcal{E}_m\right)\mathbb{P}_{(S,W)^{\otimes m}}\left(\mathcal{E}_m\right)+   \epsilon\\
    \stackrel{(a)}{\leq} &  \frac{1}{m} \mathbb{E}_{(S,W)^{\otimes m}}\left(| \sum \limits_{i=1}^m \gen(S_i,W_i)| \Big|\mathcal{E}_m\right)\mathbb{P}_{(S,W)^{\otimes m}}\left(\mathcal{E}_m\right)+\sqrt{2\sigma^2\log(2)/(n m)}+  \epsilon\\
    \stackrel{(b)}{\leq} &    \varepsilon_m+\epsilon,
\end{align*}
where $\lim_{m \ra \infty}\varepsilon_m=0$, $(a)$ is due to a known maximal inequality for subgaussian random variable \cite[Theorem~2.5]{boucheron2013concentration} and $(b)$ is derived due to the Lemma~\ref{lem:integrable}, stated in the following. Taking the limit for $m \ra \infty$ completes the proof.

To show step $(b)$, we state the following lemma, shown within the proof of \cite[Theorem~7.2.2]{Berger1975}. Here, for the sake of completeness, we state the adapted proof to our setup in Section~\ref{pr:integrable}.
\begin{lemma} \label{lem:integrable}
Assume that $\mathbb{E}_{X}\left[|X|\right]< \infty$, where $X \in \mathcal{X}$. Let $\{\mathcal{A}_m\}_{m\in \mathbb{N}}\colon \mathcal{A}_m \subseteq \mathcal{X}^m$, be a sequence of sets such that $\lim_{m \ra \infty}\mathbb{P}_{X^{\otimes m}}\left(X^m \in \mathcal{A}_m\right)=0$. Then, 
\begin{align}
\lim_{m \ra \infty}    \frac{1}{m}\int_{\mathcal{A}_m} (\sum_{i=1}^m \left|X_i\right|)~\mathrm{d}P_{X^{\otimes m}} =0.
\end{align}
\end{lemma}

\end{proof}


\subsection{Proof of Lemma~\ref{lem:typeCoveringExt}}\label{pr:typeCoveringExt}
\begin{proof}
Denote that the marginal type of $Q_m$ with respect to $\mathcal{S}^m$ as $Q_{s,m}$. Consider a random variable $\hat{W}$, defined by the conditional distribution $P_{\hat{W}|S}$, such that $\mathbb{E}\left[  d_{Q_m}(\hat{W};S)\right]\leq \epsilon$, where the expectation is with respect to joint distribution $Q_{S,\hat{W}}\coloneq Q_{s,m} \times P_{\hat{W}|S}$. 

Following the proof of \cite[Lemma~9.1]{CsisKor82}, we can find a set $\mathcal{H}_{Q_m,\hat{W}}=\{\vc{\hat{w}}_{Q_m,j}, j \in [l_{Q_m,\hat{W}}]\} \in \mathcal{\hat{W}}^m$, such that 
\begin{align*}
 l_{Q_m,\hat{W}} \leq  e^{m (I(S;\hat{W})+\varepsilon'_m)},   
\end{align*}
where the mutual information is with respect to the joint distribution $Q_{S,\hat{W}}$, and such that for each $S^m$ having the type $Q_{s,m}$, there exists a $j(S^m) \in [l_{Q_m,\hat{W}}]$, such that 
\begin{align}
    \|\hat{P}_{S^m,\vc{\hat{w}}_{Q_m,j(S^m)}}(s,\hat{w})-Q_{S,\hat{W}}(s,\hat{w}) \|_{TV}=\varepsilon''_m,
\end{align}
where $\varepsilon''_m$ vanishes as $m \ra \infty$. Note that by Carathéodory's theorem, we can assume that $\mathcal{\hat{W}}$ is a finite set as well and hence $d_{Q_m}(\hat{w};s)$ is always bounded. This yields for the picked $j(S^m)$, satisfying the above equation, we have
\begin{align}
    \sum\limits_{i=1}^m d_{Q_m}(\hat{w}_{Q_m,j(S^m),i};S_i) \leq m(\epsilon+\varepsilon_m),
\end{align}
where $\varepsilon_m$ vanishes as $m \ra \infty$. Now, 
\begin{align*}  \mathbb{P}_{(S,W)^{\otimes m}}&\left(\mathcal{T}(S^m,W^m)=Q_m ,\min_{j \in  [l_{Q_m,\hat{W}}]} \dpm\left(W^m,\vc{\hat{w}}_{Q_m,j};S^m\right)\geq \epsilon+\nu \right)\\
= &\mathbb{P}_{(S,W)^{\otimes m}} \left(\mathcal{T}(S^m,W^m)=Q_m ,\min_{j \in  [l_{Q_m,\hat{W}}]} \sum\limits_{i=1}^m \min \limits_{k\in [m]} \gen(S_k,W_k)-\gen(S_i,\hat{w}_{Q_m,j,i}) \geq m(\epsilon+\nu) \right)\\
= &\mathbb{P}_{(S,W)^{\otimes m}} \left(\mathcal{T}(S^m,W^m)=Q_m ,\min_{j \in  [l_{Q_m,\hat{W}}]} \sum\limits_{i=1}^m \min \limits_{\substack{(s',w')\colon\\ Q_m(s',w')>0}} \gen(s',w')-\gen(S_i,\hat{w}_{Q_m,j,i}) \geq m(\epsilon+\nu) \right)\\
= &\mathbb{P}_{(S,W)^{\otimes m}} \left(\mathcal{T}(S^m,W^m)=Q_m ,\min_{j \in  [l_{Q_m,\hat{W}}]} \sum\limits_{i=1}^m d_{Q_m}(\hat{w}_{Q_m,j,i};S_i)  \geq m(\epsilon+\nu) \right)\\
\stackrel{(a)}{\leq} & \mathbb{P}_{(S,W)^{\otimes m}}\left(\mathcal{T}(S^m,W^m)=Q_m ,m(\epsilon+\varepsilon_m)\geq m(\epsilon+\nu) \right)\\
=& 0,
\end{align*}
where the last step holds for $m \geq m_{\nu,Q_m}$, where $m_{\nu,Q_m}$ is a sufficiently large integer.

The required set $\mathcal{H}_{Q_m}$ would be equal to the $\mathcal{H}_{Q_m,\hat{W}}$ having the minimum cardinality number $l_{Q_m,\hat{W}}$. This completes the proof.
\end{proof}


\subsection{Proof of Lemma~\ref{lem:integrable}} \label{pr:integrable}
\begin{proof}
Let $\eta \coloneq\mathbb{E}_{X}\left[|X|\right]$ and $s_m(x^m)\coloneq \frac{1}{m} \sum_{i=1}^m \left|x_i\right|$. Define the following sets
\begin{align*}
    \mathcal{B}_m &\coloneq \{x^m \colon s_m(x^m) > \eta+\delta \},\\
    \mathcal{C}_m &\coloneq\{x^m \colon s_m(x^m) < \eta-\delta \}.
\end{align*}
Fix a $\delta>0$. Then,
\begin{align}
     \int_{\mathcal{A}_m} S_m(X^m)\mathrm{d}P_{X^{\otimes m}}&=\int_{\mathcal{A}_m} (S_m(X^m)-\eta)\mathrm{d}P_{X^{\otimes m}}+\eta \int_{\mathcal{A}_m} \mathrm{d}P_{X^{\otimes m}}\nonumber\\
     &=\int_{\mathcal{A}_m} (S_m(X^m)-\eta)\mathrm{d}P_{X^{\otimes m}}+\eta \epsilon'_m\nonumber\\
     &=\int_{\mathcal{A}_m \bigcap \mathcal{B}_m} (S_m(X^m)-\eta)\mathrm{d}P_{X^{\otimes m}}+\int_{\mathcal{A}_m \bigcap \mathcal{B}_m^c} (S_m(X^m)-\eta)\mathrm{d}P_{X^{\otimes m}}+\eta \epsilon'_m\nonumber\\
     &=\int_{\mathcal{A}_m \bigcap \mathcal{B}_m} (S_m(X^m)-\eta)\mathrm{d}P_{X^{\otimes m}}+\delta\int_{\mathcal{A}_m \bigcap \mathcal{B}_m^c} \mathrm{d}P_{X^{\otimes m}}+\eta \epsilon'_m\nonumber\\
     &\leq \int_{\mathcal{A}_m \bigcap \mathcal{B}_m} (S_m(X^m)-\eta)\mathrm{d}P_{X^{\otimes m}}+ \epsilon_m \nonumber \\
     &\stackrel{(a)}{\leq }\int_{\mathcal{B}_m} (S_m(X^m)-\eta)\mathrm{d}P_{X^{\otimes m}}+ \epsilon_m. \label{eq:integrable1}
\end{align}
where $\epsilon_m$ vanishes as $m \ra \infty$ and $(a)$ is derived since for $X^m \in \mathcal{B}_m$, $S_m(X^m)-\eta$ is positive.  
Next,
\begin{align}
    \int_{\mathcal{B}_m} (S_m(X^m)-\eta)\mathrm{d}P_{X^{\otimes m}} &= \int (S_m(X^m)-\eta)\mathrm{d}P_{X^{\otimes m}}-\int_{\mathcal{B}_m^c} (S_m(X^m)-\eta)\mathrm{d}P_{X^{\otimes m}}\nonumber\\
    &\stackrel{(a)}{=}
    \int_{\mathcal{B}_m^c} (\eta-S_m(X^m))\mathrm{d}P_{X^{\otimes m}}\nonumber\\
    &=
    \int_{\mathcal{B}_m^c\bigcap \mathcal{C}_m} (\eta-S_m(X^m))\mathrm{d}P_{X^{\otimes m}}+
    \int_{\mathcal{B}_m^c\bigcap \mathcal{C}_m^c} (\eta-S_m(X^m))\mathrm{d}P_{X^{\otimes m}}\nonumber\\
    & \leq 
    \int_{\mathcal{B}_m^c\bigcap \mathcal{C}_m} (\eta-S_m(X^m))\mathrm{d}P_{X^{\otimes m}}+\delta\nonumber\\
    &\stackrel{(b)}{\leq} 
    \eta \int_{\mathcal{B}_m^c\bigcap \mathcal{C}_m} \mathrm{d}P_{X^{\otimes m}}+\delta\nonumber\\
    &= 
    \eta \int_{\mathcal{C}_m} \mathrm{d}P_{X^{\otimes m}}+\delta\nonumber\\
    &\stackrel{(c)}{=}
    \epsilon''_m+\delta, \label{eq:integrable2}
\end{align}
where $ \epsilon''_m$ vanishes as $m \ra \infty$, $(a)$ is derived since $\mathbb{E}[S_m(X^m)]=\eta$, $(b)$ is derived since $S_m(X^m)$ is non-negative, and $(c)$ is derived since $\mathbb{P}\left(|S_m-\eta|>\delta \right)$ asymptotically vanishes by law of large numbers.

The inequalities \eqref{eq:integrable1} and \eqref{eq:integrable2} yield for any $\delta>0$,
\begin{align}
\lim_{m \ra \infty}    \frac{1}{m}\int_{\mathcal{A}_m} (\sum_{i=1}^m \left|X_i\right|)~\mathrm{d}P_{X^{\otimes m}} <\delta.
\end{align}
This completes the proof.
\end{proof}

\end{document}

%% file: defs.tex
\usepackage[utf8]{inputenc}

\usepackage{bm,mathtools,amsfonts,amssymb,bbm,float,hyperref,euscript,setspace,tikz,steinmetz,enumitem,cancel}
\usepackage{times}

\usepackage[normalem]{ulem}
\usepackage{xurl}
\usepackage{breakcites}
\usepackage{breakurl}

\usepackage[mode=buildnew]{standalone}
\usepackage{mathabx} 

\usepackage{mleftright}\mleftright

\definecolor{darkgreen}{rgb}{0, 0.5, 0}
\definecolor{darkred}{RGB}{128, 0, 0}

\newcommand{\vc}[1]{\mathbf{#1}} 

\newcommand{\ra}{\rightarrow}

\newcommand{\dd}{d}
\newcommand{\ddm}{d_m}
\newcommand{\dem}{\vartheta_m}
\newcommand{\deAm}{\xi_m}
\newcommand{\dpm}{\varphi_m}
\newcommand{\dw}{\varrho}

\newcommand{\dsm}{\vartheta_m}
\newcommand{\dsAm}{\xi_m}
\newcommand{\dspm}{\varphi_m}

\newcommand{\Scap}{\mathfrak{Z}}
\newcommand{\Smin}{\mathfrak{z}}

\newcommand{\Kbold}{\mathbf{K}}
\newcommand{\Knorm}{K}
\newcommand{\kbold}{\mathbf{k}}
\newcommand{\knorm}{k}

\newcommand{\Supp}{\supp}

\DeclareMathOperator*{\argmin}{arg\min}

\DeclareMathOperator*{\gen}{gen}

\DeclareMathOperator\supp{supp}
\DeclareMathOperator\leb{Leb}

\newtheorem*{rep@theorem}{\rep@title}
\newcommand{\newreptheorem}[2]{%
\newenvironment{rep#1}[1]{%
 \def\rep@title{#2 \ref{##1}}%
 \begin{rep@theorem}}%
 {\end{rep@theorem}}}
\newreptheorem{theorem}{Theorem}

\newtheorem*{rep@definition}{\rep@title}
\newcommand{\newrepdefinition}[2]{%
\newenvironment{rep#1}[1]{%
 \def\rep@title{#2 \ref{##1}}%
 \begin{rep@definition}}%
 {\end{rep@definition}}}
\newrepdefinition{definition}{Definition}

\newcommand{\stkout}[1]{{\color{teal}\ifmmode\text{\sout{\ensuremath{#1}}}\else\sout{#1}\fi}}

%% file: Colt2022_Final.bbl
\begin{thebibliography}{57}
\providecommand{\natexlab}[1]{#1}
\providecommand{\url}[1]{\texttt{#1}}
\expandafter\ifx\csname urlstyle\endcsname\relax
  \providecommand{\doi}[1]{doi: #1}\else
  \providecommand{\doi}{doi: \begingroup \urlstyle{rm}\Url}\fi

\bibitem[Anthony and Bartlett(1999)]{anthonyBartlett99}
Martin Anthony and Peter~L. Bartlett.
\newblock \emph{Neural Network Learning: Theoretical Foundations}.
\newblock Cambridge University Press, 1999.

\bibitem[Arimoto(1972)]{Arimoto72}
Suguru Arimoto.
\newblock An algorithm for computing the capacity of arbitrary discrete
  memoryless channels.
\newblock \emph{IEEE Transactions on Information Theory}, 18\penalty0
  (1):\penalty0 14--20, 1972.

\bibitem[Arora et~al.(2018)Arora, Ge, Neyshabur, and Zhang]{arora2018stronger}
Sanjeev Arora, Rong Ge, Behnam Neyshabur, and Yi~Zhang.
\newblock Stronger generalization bounds for deep nets via a compression
  approach.
\newblock In \emph{Proceedings of the 35th International Conference on Machine
  Learning}, volume~80, pages 254--263. PMLR, 10--15 Jul 2018.

\bibitem[Bakshi and Bansal(2005)]{bakshi2005error}
Mayank Bakshi and Rakesh~K. Bansal.
\newblock On error exponent in lossy source coding, 2005.

\bibitem[Barsbey et~al.(2021)Barsbey, Sefidgaran, Erdogdu, Richard, and
  {\c{S}}im{\c{s}}ekli]{barsbey2021heavy}
Melih Barsbey, Milad Sefidgaran, Murat~A Erdogdu, Ga{\"e}l Richard, and Umut
  {\c{S}}im{\c{s}}ekli.
\newblock Heavy tails in {SGD} and compressibility of overparametrized neural
  networks.
\newblock In \emph{Thirty-Fifth Conference on Neural Information Processing
  Systems}, 2021.

\bibitem[Baykal et~al.(2019)Baykal, Liebenwein, Gilitschenski, Feldman, and
  Rus]{baykal2018datadependent}
Cenk Baykal, Lucas Liebenwein, Igor Gilitschenski, Dan Feldman, and Daniela
  Rus.
\newblock Data-dependent coresets for compressing neural networks with
  applications to generalization bounds.
\newblock In \emph{International Conference on Learning Representations}, 2019.

\bibitem[Berger(1975)]{Berger1975}
Toby Berger.
\newblock \emph{Rate Distortion Theory and Data Compression}, pages 1--39.
\newblock Springer Vienna, Vienna, 1975.
\newblock ISBN 978-3-7091-2928-9.

\bibitem[Birdal et~al.(2021)Birdal, Lou, Guibas, and
  {\c{S}}im{\c{s}}ekli]{birdal2021intrinsic}
Tolga Birdal, Aaron Lou, Leonidas Guibas, and Umut {\c{S}}im{\c{s}}ekli.
\newblock Intrinsic dimension, persistent homology and generalization in neural
  networks.
\newblock In \emph{Advances in Neural Information Processing Systems
  (NeurIPS)}, 2021.

\bibitem[Blahut(1972)]{Blahut72}
Richard Blahut.
\newblock Computation of channel capacity and rate-distortion functions.
\newblock \emph{IEEE Transactions on Information Theory}, 18\penalty0
  (4):\penalty0 460--473, 1972.

\bibitem[Blum and Langford(2003)]{blum2003pac}
Avrim Blum and John Langford.
\newblock Pac-mdl bounds.
\newblock In \emph{Learning theory and kernel machines}, pages 344--357.
  Springer, 2003.

\bibitem[Boucheron et~al.(2013)Boucheron, Lugosi, and
  Massart]{boucheron2013concentration}
Stéphane Boucheron, Gábor Lugosi, and Pascal Massart.
\newblock \emph{Concentration Inequalities: A Nonasymptotic Theory of
  Independence}.
\newblock OUP Oxford, 2013.

\bibitem[Bu et~al.(2020)Bu, Zou, and Veeravalli]{Bu2020}
Yuheng Bu, Shaofeng Zou, and Venugopal~V. Veeravalli.
\newblock Tightening mutual information-based bounds on generalization error.
\newblock \emph{IEEE Journal on Selected Areas in Information Theory},
  1\penalty0 (1):\penalty0 121–130, May 2020.
\newblock ISSN 2641-8770.

\bibitem[Bu et~al.(2021)Bu, Gao, Zou, and Veeravalli]{Bu2021ModelCompression}
Yuheng Bu, Weihao Gao, Shaofeng Zou, and Venugopal~V. Veeravalli.
\newblock Population risk improvement with model compression: An
  information-theoretic approach.
\newblock \emph{Entropy}, 23\penalty0 (10), 2021.

\bibitem[Camuto et~al.(2021)Camuto, Deligiannidis, Erdogdu, Gürbüzbalaban,
  Şimşekli, and Zhu]{camuto2021fractal}
Alexander Camuto, George Deligiannidis, Murat~A. Erdogdu, Mert Gürbüzbalaban,
  Umut Şimşekli, and Lingjiong Zhu.
\newblock Fractal structure and generalization properties of stochastic
  optimization algorithms, 2021.

\bibitem[Cover and Thomas(2006)]{CoverTho06}
Thomas~M. Cover and Joy~A. Thomas.
\newblock \emph{Elements of information theory (2. ed.)}.
\newblock Wiley, 2006.
\newblock ISBN 978-0-471-24195-9.

\bibitem[Csiszár(1995)]{Csiszar1995}
Imre Csiszár.
\newblock Generalized cutoff rates and renyi's information measures.
\newblock \emph{IEEE Transactions on Information Theory}, 41\penalty0
  (1):\penalty0 26--34, 1995.

\bibitem[Csiszár and Körner(2011)]{CsisKor82}
Imre Csiszár and János Körner.
\newblock \emph{Information Theory: Coding Theorems for Discrete Memoryless
  Systems}.
\newblock Cambridge University Press, 2 edition, 2011.

\bibitem[Cuff et~al.(2010)Cuff, Permuter, and Cover]{Cuff09}
Paul~Warner Cuff, Haim~H. Permuter, and Thomas~M. Cover.
\newblock Coordination capacity.
\newblock \emph{IEEE Transactions on Information Theory}, 56\penalty0
  (9):\penalty0 4181--4206, 2010.

\bibitem[Dziugaite and Roy(2017)]{dziugaite2017computing}
Gintare~Karolina Dziugaite and Daniel~M Roy.
\newblock Computing nonvacuous generalization bounds for deep (stochastic)
  neural networks with many more parameters than training data.
\newblock \emph{arXiv preprint arXiv:1703.11008}, 2017.

\bibitem[El~Gamal and Kim(2011)]{elgamal_kim_2011}
Abbas El~Gamal and Young-Han Kim.
\newblock \emph{Network Information Theory}.
\newblock Cambridge University Press, 2011.

\bibitem[Esposito et~al.(2020)Esposito, Gastpar, and Issa]{esposito2020}
Amedeo~Roberto Esposito, Michael Gastpar, and Ibrahim Issa.
\newblock Generalization error bounds via {R}\'enyi-, $f$-divergences and
  maximal leakage, 2020.

\bibitem[Haghifam et~al.(2020)Haghifam, Negrea, Khisti, Roy, and
  Dziugaite]{haghifam2020sharpened}
Mahdi Haghifam, Jeffrey Negrea, Ashish Khisti, Daniel~M. Roy, and
  Gintare~Karolina Dziugaite.
\newblock Sharpened generalization bounds based on conditional mutual
  information and an application to noisy, iterative algorithms, 2020.

\bibitem[Haghifam et~al.(2021)Haghifam, Dziugaite, Moran, and
  Roy]{haghifam2021towards}
Mahdi Haghifam, Gintare~Karolina Dziugaite, Shay Moran, and Daniel~M. Roy.
\newblock Towards a unified information-theoretic framework for generalization.
\newblock In \emph{Thirty-Fifth Conference on Neural Information Processing
  Systems}, 2021.

\bibitem[Han(2000)]{Han2000}
Te~Sun Han.
\newblock The reliability functions of the general source with fixed-length
  coding.
\newblock \emph{IEEE Transactions on Information Theory}, 46\penalty0
  (6):\penalty0 2117--2132, 2000.

\bibitem[Harutyunyan et~al.(2021)Harutyunyan, Raginsky, Steeg, and
  Galstyan]{harutyunyan2021}
Hrayr Harutyunyan, Maxim Raginsky, Greg~Ver Steeg, and Aram Galstyan.
\newblock Information-theoretic generalization bounds for black-box learning
  algorithms, 2021.

\bibitem[Hellstrom and Durisi(2020)]{Hellstrom2020}
Fredrik Hellstrom and Giuseppe Durisi.
\newblock Generalization bounds via information density and conditional
  information density.
\newblock \emph{IEEE Journal on Selected Areas in Information Theory},
  1\penalty0 (3):\penalty0 824–839, Nov 2020.
\newblock ISSN 2641-8770.

\bibitem[Hodgkinson et~al.(2021)Hodgkinson, Şimşekli, Khanna, and
  Mahoney]{hodgkinson2021generalization}
Liam Hodgkinson, Umut Şimşekli, Rajiv Khanna, and Michael~W. Mahoney.
\newblock Generalization properties of stochastic optimizers via trajectory
  analysis, 2021.

\bibitem[Hsu et~al.(2021)Hsu, Ji, Telgarsky, and Wang]{hsu2021generalization}
Daniel Hsu, Ziwei Ji, Matus Telgarsky, and Lan Wang.
\newblock Generalization bounds via distillation.
\newblock In \emph{International Conference on Learning Representations}, 2021.

\bibitem[Ihara and Kubo(2000)]{Ihara2000}
Shunsuke Ihara and Masashi Kubo.
\newblock Error exponent for coding of memoryless gaussian sources with a
  fidelity criterion.
\newblock \emph{IEICE Trans. Fundamaentals, A}, 83\penalty0 (10):\penalty0
  1891--1897, oct 2000.
\newblock ISSN 09168508.

\bibitem[Iriyama(2005)]{Iriyama2005}
Kiminori Iriyama.
\newblock Probability of error for the fixed-length lossy coding of general
  sources.
\newblock \emph{IEEE Transactions on Information Theory}, 51\penalty0
  (4):\penalty0 1498--1507, 2005.

\bibitem[Kawabata and Dembo(1994)]{Kawabata1994}
Tsutomu Kawabata and Amir Dembo.
\newblock The rate-distortion dimension of sets and measures.
\newblock \emph{IEEE Transactions on Information Theory}, 40\penalty0
  (5):\penalty0 1564--1572, 1994.

\bibitem[Kuhn et~al.(2021)Kuhn, Lyle, Gomez, Rothfuss, and
  Gal]{kuhn2021robustness}
Lorenz Kuhn, Clare Lyle, Aidan~N. Gomez, Jonas Rothfuss, and Yarin Gal.
\newblock Robustness to {{Pruning Predicts Generalization}} in {{Deep Neural
  Networks}}.
\newblock \emph{arXiv:2103.06002 [cs, stat]}, March 2021.

\bibitem[Langford and Caruana(2001)]{langford2001not}
John Langford and Rich Caruana.
\newblock (not) bounding the true error.
\newblock \emph{Advances in Neural Information Processing Systems}, 14, 2001.

\bibitem[Littlestone and Warmuth(1986)]{littlestone1986relating}
Nick Littlestone and Manfred Warmuth.
\newblock Relating data compression and learnability.
\newblock \emph{Citeseer}, 1986.

\bibitem[MacKay(1995)]{mackay1995probable}
David~JC MacKay.
\newblock Probable networks and plausible predictions-a review of practical
  bayesian methods for supervised neural networks.
\newblock \emph{Network: computation in neural systems}, 6\penalty0
  (3):\penalty0 469, 1995.

\bibitem[Marton(1974)]{marton1974}
Katalin Marton.
\newblock Error exponent for source coding with a fidelity criterion.
\newblock \emph{IEEE Transactions on Information Theory}, 20\penalty0
  (2):\penalty0 197--199, 1974.

\bibitem[Masiha et~al.(2021)Masiha, Gohari, Yassaee, and Aref]{masiha2021}
Mohammad~Saeed Masiha, Amin Gohari, Mohammad~Hossein Yassaee, and Mohammad~Reza
  Aref.
\newblock Learning under distribution mismatch and model misspecification.
\newblock In \emph{2021 IEEE International Symposium on Information Theory
  (ISIT)}, pages 2912--2917. IEEE, 2021.

\bibitem[McAllester(1999)]{mcallester1999some}
David~A McAllester.
\newblock Some pac-bayesian theorems.
\newblock \emph{Machine Learning}, 37\penalty0 (3):\penalty0 355--363, 1999.

\bibitem[Negrea et~al.(2020{\natexlab{a}})Negrea, Dziugaite, and
  Roy]{negrea2020defense}
Jeffrey Negrea, Gintare~Karolina Dziugaite, and Daniel Roy.
\newblock In defense of uniform convergence: Generalization via derandomization
  with an application to interpolating predictors.
\newblock In \emph{International Conference on Machine Learning}, pages
  7263--7272. PMLR, 2020{\natexlab{a}}.

\bibitem[Negrea et~al.(2020{\natexlab{b}})Negrea, Haghifam, Dziugaite, Khisti,
  and Roy]{negrea2020it}
Jeffrey Negrea, Mahdi Haghifam, Gintare~Karolina Dziugaite, Ashish Khisti, and
  Daniel~M. Roy.
\newblock Information-theoretic generalization bounds for sgld via
  data-dependent estimates, 2020{\natexlab{b}}.

\bibitem[Neyshabur et~al.(2018)Neyshabur, Bhojanapalli, and
  Srebro]{neyshabur2018pacbayesian}
Behnam Neyshabur, Srinadh Bhojanapalli, and Nathan Srebro.
\newblock A pac-bayesian approach to spectrally-normalized margin bounds for
  neural networks, 2018.

\bibitem[Polyanskiy and Wu(2014)]{polyanskiy2014lecture}
Yury Polyanskiy and Yihong Wu.
\newblock Lecture notes on information theory.
\newblock \emph{Lecture Notes for ECE563 (UIUC) and}, 6\penalty0
  (2012-2016):\penalty0 7, 2014.

\bibitem[Riegler et~al.(2018)Riegler, Bölcskei, and Koliander]{Riegler2018}
Erwin Riegler, Helmut Bölcskei, and Günther Koliander.
\newblock Rate-distortion theory for general sets and measures.
\newblock In \emph{2018 IEEE International Symposium on Information Theory
  (ISIT)}, pages 101--105, 2018.

\bibitem[Russo and Zou(2016)]{russozhou16}
Daniel Russo and James Zou.
\newblock Controlling bias in adaptive data analysis using information theory.
\newblock In Arthur Gretton and Christian~C. Robert, editors, \emph{Proceedings
  of the 19th International Conference on Artificial Intelligence and
  Statistics}, volume~51 of \emph{Proceedings of Machine Learning Research},
  pages 1232--1240, Cadiz, Spain, 09--11 May 2016. PMLR.

\bibitem[Sauer(1972)]{SAUER1972145}
Norbert Sauer.
\newblock On the density of families of sets.
\newblock \emph{Journal of Combinatorial Theory, Series A}, 13\penalty0
  (1):\penalty0 145--147, 1972.
\newblock ISSN 0097-3165.

\bibitem[Shalev-Shwartz and Ben-David(2014)]{shalev2014understanding}
Shai Shalev-Shwartz and Shai Ben-David.
\newblock \emph{Understanding machine learning: From theory to algorithms}.
\newblock Cambridge University Press, 2014.

\bibitem[Shannon(1948)]{Shan49}
Claude~E. Shannon.
\newblock The mathematical theory of communication.
\newblock \emph{The Bell System Technical Journal}, 27:\penalty0 379--423, July
  1948.

\bibitem[Shelah(1972)]{Shelah72}
Saharon Shelah.
\newblock {A combinatorial problem; stability and order for models and theories
  in infinitary languages.}
\newblock \emph{Pacific Journal of Mathematics}, 41\penalty0 (1):\penalty0 247
  -- 261, 1972.

\bibitem[{\c S}im{\c s}ekli et~al.(2020){\c S}im{\c s}ekli, Sener,
  Deligiannidis, and Erdogdu]{simsekli2020hausdorff}
Umut {\c S}im{\c s}ekli, Ozan Sener, George Deligiannidis, and Murat~A Erdogdu.
\newblock Hausdorff dimension, heavy tails, and generalization in neural
  networks.
\newblock In H.~Larochelle, M.~Ranzato, R.~Hadsell, M.~F. Balcan, and H.~Lin,
  editors, \emph{Advances in Neural Information Processing Systems}, volume~33,
  pages 5138--5151. Curran Associates, Inc., 2020.

\bibitem[Steinke and Zakynthinou(2020)]{steinke2020reasoning}
Thomas Steinke and Lydia Zakynthinou.
\newblock {R}easoning about generalization via conditional mutual information.
\newblock In Jacob Abernethy and Shivani Agarwal, editors, \emph{Proceedings of
  Thirty Third Conference on Learning Theory}, volume 125 of \emph{Proceedings
  of Machine Learning Research}, pages 3437--3452. PMLR, 09--12 Jul 2020.

\bibitem[Suzuki et~al.(2020{\natexlab{a}})Suzuki, Abe, Murata, Horiuchi, Ito,
  Wachi, Hirai, Yukishima, and Nishimura]{suzuki2018spectral}
Taiji Suzuki, Hiroshi Abe, Tomoya Murata, Shingo Horiuchi, Kotaro Ito, Tokuma
  Wachi, So~Hirai, Masatoshi Yukishima, and Tomoaki Nishimura.
\newblock Spectral pruning: Compressing deep neural networks via spectral
  analysis and its generalization error.
\newblock In \emph{International Joint Conference on Artificial Intelligence},
  pages 2839--2846, 2020{\natexlab{a}}.

\bibitem[Suzuki et~al.(2020{\natexlab{b}})Suzuki, Abe, and
  Nishimura]{suzuki2020compression}
Taiji Suzuki, Hiroshi Abe, and Tomoaki Nishimura.
\newblock Compression based bound for non-compressed network: unified
  generalization error analysis of large compressible deep neural network.
\newblock In \emph{International Conference on Learning Representations},
  2020{\natexlab{b}}.

\bibitem[Vapnik(1998)]{Vapnik1998}
Vladimir~N. Vapnik.
\newblock \emph{Statistical Learning Theory}.
\newblock Wiley-Interscience, 1998.

\bibitem[Vershynin(2018)]{vershynin2018high}
Roman Vershynin.
\newblock \emph{High-dimensional probability: An introduction with applications
  in data science}, volume~47.
\newblock Cambridge university press, 2018.

\bibitem[Wu and Verdú(2010)]{Wu2010Renyi}
Yihong Wu and Sergio Verdú.
\newblock Rényi information dimension: Fundamental limits of almost lossless
  analog compression.
\newblock \emph{IEEE Transactions on Information Theory}, 56\penalty0
  (8):\penalty0 3721--3748, 2010.

\bibitem[Xu and Raginsky(2017)]{xu2017information}
Aolin Xu and Maxim Raginsky.
\newblock Information-theoretic analysis of generalization capability of
  learning algorithms.
\newblock In \emph{NeurIPS}, 2017.

\bibitem[Zhang et~al.(2017)Zhang, Bengio, Hardt, Recht, and Vinyals]{Zhang16}
Chiyuan Zhang, Samy Bengio, Moritz Hardt, Benjamin Recht, and Oriol Vinyals.
\newblock Understanding deep learning requires rethinking generalization.
\newblock In \emph{International Conference on Learning Representations}, 2017.

\end{thebibliography}
